\documentclass[aap]{imsart}

\RequirePackage{amsthm,amsmath,amsfonts,amssymb}
\RequirePackage[numbers]{natbib}
\RequirePackage[colorlinks,citecolor=blue,urlcolor=blue]{hyperref}
\RequirePackage{graphicx}

\startlocaldefs
\theoremstyle{plain}

\newtheorem{theorem}{Theorem}[section]
\newtheorem{lemma}[theorem]{Lemma}
\newtheorem{proposition}[theorem]{Proposition}

\theoremstyle{remark}
\newtheorem{definition}[theorem]{Definition}
\newtheorem{assumption}[theorem]{Assumption}
\newtheorem{example}{Example}
\newtheorem{problem}[theorem]{Problem}
\newtheorem{remark}[theorem]{Remark}

\usepackage{tikz}
\usepackage{pgfplots}
\usepackage{xcolor}
\usepackage{accents}
\usepackage[normalem]{ulem}
\usepackage{xcolor}
\usepackage{subfig}
\usepackage{graphicx}
\usepackage{enumitem}  
\usepackage{mathrsfs} 
\def\st{\mathrm{s.t.}}
\def\d{\mathrm{d}}
\newcommand{\ubar}[1]{\underaccent{\bar}{#1}}
\usepackage{macro}
\definecolor{darkgreen}{rgb}{0.13, 0.55, 0.13}
\newcommand{\mc}{\mathcal}
\newcommand{\mb}{\mathbb}
\newcommand{\dfeq}{\stackrel{def}=}
\newcommand{\tpose}{^\mathsf{T}}

\def\E{\mathcal{E}}

\def\statespace{\mathcal{S}}

\newcommand{\opintm}{\mathcal{O}}

\newcommand{\met}[1]{\left\|#1\right\|}   
 
\newcommand{\sblvl}{\Gamma} 
\newcommand{\funcls}{\mfk{F}} 
\newcommand{\equicls}{\mathcal{E}}

\definecolor{MYBLUE}{rgb}{0,0,225}
\definecolor{MYGREEN}{rgb}{0.0, 0.55, 0.27}
\newcommand{\grn}{\color{black}}

\endlocaldefs

\begin{document}

\begin{frontmatter}
\title{{Optimal Confidence Intervals\\via Moderate Deviations Theory}}
\runtitle{Optimal interval estimation}

\begin{aug}
\author[A]{\fnms{Arnab}~\snm{Ganguly}\ead[label=e1]{aganguly@lsu.edu}} \and
\author[B]{\fnms{Tobias}~\snm{Sutter}\ead[label=e2]{tobias.sutter@unisg.ch}\orcid{0000-0003-1226-6845}}
\address[A]{Department of Mathematics,
Louisiana State University\printead[presep={,\ }]{e1}}

\address[B]{Department of Economics,
University of St.Gallen\printead[presep={,\ }]{e2}}
\end{aug}

\begin{abstract}
This paper proposes a statistically optimal approach for learning a function value using a confidence interval in a wide range of models, including general non-parametric estimation of an expected loss described as a stochastic programming problem or various SDE models. More precisely, we develop a systematic construction of highly accurate confidence intervals by using a moderate deviation principle-based approach.
It is shown that the proposed confidence intervals are statistically optimal in the sense that they satisfy criteria regarding
exponential accuracy, minimality, consistency, mischaracterization probability, and eventual uniformly most accurate (UMA) property. The confidence intervals suggested by this approach are expressed as solutions to robust optimization problems, where the uncertainty is expressed via the underlying moderate deviation rate function induced by the data-generating process. We demonstrate that for many models these optimization problems admit tractable reformulations as finite convex programs even when they are infinite-dimensional.
\end{abstract}



\end{frontmatter}
\section{Introduction}\label{sec:intro}
Proper understanding of quantitative disciplines like physics, systems biology, mathematical finance, etc. requires building accurate mathematical models and calibrating them according to available data. The model and the associated data can be in a wide array of formats ranging from i.i.d.~observations from a probability distribution to data from a stochastic process (e.g., solution of a stochastic dynamical system) having intricate temporal correlation structure. 
Calibration of the model according to the available data often depends on robust and accurate estimation of relevant cost functions. This in turn requires construction of appropriate confidence intervals for the cost function, which is the main theme of this paper. More precisely,  given a function $J:\Theta\to\mathbb{R}$ and an unknown model parameter $\theta \in \Theta$, the paper aims to learn the function value $J(\theta)$ via a novel confidence interval estimation procedure.

As mentioned, $J$ is typically interpreted as a {\em cost function} associated with a model. For example, in systems biology, the term $J(\theta)$ may describe the energy cost required to steer a chemical species in a reaction system to the desired configuration in finite time, where the chemical reaction dynamics depend on an unknown parameter $\theta$~\cite{ref:wilkinson2011stochastic}. Problems in decision theory often involve working with the long-run average cost $J(\theta)$ of a discrete-time Markov Decision Process $X$ under a policy $\pi_\theta$ parameterized by $\theta$, i.e., 
\begin{equation*}
    J(\theta) = \limsup_{m\to\infty} \frac{1}{m} \sum_{j=0}^{m-1} \mathbb{E}_\theta[c(X_{t_j})],
\end{equation*}
where $c:\statespace\to\mathbb{R}$ denotes some cost-per-stage function and $\mathbb{E}_\theta$ denotes the expectation of the $\statespace$-valued Markov Decision Process under policy $\pi_\theta$; see \cite{hernandez2012discrete} for a rigorous construction. Estimating $J(\theta)$ of the form above, when the underlying transition kernel is unknown, is a key building block in reinforcement learning~\cite{ref:book:meyn-22}. Depending on how the data is generated, the reinforcement learning jargon refers to the estimation problem of $J(\theta)$ as policy evaluation \cite{ref:BerTsi-96,ref:Sutton-1998} or off-policy evaluation \cite{ref:Precup-00,ref:Dudik-14,ref:Levine-20}. 

The parameter  $\theta$ in this paper can be either finite or infinite-dimensional, and we specifically note that the term `parameter' is used here in a rather broad sense. In particular, by taking $\Theta$ to be the space of probability measures (or a  subclass of it), our framework covers so-called non-parametric models. 
An important example in this direction is the non-parametric estimation of an expected loss of a stochastic programming problem. Taking $\Theta$  to be a suitable subclass of $\SC{P}(\statespace)$, the space of probability measures on $\statespace\subset \mathbb{R}^n$, the aim is to estimate 
\begin{equation}\label{eq:stochastic:program}         J(\theta) = \min_{z\in\mathcal{Z}}\mathbb{E}_\theta[\ell(X,z)]
\end{equation}
 where  $\ell:\statespace\times \mathcal{Z}\to\mathbb{R}$ is a given loss function, and the unknown probability measure $\theta \in \Theta$ is the distribution of $X$. Such cost functions and their variations are at the heart of a large class of stochastic optimization problems~\cite{shapiro2014lectures}, and constructing suitable confidence intervals for $J(\theta)$ is necessary for producing distributionally robust solutions (for example, see \cite{delage2010distributionally, ref:BenTal-13, BeGuKa18}).

In this paper, we consider a general function $J:\Theta \rt \R$ and do not restrict ourselves to a specific structure. For estimating the quantity $J(\theta)$ from the available data  via a confidence interval (CI) denoted as $\widehat{\mathcal{I}}_T$, the traditional approach involves finding a CI $\widehat{\mathcal{I}}_T$ with $(1-\alpha)\%$ confidence for a fixed threshold $\alpha$; specifically, for a fixed $\alpha$, $\widehat{\mathcal{I}}_T$ needs to satisfy  $\mathbb{P}_\theta(J(\theta)\in \widehat{\mathcal{I}}_T)\geq 1-\alpha$ for all $\theta \in \Theta$.
The construction of an `ideal' CI of course also needs to take into account other optimality and accuracy properties besides a chosen high-level of confidence.  In the world of CIs corresponding to a fixed threshold $\alpha$,  \textit{uniformly most accurate confidence intervals} (UMA-CIs) are often the ideal prototypes. A UMA-CI of level $1-\alpha$ is a confidence interval $\widehat{\mathcal{I}}_T$ such that $J(\theta)$ lies in $\widehat{\mathcal{I}}_T$ with probability at least $1-\alpha$ for all $\theta\in\Theta$, and in addition, any other $1-\alpha$ confidence interval $\widehat{\mathcal{I}}'_T$ is such that $\mathbb{P}_\theta(J(\theta')\in \widehat{\mathcal{I}}_T) \leq \mathbb{P}_\theta(J(\theta')\in \widehat{\mathcal{I}}'_T)$ for all $\theta'\in\Theta$ such that $J(\theta') \neq J(\theta)$, i.e., confidence intervals with small probabilities of covering wrong values of $J(\theta)$ are preferred, see \cite[Section~3.5]{ref:lehmann-testing}. 
UMA-CIs are typically constructed via uniformly most powerful (UMP) tests. In the case of simple hypotheses testing with known likelihoods, the Neyman-Pearson lemma~\cite[Section 3.2]{ref:lehmann-testing} ensures that the likelihood ratio defines a UMP test statistic, which then leads to a UMA-CI. The Neyman-Pearson lemma cannot be used when it is necessary to work with approximate likelihood functions. More importantly, except for some toy problems, the Neyman-Pearson lemma is no longer applicable for composite hypotheses testing (where the alternate hypothesis consists of a family of probability measures), which is what is required for UMA-CIs. 
Indeed the requirement of uniformly small $\mathbb{P}_\theta(J(\theta')\in \widehat{\mathcal{I}}_T)$ for every $\theta'$ such that $J(\theta')\neq J(\theta)$ and each $T$ is too strong, and it is not possible to characterize cases where UMA-CIs will exist. 

It is thus necessary to work with limiting notions of accuracy and optimality, which is what we use in the paper. The most common way to obtain a CI, $\widehat{\SC{I}}_T$, with asymptotic confidence $(1-\alpha)$, i.e.,  $\lim_{T \rt \infty} \PP_{\theta}(J(\theta) \in \widehat{\SC{I}}_T) \geq 1-\alpha$, is through a CLT. Given an estimator $\widehat \theta_T$, the CLT method gives an interval of the form $[J(\widehat \theta_T) - \widehat \s_T z/\sqrt T, J(\widehat \theta_T) + \widehat \s_T z/\sqrt T]$ with $\widehat \s_T$ being a suitable estimate of variance of $\widehat \theta_T$. The constant $z$ depends on the threshold $\alpha$ and the limiting distribution of $\sqrt T (\widehat \theta_T - \theta)/\sqrt{\text{Var}(\widehat \theta_T)}$, which is typically standard normal. For the case of standard sample average estimate $\widehat \theta_T = \f{1}{T}\sum_{s=1}^T X_s$ of population mean $\theta$ based on i.i.d.~data points $\{X_s:s=1,2,\hdots,T\}$ (with $\EE[X_s] = \theta$), Edgeworth corrected CIs achieving higher accuracy than the typical CLT based CI have been studied in \cite{Hall83, AbSi85, HaJi95} with the last paper studying  uniform coverage bounds (over a suitable class of probability distributions) of these CIs.    

In contrast, our work takes a different route and focuses on {\em exponentially accurate} CIs for which the error probability, $\mathbb{P}_\theta(J(\theta)\notin \widehat{\mathcal{I}}_T)$, decays exponentially fast with a chosen  rate $r>0$ and speed $b_T \rt \infty$. There is almost nothing available in the literature on such intervals. For  $\R^d$-valued i.i.d.~random variables $\{X_s : s=1,2,\dots,T\}$ having density $f$, the paper \cite{MoPe06} constructs an exponentially accurate interval for certain kernel density estimators $\widehat f_T$.  Our goal in this paper is much more ambitious --- we aim to build exponentially accurate CIs in a general framework covering datasets from a wide variety of different models. Of course the construction of an  exponentially accurate CI is just part of the goal --- it is important to construct one which exhibits various optimality properties among its peers. Minimality of such an interval $\widehat{\mathcal{I}}^\star_T$, that is, requiring $\widehat{\mathcal{I}}^\star_T\subset \widehat{\mathcal{I}}_T$ for any other CI $\widehat{\mathcal{I}}_T$ satisfying the same exponential accuracy condition at least for large $T$, is a natural criterion in this respect. 
However, such a stringent notion of minimality is in general difficult to
achieve, especially in broad and model-agnostic settings. While it can be
established in certain structured problems (e.g., finite-state or
i.i.d.\ settings as in \cite{ref:vanParys:fromdata-17}), these approaches rely on specific
properties of the underlying model and do not extend to the general
framework considered in this paper.
In most cases, it is necessary to restrict the search for optimal CIs within a smaller class and view optimality through a bit different lens. Towards this end, we search for an ideal candidate in the class of CIs that are based on a given point estimator $\widehat \theta_T$.  More precisely, the confidence intervals of interest to us are the form $\widehat{\mathcal{I}}_T=\mathcal{I}_T(\widehat \theta_T)$ where 
$\mathcal{I}_T(u)= [\ubar{H}_{T,r}(u), \bar{H}_{T,r}(u)]$, with $\ubar{H}_{T,r}, \bar{H}_{T,r}$ belonging to suitable classes of functions from $\esttvs$ to $\R$. Here $\esttvs$ is the space where the estimator $\widehat \theta_T$ takes its values and could be larger than the parameter set $\Theta$.  The primary objective of the paper is to develop a systematic approach for constructing an interval-family $\{\mathcal{I}^\star_T(\cdot)= [\ubar{J}_{T,r}(\cdot), \bar{J}_{T,r}(\cdot)]: T>0\}$ that is an ideal choice from multiple angles. 

At this stage it is useful to make a note about a couple of terminologies that we will use in the paper. For convenience, we will refer to $\{\mathcal{I}_T = [\ubar{H}_{T,r}, \bar{H}_{T,r}] : T>0\}$ as an {\em interval-family}, although a longer term, {\em interval-function-family}, perhaps is more appropriate. Notice that  for each  point $u$ in the domain of the functions $\ubar{H}_{T,r}$ and $\bar{H}_{T,r}$ (with $\ubar{H}_{T,r} \leq \bar{H}_{T,r}$ pointwise)  and $T>0$, $\mathcal{I}_T(u) = [\ubar{H}_{T,r}(u), \bar{H}_{T,r}(u)]$ is a legitimate interval, and the term {\em confidence interval} arising in this context will be reserved for the (random)  interval, $\mathcal{I}_T(\widehat \theta_T) = [\ubar{H}_{T,r}(\widehat \theta_T), \bar{H}_{T,r}(\widehat \theta_T)]$. 

We now provide a list of natural accuracy criteria that we want our optimal interval-family $\{\mathcal{I}^\star_T= [\ubar{J}_{T,r}, \bar{J}_{T,r}]:T>0\}$ to satisfy.

\begin{enumerate}[label=(\roman*)]
    \item \label{eq:item:1:conf:interval} {\em Exponential accuracy}: Given a rate $r \in (0,\infty)$ and speed $\{b_T\}_{T\in\mathbb{N}}$ such that $b_T \rt \infty$ of exponential decay, the confidence interval $\mathcal{I}^\star_T(\widehat \theta_T)$ is such that
     \begin{equation*}
        \mathbb{P}_\theta\lf(J(\theta)\notin \mathcal{I}^\star_T(\widehat \theta_T)\ri) = O(e^{-rb_T})\quad \forall \theta\in\Theta.
    \end{equation*}
\vs{.1cm}

    \item \label{eq:item:1:conf:minimality} {\em Minimality}: {\grn Let  $\mathcal{I}_T(\theta) = [\ubar{H}_{T,r}(\theta), \bar{H}_{T,r}(\theta)]$ be an interval satisfying the property \ref{eq:item:1:conf:interval}. Then for any inflation factor $\alpha_T =o(a_T\equiv \sqrt{T/b_T})$, the inflated interval $\alpha_T\mathcal{I}_T(\theta)$  will eventually (as $T$ is sufficiently large) contain $\alpha_T\mathcal{I}^\star_T(\theta)=[\alpha_T\ubar{J}_{T,r}(\theta), \alpha_T\bar{J}_{T,r}(\theta)]$ for every $\theta \in \Theta$; that is, for every $\theta$, $\alpha_T\ubar{J}_{T,r}(\theta)$ will be eventually larger than $\alpha_T\ubar{H}_{T,r}(\theta)$, and $\alpha_T\bar{J}_{T,r}(\theta)$ eventually smaller than $\alpha_T\bar{H}_{T,r}(\theta)$. This highlights that the natural length scale for an optimal interval-family is $1/a_T$. The significance of the factor $a_T$ will become clear when we discuss the assumptions on the estimator $\widehat \theta_T$.}
\vs{.3cm}

    \item \label{eq:item:intro:asymptotic:consistency}{\em Consistency}: For every $\theta\in\Theta$ the  CI $\mathcal{I}^\star_T(\widehat \theta_T)$ is such that under $\PP_\theta$, both its lower and upper endpoints, $\ubar{J}_{T,r}(\widehat \theta_T), \bar{J}_{T,r}(\widehat \theta_T)\stackrel{T\rt \infty}\Rt J(\theta)$ either almost surely or in probability. In fact, since the length of the optimal CI is expected to be $O(1/a_T)$, the convergences should happen with rate $\alpha_T = O(a_T)$.   
\vs{.3cm}

     \item \label{eq:item:4:mis:chracterization} {\em Mischaracterization probability}: 
    Consider any fixed $\theta \in \Theta$. For any $\tilde\theta \in \Theta$ with 
$J(\tilde\theta) \neq J(\theta)$, the interval family $\mathcal{I}^\star_T$ is such
that the probability of 
incorrectly including $J(\tilde\theta)$ in the confidence interval does 
not decay exponentially fast. More precisely,
\[
\lim_{T\to\infty} \frac{1}{b_T} 
\log \mathbb{P}_\theta\big(J(\tilde\theta) \notin \mathcal{I}^\star_T(\widehat\theta_T)\big) = 0.
\]
Thus, while the true value $J(\theta)$ is covered with exponentially high
probability, any incorrect value $J(\tilde\theta)$ still has a
non-negligible probability, on the exponential scale, of being excluded.

\vs{.3cm}
    \item \label{eq:item:3:conf:UMA} {\em Eventually UMA}:
Any interval family $\{\mathcal{I}_T\}
$ with the property \ref{eq:item:1:conf:interval} will eventually (as $T$ is sufficiently large) be such that for any $\theta\in\Theta$
\begin{equation*}
    \mathbb{P}_\theta\lf( J(\tilde\theta) \in \mathcal{I}^\star_{T}(
    \widehat
    \theta_T)\ri) \leq    \mathbb{P}_\theta\lf( J(\tilde\theta) \in \mathcal{I}_{T}(\widehat\theta_T)\ri) \quad \forall \tilde\theta\in\Theta : J(\tilde\theta)\neq J(\theta).
\end{equation*}
\end{enumerate}
These criteria will be stated and discussed rigorously in Section \ref{sec:math:framework} --- see Problem \ref{prob:interval} and its solution presented in our main results, Theorems~\ref{thm:minimal:interval} and \ref{th:prop-J}. Here, we just add a few high-level remarks about these five criteria. 
The criterion~\ref{eq:item:4:mis:chracterization} is an asymptotic version of what in the literature is often referred to as an \textit{unbiased confidence interval}, see \cite[Section~5.5]{ref:lehmann-testing}.
Both criteria \ref{eq:item:1:conf:minimality} and \ref{eq:item:3:conf:UMA} pertain to the smallness of the size of the desired interval, but one does not imply the other. {\grn Property~\ref{eq:item:1:conf:minimality} ensures the optimality of the functions $\ubar{J}_{T,r}, \bar{J}_{T,r}$ defining the lower and upper bounds of the interval-family, by saying that the corresponding interval family is the smallest in limit even when zoomed or inflated by appropriate amount. Property~\ref{eq:item:3:conf:UMA}, on the other hand, requires that, for all sufficiently large $T$, the proposed confidence interval minimizes, among all competing interval families satisfying~\ref{eq:item:1:conf:interval}, the probability of covering any incorrect value $J(\tilde\theta)\neq J(\theta)$, uniformly over $\theta,\tilde\theta\in\Theta$."
 }   
While nonasymptotic versions of the criteria \ref{eq:item:3:conf:UMA} as well as  \ref{eq:item:4:mis:chracterization} are standard textbook concepts and their corresponding versions at least for CIs with usual asymptotic level $(1-\alpha)$ is well-known, except for simple models, their uses for establishing optimality of a CI in limiting regimes are rather rare. 
Minimality~\ref{eq:item:1:conf:minimality} and closely related notions of tightness or least conservatism have previously been studied in data-driven DRO and
confidence-set estimation; see, for example,
\cite{ref:vanParys:fromdata-17,ref:Sutter-19,ref:Bennouna-24}.

{\grn
We now discuss the conditions needed for the above properties. Of course, consistency of the underlying estimator $\widehat{\theta}_T$ is a bare minimum requirement for consistency of the confidence interval. In fact, to establish \ref{eq:item:1:conf:interval} and \ref{eq:item:1:conf:minimality}, one needs more detailed information about the rate at which $\widehat{\theta}_T$ converges to the true parameter $\theta$. It is not hard to see that the exponential decay of the error probability in \ref{eq:item:1:conf:interval}, with rate $r$ and speed $b_T$, requires the probabilities of deviations, $\widehat{\theta}_T-\theta$, to exhibit exponential decay with the same speed. This happens in an appropriate scaling regime. It is well known that such exponential decay occurs when the deviations  $\widehat{\theta}_T-\theta$ are of order $O(1/a_T)$, where $a_T=\sqrt{T/b_T}$. Mathematically, this means that the rescaled estimation error $a_T(\widehat{\theta}_T-\theta)$ satisfies a {\em large deviation principle} (LDP) with speed $b_T$ and rate function $I^M$. 
Here we note that when $a_T\equiv 1$, one says that $\widehat{\theta}_T$ satisfies an LDP, whereas it is said to satisfy a {\em moderate deviation principle} (MDP) when $1\ll a_T\ll \sqrt{T}$ (or equivalently, $1\ll b_T\ll T$).

This paper focuses exclusively on the MDP regime, in which the  deviations of $\widehat{\theta}_T$ from the true parameter $\theta$ are of order $O(1/a_T)$. This also intuitively explains why the optimal interval family is expected to have length of the same order. As we shall see, the MDP rate function $I^M$ lies at the core of the construction of the optimal interval.
}

It is clear that an MDP, by analyzing the deviations that are neither too near the limit nor too far from it, bridges the gap between a central limit theorem and the usual LDP.  Moderate deviation asymptotics have been frequently used in mathematical statistics with early works focusing on sample averages based on i.i.d.~sequences and arrays \cite{RuSe65, Mich74, Amos80, DeZa97}. More general framework including that of general topological spaces, weakly dependent sequences, Markov processes, multiscale systems, stochastic PDEs have been considered in various works, a very limited list of references for which is  \cite{Ghos74, GhBa77, DeBa81, DeAc92, Wu95, Gram97, Arco03, DeAc97, DeAc98, LiSp99, Djel02, Guil01, Guil03, BuDuGa16, GaSu21}. 
MDP for different statistical estimators used in regression analysis, time-series analysis, inference of specific stochastic differential equations have been obtained in \cite{DjGuWu99, MoPeTh08, ref:Yu-09, ref:Gao-09, LiZh13}.

\textbf{Further related work.}
The notion of minimality~\ref{eq:item:1:conf:minimality} in combination with exponential accuracy is inspired by the recent works~\cite{ref:Sutter-19,ref:salac-2025}, where the goal is to derive high confidence upper bound estimates $\bar H_{T,r}(\widehat\theta_T)$ to $J(\theta)$ similar to \ref{eq:item:1:conf:interval}, but with speed $b_T=T$, while ensuring a slightly modified concept to \ref{eq:item:1:conf:minimality} regarding minimality. The estimator $\bar H_{T,r}(\widehat\theta_T)$ in \cite{ref:Sutter-19} relies on $\{\widehat \theta_T\}$ satisfying a large deviation principle and, importantly, the function $\bar H_{T,r}$ does not depend on time. Consequently, the resulting estimator lacks asymptotic consistency, i.e., $\bar H_{T,r}(\widehat\theta_T)$ does not converge to $J(\theta)$ as $T\to \infty$. Another limitation of \cite{ref:Sutter-19} is that the results rely on specific properties of the large deviation rate function, such as the ``radial monotonicity" (see \cite[Definition 3.3]{ref:Sutter-19}), which is a restrictive assumption. Our approach works under much milder assumptions on the underlying rate function $I^M$. The paper \cite{ref:Sutter-19} generalizes an approach restricted to i.i.d.~data with finite support~\cite{ref:vanParys:fromdata-17}.
A more recent work~\cite{ref:Amine-21} extends the approach of~\cite{ref:vanParys:fromdata-17} intending to achieve asymptotic consistency of the high-confidence upper bound estimates by exploiting a moderate deviation principle ensuring asymptotic consistency of the estimators. This work, however, is restricted to the setting of finite state i.i.d.~processes and uses a slightly different notion of minimality in \ref{eq:item:1:conf:minimality} compared to us.

When $J(\theta)$ is described by an expected loss with respect to a probability measure $\theta$, i.e., $J(\theta) = \mathbb{E}_{\theta}[\ell(X)]$, and $\widehat\theta_T$ denotes the empirical probability measure constructed from $T$ i.i.d.~samples drawn from $\theta$, \cite{ref:Lam-16, ref:Duchi-Glynn-21} discussed how to construct functions $\ubar H_{T,r}$ and $\bar H_{T,r}$ such that the corresponding CI, $\mathcal{I}_T(\widehat\theta_T) = [\ubar H_{T,r}(\widehat\theta_T),\bar H_{T,r}(\widehat\theta_T)]$, provides an asymptotically exact coverage corresponding to a fixed threshold  $\alpha$, i.e., $\lim_{T\to\infty} \mathbb{P}_\theta(J(\theta)\notin \mathcal{I}_T(\widehat\theta_T)) = \alpha$, and such that $\mathcal{I}_T(\widehat\theta_T) \to \{J(\theta)\}$ as $T\to\infty$.
 This of course differs from our framework of exponential accuracy as described in \ref{eq:item:1:conf:interval}. Moreover, \cite{ref:Lam-16,ref:Duchi-Glynn-21} did not study optimality of their proposed intervals. 

The above CIs are of the form
\begin{equation}\label{eq:DRO:intro}
   \ubar{H}_{T,r}(\widehat \theta_T) = \inf_{\theta\in\Theta} \{J(\theta) : \mathsf{d}_T(\widehat\theta_T,\theta)\leq r\},
   \quad \bar{H}_{T,r}(\widehat \theta_T) = \sup_{\theta\in\Theta} \{J(\theta) : \mathsf{d}_T(\widehat\theta_T,\theta)\leq r\},
\end{equation}
where $\mathsf{d}_T$ is some distance-like function.   Interestingly, the functions $\ubar{J}_{T,r}, \bar{J}_{T,r}$ defining the optimal confidence interval bounds in our paper are also of the above form, with the noteworthy feature that $\mathsf{d}_T$ depends on the data-generating process through the MDP rate function. %
Intuitively, the constraint $\mathsf{d}_T(\widehat\theta_T,\theta)\leq r$ forces the optimization variable $\theta$ to be ``close" to the estimator $\widehat \theta_T$ in a sense that will be made precise within this paper. In other words, an estimator of the form $\bar{H}_{T,r}(\widehat \theta_T)$ in \eqref{eq:DRO:intro}  describes the worst-case value of $J$ that can attain in a neighborhood close to the estimator $\widehat \theta_T$. Estimators of the form~\eqref{eq:DRO:intro} are popular objects in robust optimization \cite{ref:benTal-09, ref:Bertsimas-11} and distributionally robust optimization \cite{ref:DROtutorial-19, ref:Rahimian-19}. While there is a vast amount of literature on the choice of $\mathsf{d}_T$, it is usually chosen ad hoc and common examples are scaled $\phi$-divergences (e.g., relative entropy) or integral probability metrics (e.g., Wasserstein distances). Then, for a fixed structure of $\mathsf{d}_T$ and asymptotically fixed threshold $\alpha$, the statistical properties of $\bar{H}_{T,r}(\widehat \theta_T)$ are analyzed, see \cite{ref:Lam-16, ref:Duchi-Glynn-21, ref:MohajerinEsfahani-2017, ref:Blanchet-23}. 
In particular, to ensure consistency of the two estimators $\ubar H_{T,r}(\widehat\theta_T)$ and $\bar H_{T,r}(\widehat\theta_T)$, the ambiguity set $\{\theta\in\Theta : \mathsf{d}_T(\widehat \theta_T,\theta)\leq r\}$ asymptotically should reduce to a singleton described by the true unknown parameter $\theta$ \cite{ref:Duchi-Glynn-21,ref:Blanchet-21}.
Our approach is fundamentally different. We use a top-down approach --- starting from a list of natural criteria, \ref{eq:item:1:conf:interval} - \ref{eq:item:3:conf:UMA}, that one expects the  functions, $\ubar{J}_{T,r}$ and $\bar{J}_{T,r}$,   defining the lower and upper bounds of the optimal CI to satisfy,     we describe a method to construct them. In our case, the data-generating process, or the rate function of the underlying MDP to be precise, naturally induces the optimal structure of $\mathsf{d}_T$ in \eqref{eq:DRO:intro}. This, for example, allows us to quantify the optimal rate at which the ambiguity set shrinks to a singleton. For the non-parametric estimation problem involving i.i.d data, the MDP rate function dictates that the optimal choice of $\mathsf{d}_T$ is a (scaled) $\chi^2$-divergence (c.f. Section \ref{ssec:non-parametric:iid}).

A work particularly close to our minimality perspective is
\cite{ref:Bennouna-24}. The authors study confidence-set estimators under
adversarial contamination and show that a KL--TV ambiguity set is uniformly
minimal: every regular confidence-set estimator satisfying the same
exponential coverage requirement must contain their proposed set. Evaluating a
scalar functional $J$ over these sets therefore yields a corresponding
endpointwise minimality property for the induced interval.
Their setting is based on empirical distributions in an i.i.d.\ contamination
model and the usual large-deviation speed. Our framework instead accommodates
general estimators and potentially dependent stochastic processes, and operates
in the moderate-deviation regime $1\ll b_T\ll T$, where exponential accuracy
coexists with a vanishing interval width of order
$\sqrt{b_T/T}$. Moreover, we establish local-uniform minimality together with
consistency, the mischaracterization property, and eventual UMA for two-sided
confidence intervals of a general functional $J(\theta)$. Thus, our
contribution lies not in introducing minimality itself, but in deriving a
general MDP-based construction that simultaneously provides these guarantees.

\vspace{2mm}

\textbf{Structure:} The rest of the paper proceeds as follows. Section~\ref{sec:heuristic:approaches} describes heuristic approaches to derive a confidence interval satisfying the exponential accuracy condition~\ref{eq:item:1:conf:interval} based on the central limit theorem. We then, in Section~\ref{sec:math:framework}, mathematically introduce a framework to rigorously address the derivation of confidence intervals as motivated by the properties \ref{eq:item:1:conf:interval}-\ref{eq:item:3:conf:UMA} via our main results, Theorems~\ref{thm:minimal:interval} and \ref{th:prop-J}. In particular, we show as our main result that these optimal confidence intervals can be explicitly constructed using a moderate deviation rate function.
Section~\ref{sec:proof:main:result} then derives the formal proofs of the main results. Finally, Section~\ref{sec:models} presents several model classes where our main results apply and shows how the respective optimal confidence intervals can be efficiently computed. Section~\ref{sec:conclusion} concludes.
Basic definitions on large and moderate deviation theory are provided in Appendix~\ref{appA:MDP:basics} and auxiliary proofs are provided in Appendix~\ref{appB}.

\vspace{2mm}

\textbf{Notation:} For a set $A$ we denote its closure by $\bar A$ and its interior by $A^\circ$. For  two families of non-negative numbers, $\{a_T\}$ and $\{b_T\}$,  we use $ a_T \ll b_T$ to express that $a_T /b_T \to 0$  as $T \to\infty$. A family of random variable $\{X_T\}$ converging in probability with respect to $\PP$ to a random variable $Y$, is denoted as $X_T \stackrel{\PP} \rt Y$ as $T \rt \infty$.

For a topological space $\esttvs$, $\mathcal{P}(\esttvs)$ will denote the space of probability measures on $\esttvs$, and $\mathcal{M}(\esttvs)$ will denote the vector space of finite signed measures on $\esttvs$. For a mapping $f:\esttvs\to\mathbb{R}$ and a set $\esttvs_0\subset \esttvs$, we denote by $f|_{\esttvs_0}: \esttvs_0 \rt \R$ the restriction of $f$ to $\esttvs_0$.

For each $T>0$, let the functions $\ubar J_T, \bar J_T: \esttvs \to \R$ be such that $\ubar J_T(u) \leq \bar J_T(u)$ for all $u \in \esttvs$. Then the family $\{[\ubar J_T, \bar J_T] : T>0\}$ which define an interval $[\ubar J_T(u), \bar J_T(u)]$ for each $u\in\esttvs$ and $T>0$, is referred to as an {\em interval-family}. 

Let $\esttvs$ be a topological vector space equipped with a norm $\met{\cdot}$.  For any $a>0$ and $u \in\esttvs$, the set $B(u,a)$ denotes the ball centred at $\theta$ with radius $a$, i.e., $B(u,a) = \{u'\in\esttvs: \met{u'-u}\leq a\}$.
For a set $\esttvs_0 \subset \esttvs$, and $a \in \R, u\in \esttvs$, the set $a \esttvs_0+u$ is naturally defined as
\begin{align*}
a \esttvs_0+u \equiv \lf\{a\vart+u: \vart \in \esttvs_0\ri\}.
\end{align*}

\section{Discussion of standard approaches to estimation}\label{sec:heuristic:approaches}
\

We start by discussing whether standard estimators of $J(\theta)$ can be adapted to work in our desired setting of exponential accuracy \ref{eq:item:1:conf:interval}. 

\vspace{2mm}

 \np
{\bf Limitation of the canonical estimator:}
From the point estimation perspective, a natural question is if given an estimator $\widehat \theta_T$ of $\theta$ and a function $J$, the canonical estimator $J(\widehat \theta_T)$ is an ideal choice for estimating $J(\theta)$. The estimator $J(\widehat \theta_T)$ is of course consistent (assuming that $\widehat \theta_T$ is consistent, and $J$ is continuous), but the answer, perhaps counter-intuitively, is ``not always". Indeed $J(\widehat \theta_T)$ is not a good candidate for estimating  $J(\theta)$ in situations where the consequences of the true cost being greater than the estimated one are catastrophic. For a specific scenario, consider the constrained optimization problem of the form 
\begin{align}
\min_{x \in \E_c} g(x),\quad  \text{ where }\  \E_c = \{x \in \E: J(\theta,x) \leq c\} .
\end{align}
In absence of knowledge of $\theta$, one needs to use a suitable estimator $\widehat J_T(x)$ for $J(\theta,x)$. Accuracy of the approximation of the original solution by the solution of the data-driven version of the problem,
$\min_{x \in \E} g(x)$ subject to $\widehat {J}_T(x)\leq c,$ as well as its robustness requires the estimator $\widehat  J_T(x)$ not only to be consistent but be such that the probability of $J(\theta,x)$ being larger than $\widehat  J_T(x)$ is very small. This  ensures that with high probability the estimated constraint region, $\{x:\widehat {J}_T(x)\leq c \}$, is an accurate representation of actual constraint region, $\E_c$, and is in fact contained in it for large $T$.

In other words, in situations like above we really need an upper confidence bound $\widehat  J_T$ for $J(\theta)$, which, besides being a consistent estimator of $J(\theta)$, will ensure that $\PP_{\theta}(J(\theta)>\widehat  J_T)$ is small -- exponentially small in our framework; that is, for a given rate of decay $r$ and speed $b_T \rt \infty$, we seek an estimator $\widehat  J_T \equiv \widehat  J_{T,r}$
\begin{equation*}
  \PP_{\theta}\lf(J(\theta) \notin (-\infty, \widehat  J_{T}]\ri) = \PP_{\theta}\lf(J(\theta)> \widehat  J_T\ri)  = O(e^{-rb_T}).
\end{equation*}
 Rigorously stated, the upper bound needs to ensure that 
\begin{equation}\label{eq:constraint:overestimation}
    \limsup_{T\to\infty} \frac{1}{b_T} \log \mathbb{P}_\theta \lf( J(\theta) > \widehat J_T \ri) \leq -r \quad \forall \theta\in\Theta.
\end{equation}
But $\widehat  J_T \equiv J(\widehat \theta_T)$  does not satisfy \eqref{eq:constraint:overestimation}. To see this first note  that  the following trichotomy holds:
\begin{equation}\label{eq:trichotomy}
    \lim_{T\to\infty}\mathbb{P}_\theta\lf( J(\theta) > J(\widehat \theta_T)+\kappa\ri) =
    \begin{cases}
1,& \quad \text{ if } \kappa < 0\\
\frac{1}{2}, &\quad \text{ if } \kappa = 0\\
0,& \quad \text{ if } \kappa >0.
\end{cases}
\end{equation}
Here the first and the last equality simply follow from consistency of $J(\widehat \theta_T)$, the second requires the additional assumption that $\widehat \theta_T$ satisfies a CLT (with respect to $\PP_\theta$, for every $\theta$) with some asymptotic covariance matrix $S(\theta)$ and $J$ is continuous. Indeed in this case
\begin{equation*}
   \lim_{T\to\infty} \mathbb{P}_\theta\lf(J(\theta) > J(\widehat \theta_T)\ri) = \lim_{T\to\infty} \mathbb{P}_\theta\lf(\sqrt{T}(J(\widehat \theta_T) - J(\theta))<0\ri) = \frac{1}{2}
\end{equation*}
and consequently for $\kappa \leq 0$, 
\begin{align*}
\limsup\limits_{T\to\infty}\frac{1}{b_T} \log \mathbb{P}_\theta\lf( J(\theta) > J(\widehat \theta_T)+\kappa\ri) = 0.
\end{align*}
 While adding a fixed bias $\kappa$ to $J(\widehat \theta_T)$ may result in the exponential decay of error probabilities, the desirable property of  consistency \ref{eq:item:intro:asymptotic:consistency} is lost. Even if we decide to forgo consistency, it is not clear what an allowable size of $\kappa$ representing the fixed or asymptotic bias is for a given problem. \\

 \np
{\bf CLT Approach:} It is thus imperative that the parameter $\kappa \equiv \kappa_T$ is scaled with time $T$ no matter which lens we use to view the problem. Indeed for the interval estimation, a natural approach is to search for CIs of the form $[J(\widehat \theta_T)-\kappa_T, J(\widehat \theta_T)+\kappa_T]$ (or $(-\infty, J(\widehat \theta_T)+\kappa_T]$ for the one-sided upper CI problem). A common way to achieve an asymptotic level $(1-\alpha)$ confidence interval for a {\em fixed} level $\alpha$ is through the CLT. Let $\theta$ be the true value of the parameter, and assume that $\widehat\theta_T$ satisfies a CLT with limiting covariance matrix $S(\theta)$, that is, $\sqrt{T}(\widehat\theta_T - \theta)\overset{d}{\longrightarrow} \mathcal{N} (0, S(\theta))$ (w.r.t.~$\PP_{\theta}$). If $J$ is continuously differentiable at the point $\theta$ and $\nabla J(\theta) \neq 0$, then standard laws of probability guarantee that
\begin{equation*}
\frac{\sqrt T (J(\widehat\theta_T) - J(\theta))}{ \sqrt{\nabla J(\widehat \theta_T)^\top S(\widehat\theta_T)\nabla J(\widehat \theta_T)}} \overset{d}{\longrightarrow} \mathcal{N}(0, 1).
\end{equation*}
Consequently,  a central limit confidence interval (CLT-CI) of asymptotic level $\alpha$ for $J(\theta)$ is given by
\begin{equation}\label{eq:CLT:CI}
\mathcal{I}^{\mathsf{CLT}}_{T,\alpha}(\widehat \theta_T) = \Big[J(\widehat \theta_T) - \kappa^{\mathsf{CLT}}_T(\alpha), J(\widehat \theta_T) + \kappa^{\mathsf{CLT}}_T(\alpha)\Big],
\end{equation}
with
$$\kappa^{\mathsf{CLT}}_T(\alpha) \equiv \Phi^{-1}(1 - \alpha/2)\sqrt{\nabla J(\widehat \theta_T)^\top S(\widehat\theta_T)\nabla J(\widehat \theta_T)}/ \sqrt{T},$$
where $\Phi$ is the distribution function of a standard normal random variable. This guarantees that $\lim_{T\rt \infty}\PP_\theta(J(\theta) \notin \mathcal{I}^{\mathsf{CLT}}_{T,\alpha}(\widehat \theta_T)) \leq \alpha$. But a CLT-based interval does not lead to the desired exponential decay of the error probability, although a practitioner can intuitively use $\alpha=e^{-rb_T}$ with the hope that the resulting CI, $ \mathcal{I}^{\mathsf{CLT}}_{T,\alpha=e^{-rb_T}}(\widehat \theta_T)$, at least satisfies the {\em exponentially high confidence estimate}~\ref{eq:item:1:conf:interval} in the introduction, i.e.,  $\PP_\theta(J(\theta) \notin \mathcal{I}^{\mathsf{CLT}}_{T,\alpha=e^{-rb_T}}(\widehat \theta_T)) = O(e^{-rb_T})$, if not \ref{eq:item:1:conf:minimality}.  However, this is purely a heuristic technique and is not mathematically guaranteed to work.  The level $\alpha$ needs to be {\em constant} (with respect to time) for the CLT-CI in \eqref{eq:CLT:CI} --- one cannot simply substitute $\alpha = e^{-rb_T}$ and rigorously justify that the above {\em high confidence estimate} holds for $\mathcal{I}^{\mathsf{CLT}}_{T,\alpha}(\widehat \theta_T)$. 
Moreover, achieving the guarantees \ref{eq:item:4:mis:chracterization} or \ref{eq:item:3:conf:UMA} via this CLT-CI remains entirely unclear. But notice that length of this heuristic CI, $\mathcal{I}^{\mathsf{CLT}}_{T,\alpha=e^{-rb_T}}(\widehat \theta_T))$, is $O(\Phi^{-1}(1-e^{-rb_T}/2))/\sqrt T = O(1/a_T)$  where $a_T = \sqrt{T/b_T}$ and we used the approximation $\Phi^{-1}(1-\alpha) = |\Phi^{-1}(\alpha)| \approx \sqrt{-2\log(\alpha)}$ for small $\alpha$ (see \cite{ref:Schloeter-12}). 
Thus even though the above CLT-based CI is not based on rigorous mathematical principles, it does provide an important insight --- the length of the optimal exponentially accurate CI is likely $O(1/a_T)$.

The reason that the above method does not assure exponential decay of error probabilities is because the CLT is meant to look at deviations near the mean which are of order $O(1/\sqrt T)$. This is not where exponential decay occurs; it occurs when the deviations from the mean are larger -- of the order $O(1/a_T)$ with $a_T \ll \sqrt T$. It is thus natural that a rigorous methodology for constructing the optimal interval estimators satisfying the properties~\ref{eq:item:1:conf:interval}-\ref{eq:item:3:conf:UMA} needs to be based on large deviation analysis rather than the CLT. As customary, we will reserve the phrase {\em large deviation principle (LDP)} when $a_T=1$ and {\em moderate deviation principle (MDP)} (which also falls under the realm of large deviation analysis) when $1 \ll a_T \ll \sqrt T$.
An MDP of $\widehat \theta_T$ says that probability of deviation of order $1/a_T$ from the true value $\theta$ decays exponentially with speed $b_T$. This intuitively explains why the optimal exponentially accurate CI for $J(\theta)$ is expected to have length of $O(1/a_T)$.

However, constructing such optimal exponentially accurate confidence intervals is considerably more subtle, and the corresponding notions of optimality require careful formulation.

Before we do that in the next section, we conclude by introducing the following running example of an Ornstein-Uhlenbeck process where we show the explicit form of the optimal exponentially accurate CI.

\begin{example}[Estimation of stationary variance of an Ornstein-Uhlenbeck process] \label{ex:OU:part1}
Consider an Ornstein-Uhlenbeck (OU) process defined as the SDE
\begin{equation}\label{eq:OU:SDE:mot}
    \d X_t = -\theta X_t \d t + \d W_t, \quad X_0 = 0,
\end{equation}
where $\theta\in\Theta=\mathbb{R}_{>0}$ is an unknown drift parameter and $W$ a Brownian motion
independent of $X_0$. Our goal is to find a CI for the stationary variance of the process (see~\cite[Example~6.8]{ref:karatzas-91}) given by
\begin{equation}\label{eq:OU:asymptotic:variance:mot}
    J(\theta) = \lim_{T\to\infty} \frac{1}{T}\sum_{t=0}^{T-1} \left( \mathbb{E}_\theta[X_t^2] - \mathbb{E}_\theta[X_t]^2\right)= \frac{1}{2\theta}.
\end{equation}
We base the estimation of the CI on the maximum likelihood estimator (MLE) $\widehat \theta_T\in\esttvs=\mathbb{R}$, which is constructed from a single trajectory of data $\{X_t\}_{t\in[0,T]}$ by 
\begin{equation}\label{eq:MLE:OU}
    \widehat \theta_T = -\frac{\int_0^T X_t \d X_t}{\int_0^T X^2_t \d t} 
    = -\frac{X_T^2-T}{2\int_0^T X^2_t \d t}.
\end{equation}
Here the second equality, easily obtained through stochastic integration by parts,  describes an alternative computable expression of the MLE without the stochastic integral (see \cite[Remark~2.1]{ref:Kleptsyna-02}).

Our MDP based approach, which we will describe shortly, in this case leads to the confidence interval 
\begin{equation}\label{eq:optimal:CI:OU} \mathcal{I}^\star_{T,r}(\widehat\theta_T)\!=\![J(\widehat\theta_T) \!+\!   \kappa^{-}_T, J(\widehat\theta_T) \!+\! \kappa^{+}_T], \quad \kappa^{\pm}_T = \frac{1}{\widehat\theta_T^2}\left( r_T \pm \sqrt{r_T^2+\widehat\theta_T r_T} \right),
\end{equation}
where $r_T = r b_T/T$.
The length of this interval is
\begin{equation*}
\kappa^{+}_T - \kappa^{-}_T = \frac{2 \sqrt{r_T^2 + r_T \widehat \theta_T}}{\widehat\theta_T^2}
\sim \frac{2 \sqrt{r_T}}{\widehat{\theta}_T^{3/2}}.
\end{equation*}
The details of this calculation will be presented in Example \ref{ex:OU:part2}. 
Not surprisingly, the length of this confidence interval is (for large $T$) the same as the heuristic CLT-CI, $\mathcal{I}^{\mathsf{CLT}}_{T,\alpha = e^{-r b_T}}(\widehat \theta_T)=[J(\widehat\theta_T) - \kappa^{\mathsf{CLT}}_T(e^{-r b_T}),J(\widehat\theta_T) + \kappa^{\mathsf{CLT}}_T(e^{-r b_T})] $, which is not theoretically guaranteed to give the exponential decay.

Indeed, it is known that the MLE~\eqref{eq:MLE:OU} satisfies the following CLT, $\sqrt{T}(\widehat\theta_T - \theta) \stackrel{d}{\rightarrow} \mathcal{N}(0, 2\theta)$  \cite{ref:basawa1980statistical}. Thus the heuristic CLT-CI corresponding to $O(e^{-rb_T})$-decay for error probability is given as in \eqref{eq:CLT:CI} with 
\begin{equation}\label{eq:OU:CLT:CI}
\kappa^{\mathsf{CLT}}_T(\alpha) \equiv \f{\Phi^{-1}(1 - \alpha/2)}{\widehat \theta_T^{3/2} \sqrt{2T}}, \qquad \alpha = e^{-r b_T}. 
\end{equation}
Therefore, the length of this interval ${\mathcal{I}}^{\mathsf{CLT}}_{T,\alpha=e^{-r b_T}}(\widehat\theta_T)$ in \eqref{eq:CLT:CI} scales as
\begin{equation}\label{eq:interval:length:CLT}
2\kappa^{\mathsf{CLT}}_T(\alpha)  = 2\f{\Phi^{-1}(1 - \alpha/2)}{\widehat \theta_T^{3/2} \sqrt{2T}} 
 = 2\f{|\Phi^{-1}(\alpha/2)|}{\widehat \theta_T^{3/2} \sqrt{2T}}
 \sim \frac{2 \sqrt{r_T}}{\widehat \theta_T^{3/2}}, 
\end{equation}
where the second equality used the symmetry of the standard normal density and the third step used the approximation, $\Phi^{-1}(\alpha) \approx \sqrt{-2\log(\alpha)}$ for small $\alpha$ (see \cite{ref:Schloeter-12}), here $\alpha = e^{-r b_T}$. 
The MDP-based interval \eqref{eq:optimal:CI:OU} and the heuristic CLT interval have the same
first-order length. More precisely,
\begin{equation*}
\frac{
\kappa_T^+
-
\kappa_T^-
}{
2\kappa_T^{\mathsf{CLT}}(e^{-rb_T})
}
\xrightarrow{\mathbb P_\theta}1.
\end{equation*}
Thus, their distinction does not lie in their first-order width. 
Nevertheless the two confidence intervals are fundamentally different, while the CLT-based CI is symmetric around the plug-in value $J(\widehat \theta_T)$, the MDP-based CI is symmetric around
$J(\widehat\theta_T)
+ r_T / \widehat\theta_T^{\,2}$.
More importantly, under the assumptions established in Section
\ref{sec:math:framework}, Theorems~\ref{thm:minimal:interval} and
\ref{th:prop-J} show that the MDP-based interval satisfies exponential
accuracy and the optimality properties
\ref{eq:item:1:conf:interval}--\ref{eq:item:3:conf:UMA}.
By contrast, the ordinary CLT justifies
\eqref{eq:CLT:CI} only for a fixed confidence level. The substitution
$\alpha_T=e^{-rb_T}$ is a moving-tail heuristic and does not, by itself,
establish exponential accuracy or the remaining optimality guarantees.
The optimal MDP-based CI MDP-based CI $\mathcal{I}^\star_{T,r}(\widehat\theta_T)$ is  graphically visualized in Figure~\ref{sfig:OU:new}.

\begin{figure}[t] 
\centering
{\begin{tikzpicture}
\begin{axis}[
  width=0.5\linewidth,
  height=0.3\linewidth,
  scale only axis,
  xmode=log,
  log basis x=10,
  xmin=100,
  xmax=100000,
  xtick={100,1000,10000,100000},
  xticklabels={$10^2$,$10^3$,$10^4$,$10^5$},
  minor x tick num=8,
  ymin=1.29288128139,
  ymax=4.33303672764,
  xlabel={$T$},
  ylabel={confidence bounds},
  tick align=outside,
  grid=major,
  grid style={draw=black!10},
  axis line style={black!65},
  legend style={
    at={(0.5,-0.23)},
    anchor=north,
    legend columns=3,
    draw=black!35,
    fill=white,
    font=\small,
    /tikz/every even column/.append style={column sep=7pt}
  }
]

\addplot[area legend,draw=none,fill=red,fill opacity=0.18,forget plot]
table[row sep=crcr] {
x	y\\
100	1.75256932766\\
158.489319246	1.90719727159\\
251.188643151	2.03927238935\\
398.107170553	2.18392639935\\
630.95734448	2.25944411481\\
1000	2.34263866725\\
1584.89319246	2.38773968537\\
2511.88643151	2.42303815516\\
3981.07170553	2.45234051307\\
6309.5734448	2.4808433616\\
10000	2.49114892681\\
15848.9319246	2.49483289726\\
25118.8643151	2.49869465202\\
39810.7170553	2.50761061063\\
63095.734448	2.51321813572\\
100000	2.51464590687\\
100000	2.58118542655\\
63095.734448	2.59951797285\\
39810.7170553	2.61559947165\\
25118.8643151	2.64006509783\\
15848.9319246	2.66254669114\\
10000	2.69677787498\\
6309.5734448	2.73181143895\\
3981.07170553	2.79829084158\\
2511.88643151	2.8486706674\\
1584.89319246	2.94166018974\\
1000	3.0334735823\\
630.95734448	3.15609814634\\
398.107170553	3.32008751184\\
251.188643151	3.5776990187\\
158.489319246	3.81014266678\\
100	4.19484784372\\
}--cycle;

\addplot[color=red,dashed,line width=1.5pt]
table[row sep=crcr] {
x	y\\
100	2.82467164174\\
158.489319246	2.79685548919\\
251.188643151	2.76719073107\\
398.107170553	2.73250981292\\
630.95734448	2.7021108734\\
1000	2.6818329881\\
1584.89319246	2.65545548839\\
2511.88643151	2.63574719572\\
3981.07170553	2.62152454095\\
6309.5734448	2.60483750222\\
10000	2.59476723336\\
15848.9319246	2.58097183537\\
25118.8643151	2.57113169622\\
39810.7170553	2.56263742186\\
63095.734448	2.55577926078\\
100000	2.54905703898\\
};
\addlegendentry{$J(\widehat\theta_T) \!+\! \kappa^{+}_T$}

\addplot[area legend,draw=none,fill=blue,fill opacity=0.18,forget plot]
table[row sep=crcr] {
x	y\\
100	1.43107016531\\
158.489319246	1.5816072461\\
251.188643151	1.71804633983\\
398.107170553	1.86658097897\\
630.95734448	1.96148915759\\
1000	2.06231555766\\
1584.89319246	2.13075130778\\
2511.88643151	2.18919455235\\
3981.07170553	2.24058982913\\
6309.5734448	2.28940027646\\
10000	2.3200835672\\
15848.9319246	2.34269343945\\
25118.8643151	2.36346743548\\
39810.7170553	2.38713069373\\
63095.734448	2.40614660595\\
100000	2.41977191962\\
100000	2.48255789424\\
63095.734448	2.48694605357\\
39810.7170553	2.48735302581\\
25118.8643151	2.49336845787\\
15848.9319246	2.49506694254\\
10000	2.50450607285\\
6309.5734448	2.51123405753\\
3981.07170553	2.54134033819\\
2511.88643151	2.55241852846\\
1584.89319246	2.59339574493\\
1000	2.62557716464\\
630.95734448	2.67310808518\\
398.107170553	2.74038565874\\
251.188643151	2.8596587942\\
158.489319246	2.9385790693\\
100	3.090824143\\
}--cycle;

\addplot[color=blue,dotted,line width=1.5pt]
table[row sep=crcr] {
x	y\\
100	2.1712388927\\
158.489319246	2.22287705942\\
251.188643151	2.26305444752\\
398.107170553	2.2906579458\\
630.95734448	2.31405883515\\
1000	2.33956275892\\
1584.89319246	2.35478781513\\
2511.88643151	2.37092795883\\
3981.07170553	2.38780962415\\
6309.5734448	2.39905529659\\
10000	2.41305148107\\
15848.9319246	2.42097624116\\
25118.8643151	2.43004359647\\
39810.7170553	2.43820774583\\
63095.734448	2.44599955147\\
100000	2.45224404317\\
};
\addlegendentry{$J(\widehat\theta_T) \!+\!   \kappa^{-}_T$}

\addplot[color=darkgreen,solid,line width=1.1pt]
coordinates {(100,2.5) (100000,2.5)};
\addlegendentry{$J(\theta)$}

\end{axis}
\end{tikzpicture}%
}
\caption[]{MDP-based confidence bounds \eqref{eq:optimal:CI:OU} for the stationary variance of the
Ornstein--Uhlenbeck process. Shaded regions show the pointwise $10\%$ and
$90\%$ quantiles, while the dashed and dotted curves show the corresponding
empirical means, all based on $10^3$ independent trajectories. We set
$b_T=T^{5/11}$, $\theta=1/5$, and $r=10^{-2}$.}
\label{sfig:OU:new}
\end{figure}

\end{example}

\section{Mathematical framework, problem description and main results}\label{sec:math:framework}
Let $X$ be an $\{\mathcal{F}_t\}$-adapted stochastic process on a filtered probability space $(\Omega, \mathcal{F}, \{\mathcal{F}_t\},  \mathbb{P})$ with state space $\statespace \subset \mathbb{R}^n$. 
We denote the $\sigma$-algebra generated by the process $X_{[0,T]}$ by $\mathcal{F}_T\subset \mathcal{F}$. We assume that the probability measure $\mathbb{P}$ belongs to a parametrized family $\mathcal{P}=\{\mathbb{P}_\theta \ : \ \theta\in\Theta\}$. As mentioned in the introduction our optimal interval estimator of the cost function $J(\theta)$ will be based upon a given estimator $\{\widehat \theta_T\}$. The assumptions on the parameter space $\Theta$ and the estimator $\{\widehat \theta_T\}$ will be mentioned shortly. 

An {\em estimator} $\widehat\theta = \{\widehat\theta_T\}$ is simply a stochastic process with state space $\esttvs$ adapted to the filtration $\{\mathcal {F}_T\}$. Recall that $\widehat\theta$  is a {\em weakly consistent} estimator for $\theta_0\in\Theta$ if $\widehat \theta_T \stackrel{\PP_{\theta_0}} \Rt \theta_0$ as $T \rt \infty$ and is {\em strongly consistent} if the convergence is $\mathbb{P}_{\theta_0}$-a.s.
In many situations an estimator $\widehat \theta=\{\widehat\theta_T\}$ might not take values in the parameter space $\Theta$ at least for all $T>0$, even when it is consistent. With this in mind we introduce the following assumption.

\begin{assumption}\label{assum:para}
There exists a normed vector space $\esttvs$  equipped with a norm $\met{\cdot}$ such that
\begin{enumerate}[label = (\roman*),ref=(\roman*)]
\item  \label{ass:item:parsp-0}  $\Theta  \subset \esttvs$; %
\item\label{item:conti:J}{\grn the cost function $J:\esttvs \rt \R$ is such that for an open set $\Theta \subset \opintm \subset \esttvs$, the restriction $J \big|_\opintm: \opintm  \rt \R$ is locally Lipschitz continuous in the sense it is Lipschitz continuous on every bounded set of $\opintm$.}
\end{enumerate}
\end{assumption}

{\grn
\begin{remark} In general, the estimator $\widehat\theta_T$ takes values in $\esttvs$, and it is assumed to be a vector space as it plays the role of the ambient space in which the deviations $\widehat\theta_T-\theta$ take values. The cost function $J$ is typically defined only on the parameter space $\Theta$. In many cases, $J$ can be extended to all of $\esttvs$ while preserving the required local Lipschitz property. For example, when $J:\Theta\to\mathbb{R}$ is Lipschitz continuous, the Whitney--McShane extension theorem \cite{McSh34} guarantees a Lipschitz continuous extension to $\esttvs$. If $J$ is only locally Lipschitz continuous, analogous extension results hold under additional mild conditions; see, for example, \cite{Gut25}. In most stochastic models, however, the required extension can be verified directly from the structure of the cost function without appealing to such general extension theorems.

Assumption \ref{assum:para}-\ref{item:conti:J}, however, requires considerably less. We only require that $J:\Theta\to\mathbb{R}$ admit an extension to some open set
$\Theta\subset\opintm\subset\esttvs$ on which the required local Lipschitz property is preserved. Once such an extension to $\opintm$ is available, $J$ may be defined arbitrarily on $\esttvs\setminus\opintm$. The extension beyond $\Theta$ is needed  for technical reasons. Of course, when $\Theta$ is itself open and $J$ has the required local Lipschitz property on $\Theta$, we may simply take $\opintm=\Theta$.

    \end{remark}
    }
The topology on $\esttvs$ is the one induced by the norm $\met{\cdot}$. Note that we are not assuming that $\met{\cdot}$ is complete, that is, $\esttvs$ might not be Banach. 
This section will formalize the main problem of interest by rigorously defining the notions of exponential accuracy, minimality and other optimality criteria described in \ref{eq:item:1:conf:interval}-\ref{eq:item:3:conf:UMA} in the introduction. This in turn requires a few definitions. 

\begin{definition}[Eventually smaller]\label{def:smaller}
Let $\estsp$ be a metric space and $\SC{M}(\estsp, \R)$ the class of measurable functions from $\estsp$ to $\mathbb{R}$. A family of functions
$\{\phi_T: T>0\} \subset \SC{M}(\estsp, \R)$ is called
\begin{enumerate}[label=(\roman*)]
\item
{\em eventually smaller} than another family $\{\psi_T: T>0\} \subset \SC{M}(\estsp, \R)$ {\em at $ \metvar \in \estsp$} if
\begin{align*}
\limsup_{T\rt \infty}(\phi_T(\metvar) - \psi_T(\metvar)) \leq 0;
\end{align*}

\item {\em eventually smaller}  than $\{\psi_T: T>0\} \subset \SC{M}(\estsp, \R)$ if the former is eventually smaller than the latter at all $\metvar \in \estsp$;\\

\item {\em eventually and uniformly smaller}  than $\{\psi_T: T>0\}$ on {\em compact sets} if for any compact subset $\estsp_0 \subset \estsp$,
\begin{align*}
\limsup_{T\rt \infty} \sup_{\metvar \in \estsp_0}(\phi_T(\metvar) - \psi_T(\metvar)) \leq 0.
\end{align*}
\end{enumerate}
\end{definition}
The notions of {\em eventually larger} and {\em eventually and uniformly larger} are, of course, defined analogously.

Let $\funcls_0 \equiv \funcls_0(\esttvs, \R)$ and $\funcls_1 \equiv \funcls_1(\esttvs, \R)$  be two collections of function-families from $\esttvs$ to $\R$. Suppose we want to construct a CI for $J(\theta)$ by choosing the the lower and upper-bound functions from $\funcls_0$ and $\funcls_1$, respectively; i.e., our CI will be of the form $[\ell_T(\widehat\theta_T), u_T(\widehat\theta_T)]$, with $\{l_T\} \in \funcls_0, \ \{u_T\} \in \funcls_1.$ We now discuss the the concept of {\em minimal interval-family} that is a natural optimality criterion for choosing the `right' function-families $\{l_T\}$ and $\{u_T\}$ from the collections $\funcls_0$ and $\funcls_1$. To this end,
denote by 
\begin{align*}
\mathbb{I}(\funcls_0, \funcls_1) \equiv &\ \Big\{\{\mathcal{I}_T = [\ell_T, u_T] : T>0\} : \{\ell_T: T>0\} \in \funcls_0,\  \{u_T: T>0\}\in  \funcls_1 \Big\},
\end{align*}
and let $\mathbb{I}_0(\funcls_0, \funcls_1) \subset \mathbb{I}(\funcls_0, \funcls_1)$ be a subcollection of interval-families.
{\grn
\begin{definition}[Minimal interval-family]\label{def:minimal:interval}

Let $\lf\{\mathcal{I}^\star_T = [\ell^\star_T, u^\star_T]  : T>0\ri\}\in \mathbb{I}_0(\funcls_0, \funcls_1)$

\begin{enumerate}[label=(\roman*)]

\item Then  $\lf\{\mathcal{I}^\star_T\big|_{\Theta}\ri\}$, the interval-family restricted to $\Theta$, is a {\em pointwise minimal interval-family} of  the subcollection $\mathbb{I}_0(\funcls_0, \funcls_1)\big|_{\Theta}$ for a given {\em inflation factor} $\{\alpha_T\}$ if the following holds: for any other interval-family $\lf\{\mathcal{I}_T = [ \ell_T, u_T] : T>0\ri\} \in \mathbb{I}_0(\funcls_0, \funcls_1)$ satisfying  $\{\alpha_T(\ell_T - J): T>0\} \in \funcls_0$ and $\{\alpha_T(u_T - J): T>0\} \in \funcls_1$, the family $\lf\{\alpha_T\ell^\star_T\big|_{\Theta} : T>0\ri\}$ is eventually larger than $\lf\{\alpha_T\ell_T\big|_{\Theta} : T>0\ri\}$ and $\lf\{\alpha_T u^\star_T\big|_{\Theta} : T>0\ri\}$ is eventually smaller than $\lf\{\alpha_T u_T\big|_{\Theta} : T>0\ri\}.$

\item $\lf\{\mathcal{I}^\star_T\big|_{\Theta}\ri\}$ will be called  a {\em locally uniform minimal interval-family} of the subcollection $\mathbb{I}_0(\funcls_0, \funcls_1)\big|_{\Theta}$  for a given {\em inflation factor} $\{\alpha_T\}$ if the following holds: for any other interval-family $\lf\{\mathcal{I}_T = [ \ell_T, u_T] : T>0\ri\} \in \mathbb{I}_0(\funcls_0, \funcls_1)$ satisfying  $\{\alpha_T(\ell_T - J): T>0\} \in \funcls_0$ and $\{\alpha_T(u_T - J): T>0\} \in \funcls_1$, the family $\lf\{\alpha_T \ell^\star_T\big|_{\Theta} : T>0\ri\}$ is uniformly and eventually larger than $\lf\{\alpha_T\ell_T\big|_{\Theta} : T>0\ri\}$ on compact sets of $\Theta$ and $\lf\{\alpha_T u^\star_T\big|_{\Theta} : T>0\ri\}$ is uniformly and eventually smaller than $\lf\{\alpha_T u_T\big|_{\Theta} : T>0\ri\}$ on compact sets of $\Theta$.

\end{enumerate}
\end{definition}
}

If $\funcls_0 = \funcls_1$, we will use the symbol $\mathbb{I}(\funcls_0)$ (resp. $\mathbb{I}_0(\funcls_0)$ for a subcollection) instead of $\mathbb{I}(\funcls_0, \funcls_0)$ (resp. $\mathbb{I}_0(\funcls_0, \funcls_0)$) for simplicity.

The collections $\funcls_0$ and $\funcls_1$ relevant for this paper consist of function-families that are equi-semicontinuous when restricted to $\opintm$ at least in the long run. We thus introduce the following definition of {\em eventual equicontinuity}, which is a limiting version of the classical equicontinuity. Slight variants of this notion have been used under different names, e.g., \textit{asymptotic equicontinuity} \cite[Section~5]{rockafellar1998variational}.

\begin{definition}[Eventually equicontinuous]\label{def:equiusc}
\begin{enumerate}[label=(\roman*)] \ 
\item
A family of functions $\{\phi_T: \estsp \rt \R, T>0\}$ is called {\em eventually equi-upper semicontinuous} (ev-equi-u.s.c.) at $\metvar_0$, if  $\limsup \limits_{T \rt \infty, \metvar \rt \metvar_0}\lf(\phi_T(\metvar) - \phi_T(\metvar_0)\ri) \leq 0$.  Equivalently, for every $\vep>0$, there exist $\delta =\delta(\metvar_0,\vep)>0$ and $T_0 = T_0(\metvar_0,\vep)>0$ such that for all $\metvar \in B(\metvar_0,\delta)$ and $T\geq T_0$, $\phi_T(\metvar) - \phi_T(\metvar_0)<\vep.$ \\
The family is called {\em ev-equi-u.s.c.} if it is ev-equi-u.s.c. at all $\metvar \in \estsp$.\\

\item A family of functions $\{\phi_T: \estsp \rt \R, T>0\}$ is called \em{eventually equicontinuous} at $\metvar_0$, if  $\limsup \limits_{T \rt \infty, \metvar \rt \metvar_0} \lf|\phi_T(\metvar) - \phi_T(\metvar_0)\ri| = 0$. Equivalently, for every $\vep>0$, there exist $\delta =\delta(\metvar_0,\vep)>0$ and $T_0 = T_0(\metvar_0,\vep)>0$ such that for all $\metvar \in B(\metvar_0,\delta)$ and $T\geq T_0$, $\lf|\phi_T(\metvar) - \phi_T(\metvar_0)\ri|<\vep.$\\
The family is called {\em eventually equicontinuous} if it is eventually equicontinuous at all $\metvar \in \estsp$.

\item A family $\{\phi_T: \estsp \rt \R, T>0\}$ is {\em eventually equi-lower semicontinuous} (ev-equi-l.s.c.) if $\{-\phi_T: T>0\}$ is ev-equi-u.s.c.
\end{enumerate}
\end{definition}

{\grn Let $\SC{L}_\Theta,$  $\SC{U}_{\Theta}$ and $\equicls_{\Theta}$  respectively denote collections of function-families from $\esttvs \rt \R$, which  are respectively {\em eventually equi-l.s.c.},  {\em eventually equi-u.s.c.} and {\em eventually equicontinuous}  at all points $\theta \in \Theta \subset \opintm$. }

\begin{remark}
Notice that by openness of $\opintm$, if $\{u_T: \esttvs \rt \R\} \in \SC{U}_{\Theta}$ and $\theta \in \Theta$, then for every $\vep>0$, there exist $T_0 = T_0(\theta,\vep)>0$  and a small enough $\delta =\delta(\theta,\vep)>0$ and such that for all $\theta' \in B(\theta,\delta) \subset \opintm \subset \esttvs$ and $T\geq T_0$, $u_T(\theta') - u_T(\theta)<\vep.$ 

\end{remark}

The collection of interval families that is of primary interest in this paper is 
\begin{align*}
\mathbb{I}\lf(\equicls_{\Theta}\ri) \equiv &\ \Big\{\{\mathcal{I}_T = [\ell_T, u_T] : T>0\} : \{\ell_T: T>0\}, \{u_T: T>0\} \in \equicls_{\Theta}\Big\}.
\end{align*}
Notice that $\mathbb{I}\lf(\equicls_{\Theta}\ri) \subset \mathbb{I}\lf(\SC{L}_{\Theta},\ \SC{U}_{\Theta}\ri)$, the collection of interval families whose lower and upper end points are function-families which are respectively ev-equi-l.s.c. and ev-equi-u.s.c. on $\Theta$.

\vspace{3mm}

As mentioned in the introduction we search for optimal CI among the class of intervals which are exponentially accurate in the sense that the associated error probability decays exponentially with a certain rate and speed, see \ref{eq:item:1:conf:interval}. We first formalize this notion.

\begin{definition}[Exponentially accurate] \label{def:exponentially:accurate}
Let $r>0$ and $b_T \stackrel{T \rt \infty} \Rt \infty$.  An interval-family $\{\mathcal{I}_{T,r} : T>0\}\in \mathbb{I}\lf(\equicls_{\Theta}\ri)$ is called {\em exponentially accurate} with error rate $r$ and speed $b_T$, if 
\begin{align} \label{eq:def:UMA:cond}
\limsup\limits_{T\to\infty}\frac{1}{b_T} \log \mathbb{P}_\theta\lf( J(\theta) \notin \mathcal{I}_{T,r}(\widehat \theta_T)\ri) \leq -r, \quad \forall \theta\in\Theta \subset \esttvs.
\end{align}
\end{definition}
Denote by $\mathbb{I}_r\lf(\equicls_{\Theta}\ri) \subset \mathbb{I}\lf(\equicls_{\Theta}\ri)$  the subcollection of interval-families $\{\mathcal{I}_{T,r} \equiv [\ubar H_{T,r}, \bar H_{T,r}]:T>0\}$ with both $\{\ubar H_{T,r}: T>0\}, \{\bar H_{T,r}: T>0\} \in \equicls_{\Theta}$  which are exponentially accurate with error rate $r$ and speed $b_T$ (that is, which satisfy \eqref{eq:def:UMA:cond}). Similarly, $\mathbb{I}_r\lf(\SC{L}_{\Theta}, \SC{U}_{\Theta}\ri)$ will denote the subcollection of interval-families from $\mathbb{I}\lf(\SC{L}_{\Theta}, \SC{U}_{\Theta}\ri)$ that are exponentially accurate with error rate $r$ and speed $b_T$. \\

A common criterion to characterize the ``optimal" confidence interval among those with fixed error-threshold is the notion of \textit{uniformly most accurate} (UMA) confidence intervals. We consider an appropriately adjusted asymptotic version of this concept for our problem.

\begin{definition}[Eventually UMA] \label{eq:def:UMA}
    Fix an exponential rate parameter $r>0$ and a speed of decay $\{b_T : T>0\}$. An interval family $\{\widetilde{\mathcal{I}}_{T,r} : T>0\}\in \mathbb{I}_r\lf(\equicls_{\Theta}\ri)$ is called {\em eventually UMA} in the collection $\mathbb{I}_r\lf(\equicls_{\Theta}\ri)$, if for any other interval family $\{\mathcal{I}_{T,r} : T>0\}\in \mathbb{I}_r\lf(\equicls_{\Theta}\ri)$,
    \begin{equation*}
\limsup_{T\to\infty} \lf[ \mathbb{P}_\theta\lf( J(\tilde \theta) \in \widetilde{\mathcal{I}}_{T,r}(\widehat\theta_T)\ri) -  \mathbb{P}_\theta\lf( J(\tilde\theta) \in \mathcal{I}_{T,r}(\widehat\theta_T)\ri)\ri] \leq 0 \quad \forall \theta, \tilde\theta\in\Theta : J(\tilde\theta)\neq J(\theta).
    \end{equation*}
\end{definition}
Intuitively, in the long run confidence intervals with small probabilities of covering the wrong function value $J(\tilde\theta)$ are preferred. 
Finally, we can precisely state the key problem tackled in this paper.

\begin{problem}[Optimal Interval Estimation] \label{prob:interval}  Fix an exponential rate parameter $r>0$ and a speed of decay $\{b_T:T>0\}$  such that $b_T\to\infty$ as $T\to\infty$. Let $a_T  = \sqrt {T/b_T}$. Find an interval-family $\{\mathcal{I}^\star_{T,r} \equiv [\ubar{J}_{T,r}, \bar{J}_{T,r}]: T>0\}$ in $\mathbb{I}_r\lf(\equicls_{\Theta}\ri)$ such that the following hold.
\begin{enumerate}[label=(\alph*)]

\item \label{prob:item:minimality} {\grn For any $\alpha_T = o(a_T)$, $\{\mathcal{I}^\star_{T,r}\big|_{\Theta}: T>0\}$ is a {\em locally uniform minimal interval-family} of $\mathbb{I}_r\lf(\equicls_{\Theta}\ri)\big|_{\Theta}$ for a given inflation factor $\{\alpha_T\}$ in the sense of Definition \ref{def:minimal:interval}.}

\item \label{prob:item:cons} {\grn The length of the CI $\{\mathcal{I}^\star_{T,r}(\widehat \theta_T): T>0\}$ is $O(1/a_T)$; specifically, the families $\{a_T(\bar J_{T,r}(\widehat \theta_T) - J(\theta_0))\}$ and $\{a_T(\ubar J_{T,r}(\widehat \theta_T) - J(\theta_0))\}$ are stochastically bounded (c.f. Definition \ref{def:stoch-bdd}). }

In particular, for any $\alpha_T = o(a_T)$,  as $T \rt \infty$, 
$$\alpha_T\lf(\ubar J_{T,r}^\delta(\widehat \theta_T) - J(\theta_0)\ri), \ \alpha_T\lf(\bar J_{T,r}^\delta(\widehat \theta_T) - J(\theta_0)\ri)  \stackrel{\PP_{\theta_0}}\Rt 0.$$

\item \label{prob:item:UMA} $\{\mathcal{I}^\star_{T,r}: T>0\}$ is {\em eventually UMA} in the collection $\mathbb{I}_r\lf(\equicls_{\Theta}\ri)$.

\item \label{prob:item:unbiased} $\{\mathcal{I}^\star_{T,r}: T>0\}$ is such that
\begin{equation*}
    \lim_{T\to\infty} \frac{1}{b_T} \log \mathbb{P}_{\theta_0}\lf( J(\tilde \theta_0)\notin \mathcal{I}^\star_{T,r}(\widehat \theta_T) \ri) = 0 \quad \forall \theta_0, \tilde \theta_0\in\Theta : J(\tilde \theta_0)\neq J(\theta_0).
\end{equation*}

\end{enumerate}
\end{problem}

Problem~\ref{prob:interval} is the rigorous formulation of the interval criteria \ref{eq:item:1:conf:interval}-\ref{eq:item:3:conf:UMA} mentioned in the introduction.
Property \ref{prob:item:unbiased} of Problem~\ref{prob:interval}, which is the mathematical formulation of the criteria \ref{eq:item:4:mis:chracterization}, states that the probability of the ``wrong value" of the cost function $J(\theta')$ falling outside our confidence interval does not decay at an exponential rate. This is of course a natural criterion for optimality - we want the CI $\mathcal{I}^\star_{T,r}(\widehat \theta_T)$ to be such that the chances of true value $J(\theta)$ lying within it is exponentially high, while for any other value of the cost function there is a significant possibility that it will lie outside our CI. 

In the traditional world of confidence intervals with fixed threshold $\alpha$, the corresponding idea is that the probability of the CI covering the wrong function values $J(\theta')$ does not exceed the given confidence $(1-\alpha)$. Such an interval is typically referred to as an \textit{unbiased confidence interval} in the literature, for example, see \cite[Section~5.5]{ref:lehmann-testing}.

\vspace{3mm}

Since exponential decay of error probabilities is involved, it is clear that a large deviation analysis of the estimator $\widehat \theta_T$ is required in an appropriate scaling regime. The scaling regime considered in the paper is given by $1 \ll b_T \ll T$. As mentioned in the introduction, large deviation asymptotics in this scaling regime is usually referred to as MDP (see Definition \ref{def:MDP}). We now state the assumptions on the estimator $\{\widehat \theta_T\}$. %

\begin{assumption} \label{assum:basic}
The following hold.
\begin{enumerate}[label=(\roman*)] 
\item \label{ass:item:MDP} Let $1\ll b_T \ll T$. For each $\theta_0 \in \Theta$, $\{\widehat \theta_T\}$ is weakly consistent for $\theta_0$ with respect to $\PP_{\theta_0}$ and satisfies an MDP in $\esttvs$ with speed $b_T$ and good rate function $I^M_{\theta_0}: \esttvs \rt [0,\infty)$ (cf. Definitions~\ref{def:LDP} and \ref{def:MDP}). \vs{.1cm}

\item \label{item:4:ass:jointly:good:alternative} 
The mapping $(\theta, \vart) \in \Theta \times \esttvs \mapsto I^M_\theta(\vart)$ satisfies the following two conditions:
\begin{enumerate}[label = (\alph*),ref=\theenumi{}-(\alph*)]
\item \label{item:ass:cpt:RF} If $\{\theta_n\} \subset \Theta$ and $\{\vart_n\} \subset \esttvs$ are such that $\sup_{n}\met{\theta_n} < \infty$ and $\met{\vart_n} \rt \infty$ as $n \rt \infty$, then $I^M_{\theta_n}(\vart_n) \rt \infty.$

\item \label{item:ass:growth:RF} If $\{\theta_n\}\subset \Theta$ and $\{\vart_n\}\subset \esttvs$ are such that $\met{\theta_n} \rt \infty$ and $\met{\vart_n}/\met{\theta_n} \rt \infty$ as $n\rt \infty$, then $\limsup \limits_{n\rt \infty}I^M_{\theta_n}\lf(\vart_n\ri) = \infty$.

\end{enumerate}

\vs{.2cm}

\end{enumerate}
\end{assumption}

\begin{remark}\label{rem:ass:basic} Regarding Assumption~\ref{assum:basic}, the following two remarks are in order.
\begin{itemize}
    \item Condition \ref{item:ass:growth:RF} will be assumed to be vacuously true if a sequence  $\{\theta_n\}$ converging to infinity does not exist, that is, when $\Theta$ is a bounded set.
    \item Condition~\ref{item:ass:cpt:RF} of Assumption~\ref{assum:basic} is equivalent to the condition that for every bounded set $\Theta_0 \subset \Theta$ and $m_0 >0$, the set $\mathbb{G}_{m_0}(\Theta_0) \equiv \{(\theta,\vart)\in\Theta\times\esttvs : \theta \in \Theta_0, I^M_\theta(\vart) \leq m_0\}$ is bounded. A  justification is provided in Appendix~\ref{appB}.
\end{itemize}
\end{remark}

{\grn
One consequence of $\widehat \theta_T$ satisfying an MDP is that the estimation error converges to zero at a rate faster than $1/a_T$, which is a substantially stronger conclusion than weak consistency alone. This is the content of the following lemma. Although the proof is intuitively clear, it involves a few technicalities that need to be addressed; the details are  provided in the appendix.

\begin{lemma}
    \label{lem:est-conv-rate}
Suppose that Assumption \ref{assum:basic}-\ref{ass:item:MDP} holds. Fix any $\theta_0 \in \Theta$, and  assume that $I^M_{\theta_0}(\vart)>0,$ \  for $\vart \neq 0.$ Then  as $T \rt \infty,$\ 
\begin{align}
    \label{eq:est-conv-prob}
    a_T(\widehat\theta_T - \theta_0) \stackrel{\PP_{\theta_0}} \Rt 0.
\end{align}

\end{lemma}

}

{\grn
\subsection{Construction of optimal interval-family}
The solution to Problem \ref{prob:interval} is given by an optimization problem based on the MDP rate function $I^M_\theta$, which we motivate and describe below.

To construct the interval-family
$
\{[\ubar{J}_{T,r}(\cdot),\bar{J}_{T,r}(\cdot)]: T>0\}
$
underlying an exponentially accurate CI
$
[\ubar{J}_{T,r}(\widehat\theta_T),\bar{J}_{T,r}(\widehat\theta_T)],
$
a natural approach is to define the endpoint functions at $\theta' \in \esttvs$ as
\begin{align}\label{eq:end-fun-pre}
 \ubar {J}_{T,r}(\theta')
 &= \inf\lf\{J(\theta): \theta \in  \scr{I}_{T,r}(\theta')\ri\}, \quad
 \bar {J}_{T,r}(\theta')
 = \sup\lf\{J(\theta): \theta \in  \scr{I}_{T,r}(\theta')\ri\}.
\end{align}
The question is then how to choose the set $\scr{I}_{T,r}(\theta') \subset \Theta$. To motivate its construction, first notice that choosing $\scr{I}_{T,r}(\theta')\equiv\Theta$ simply leads to the global minimum and maximum $J_{\min}$ and $J_{\max}$ of the target function $J$, which could be $\mp\infty$. This corresponds to a confidence interval with $100\%$ coverage, but one that is completely uninformative. Thus, $\scr{I}_{T,r}(\theta')$ should contain only those parameters that are sufficiently compatible with the observed value $\theta'$ of the estimator $\widehat\theta_T$, according to the deviation behavior described by the MDP.

Recall that, informally, the MDP for $\widehat\theta_T$ means that, when the true parameter is $\theta$,
\[
\PP_{\theta}\lf
(a_T(\widehat\theta_T-\theta)\approx \vart\ri)
\approx
e^{-b_T I^M_{\theta}(\vart)},
\]
where $a_T = \sqrt{T/b_T}$.
Thus, the rate function $I^M_{\theta}(\vart)$ describes the exponential decay rate associated with observing a scaled deviation $\vart$ of the estimator from its true value $\theta$. Now, if the true parameter is $\theta_0$, we want the CI
$
[\ubar{J}_{T,r}(\widehat\theta_T),\bar{J}_{T,r}(\widehat\theta_T)]
$
to have a coverage error whose exponential decay rate is at least $r$. This suggests that, for an observed value $\theta'$ of $\widehat\theta_T$, the set
$\scr{I}_{T,r}(\theta')$ should consist of those $\theta\in\Theta$ that can
reasonably be regarded as candidate values of the true parameter, in the sense
that observing $\theta'$ under $\PP_\theta$ incurs a cost, measured in terms of the rate function $I^M_\theta$, no
larger than $r$. More precisely, under the candidate true parameter $\theta$,
the observed estimator-value $\theta'$ corresponds to the scaled deviation
$a_T(\theta'-\theta)$, whose exponential decay rate is given by
$I^M_\theta(a_T(\theta'-\theta))$.  Thus, we regard $\theta \in \Theta$ as compatible with the estimator-value $\theta'$ and include $\theta$ in $\scr{I}_{T,r}(\theta')$ whenever
$
I^M_{\theta}\lf(a_T(\theta'-\theta)\ri)\leq r.
$
This leads naturally to the definition of the compatibility set as 
\begin{equation}\label{eq:feasible:set:ub-0}
    \scr{I}_{T,r}(\theta')
    \dfeq
    \lf\{\theta\in\Theta \ : \
    I^M_{\theta}\lf(a_T(\theta'-\theta)\ri) \leq r \ri\} \equiv \lf\{\theta\in\Theta \ : \  a_T(\theta'-\theta) \in \sblvl_{\theta,r} \ri\},
\end{equation}
where 
\begin{align}\label{eq:sblvl}
\sblvl_{\theta,r} \dfeq  \lf\{\vart \in \esttvs: I^M_\theta(\vart) \leq r\ri\}
\end{align}
denotes the sublevel set of $I_\theta^M$ for a level $r>0$.
To ensure that, for every such compatible $\theta$, the corresponding value
$J(\theta)$ is covered by the resulting interval
$[\ubar{J}_{T,r}(\theta'),\bar{J}_{T,r}(\theta')]$, we naturally define its
lower and upper endpoints by taking the infimum and supremum of $J$ over
$\scr{I}_{T,r}(\theta')$, as in \eqref{eq:end-fun-pre}.

Due to certain technical reasons, explained later, we need to work with a slightly larger open set containing the sublevel set $\sblvl_{\theta,r}$. For $\delta>0$, let
\begin{align}\label{eq:sblvl-fat}
\sblvl_{\theta,r}^\delta \dfeq\lf\{\vart' \in \esttvs: \met{\vart'-\sblvl_{\theta,r}} \equiv \inf_{\vart \in \sblvl_{\theta,r}} \met{\vart'-\vart} < \delta\ri\} = \bigcup_{\vart \in \sblvl_{\theta,r} } B^\circ(\vart,\delta)
\end{align} 
denote the {\em open $\delta$-fattening} of the sublevel set $\sblvl_{\theta,r}$. Clearly, the $\sblvl_{\theta,r}^\delta $ are open sets for each $\delta>0$. Next, define $ \scr{I}^\delta_{T,r}(\theta')$,
 the feasible set of the optimization problem defining the endpoint functions, as
\begin{equation}\label{eq:feasible:set:ub}
    \scr{I}^\delta_{T,r}(\theta') \dfeq \lf\{\theta\in\Theta \ : \  a_T(\theta'-\theta) \in \sblvl_{\theta,r}^\delta \ri\}.
\end{equation}
For $\delta=0$, we will take $\sblvl^0_{\theta,r} \equiv \sblvl_{\theta,r},$ and $ \scr{I}^0_{T,r}(\theta') \equiv   \scr{I}_{T,r}(\theta').$
Next, for $\delta \geq 0$, define the functions $\bar {J}^\delta_{T,r}, \ubar {J}^\delta_{T,r}: \esttvs \rt \R$ by
\begin{align}\label{eq:optJ-mod}
\begin{aligned}
\bar {J}^\delta_{T,r}(\theta') \dfeq&\
\begin{cases}
J(\theta'),  \quad \text{ if } \scr{I}^\delta_{T,r}(\theta') = \emptyset,\\
\sup\lf\{ J(\theta):  \theta \in  \scr{I}^\delta_{T,r}(\theta')\ri\} \equiv \sup_{\theta\in\Theta} \lf\{ J(\theta) : a_T(\theta'-\theta) \in \sblvl^\delta_{\theta,r}\ri\},    \ \ \text{otherwise.} 
\end{cases}
\\
\ubar {J}^\delta_{T,r}(\theta') \dfeq&\
\begin{cases}
J(\theta'),  \quad \text{ if } \scr{I}^\delta_{T,r}(\theta') = \emptyset,\\
\inf\lf\{ J(\theta):   \theta \in  \scr{I}^\delta_{T,r}(\theta')\ri\} \equiv \inf_{\theta\in\Theta} \lf\{ J(\theta): \ a_T(\theta'-\theta) \in \sblvl^\delta_{\theta,r}\ri\},  \ \  \text{otherwise.}
\end{cases}
\end{aligned}
\end{align}
Note that for any $r,\delta \geq 00$, $\scr{I}^\delta_{T,r}(\theta') \neq \emptyset$ for $\theta' \in \Theta$.
The main objective of the paper is to show that for any fixed $\delta>0$, the interval family defined via \eqref{eq:optJ-mod} as $\{\mathcal{I}^{\delta,\star}_{T,r}\dfeq [\ubar J^\delta_{T,r},\bar J^\delta_{T,r}]:T>0\}$
is a solution to  Problem \ref{prob:interval}. Obviously a small $\delta$ is preferred in practice as it leads to a smaller interval for each fixed $T$, even though in the limit  such differences disappear. The ideal choice of $\delta$ is of course $0$, and we show that $\delta = 0$ can be taken under the additional assumption that for each fixed $\theta \in\Theta$, $I^M_\theta: \esttvs \rt [0,\infty)$ is continuous.  In other words, when $I^M_\theta$ is continuous, a solution to  Problem \ref{prob:interval} is given by $\{\mathcal{I}^{0,\star}_{T,r}\dfeq [\ubar J^0_{T,r},\bar J^0_{T,r}]:T>0\}$
where \eqref{eq:optJ-mod} for $\delta=0$ is the same as
\begin{align}\label{eq:optJ}
\begin{aligned}
\bar {J}^0_{T,r}(\theta') \equiv \bar {J}_{T,r}(\theta') =&\ 
\begin{cases}
J(\theta'),  \quad \text{ if } \scr{I}^\delta_{T,r}(\theta') = \emptyset\\
\sup_{\theta\in\Theta} \lf\{ J(\theta)\ : \ I^M_\theta\lf(a_T(\theta'-\theta)\ri)\leq r\ri\}, \quad \text{otherwise.}
\end{cases}
\\
\ubar {J}^0_{T,r}(\theta') \equiv \ubar {J}_{T,r}(\theta') =&\ 
\begin{cases}
J(\theta'),  \quad \text{ if } \scr{I}^\delta_{T,r}(\theta') = \emptyset\\
\inf_{\theta\in\Theta} \lf\{ J(\theta)\ : \ I^M_\theta\lf(a_T(\theta'-\theta)\ri)\leq r\ri\}, \quad \text{otherwise.}
\end{cases}
\end{aligned}
\end{align}
In fact the continuity of the rate function $I^M_\theta$ is only needed to establish exponential accuracy of the interval $\mathcal{I}^{0,\star}_{T,r}$.

The main results of the paper are presented in Theorem~\ref{thm:minimal:interval} and Theorem \ref{th:prop-J}. Theorem \ref{thm:minimal:interval}-\ref{item:opt:interval:exp-accu}, which establishes exponential accuracy of our interval, is  similar in spirit to  \cite[Theorem~3.1]{ref:Sutter-19} which looked at the optimal CI problem in the usual LDP scaling regime ($b_T=T$).
Not all parts of Assumptions \ref{assum:para} and \ref{assum:basic}
are needed for every statement. We will identify the specific assumptions that are needed for a particular assertion.
}

\begin{theorem}[Minimal interval]\label{thm:minimal:interval}
Suppose that Assumption~\ref{assum:para} and Assumption~\ref{assum:basic} hold. Let  
$\ubar J^\delta_{T,r}$ and $\bar J^\delta_{T,r}$ be defined by \eqref{eq:optJ-mod}. 
Then the following statements hold. 
\begin{enumerate}[label=(\roman*)] 
    \item \label{item:opt:interval:exp-accu}  For any $\delta>0$ and $\theta_0 \in \Theta$
\begin{equation}\label{eq:feas-1}
\begin{aligned}
    \limsup\limits_{T\to\infty}\frac{1}{b_T} \log \mathbb{P}_{\theta_0}\lf( J(\theta_0)>\bar J^\delta_{T,r}(\widehat \theta_T)\ri) \leq &\ -r, \\
    \limsup\limits_{T\to\infty}\frac{1}{b_T} \log \mathbb{P}_{\theta_0}\lf( J(\theta_0)< \ubar J^\delta_{T,r}(\widehat \theta_T)\ri) \leq &\  -r.
    \end{aligned}
\end{equation}
    In particular, the interval-family $\{\mathcal{I}^{\delta,\star}_{T,r}=[\ubar J^\delta_{T,r},\bar J^\delta_{T,r}]:T>0\}$ is exponentially accurate with error rate $r$ and speed $b_T$. %
    \item \label{item:thm:uniformly:compact} {\grn For any bounded $\opintm_0 \subset \opintm$, there exist constants $T_0 \equiv T_0(\opintm_0)$ and  $\const_0 \equiv \const_0(\opintm_0)$ such that for all $T \geq T_0$
    $$\sup_{\theta' \in \opintm_0 }\lf|\ubar{J}^\delta_{T,r}(\theta')-J(\theta')\ri| \vee \sup_{\theta' \in \opintm_0}\lf|\bar{J}^\delta_{T,r}(\theta')-J(\theta')\ri| \leq \const_0/a_T.$$
    In particular, for any $\alpha_T = o(a_T)$, $\alpha_T (\ubar J^\delta_{T,r} - J)$, $\alpha_T (\bar J^\delta_{T,r} - J)$ converge to $0$ uniformly on bounded sets of $\opintm$ as $T \rt \infty$.  }
   
   \item \label{item:opt:interval:equi}  {\grn For any $\alpha_T = o(a_T)$ and $\delta > 0$, the families $\{\alpha_T(\ubar J^\delta_{T,r}-J): T>0\}$ and $\{\alpha_T(\bar J^\delta_{T,r}-J): T>0\}$ are eventually equi-continuous at any $\theta' \in \opintm$. }
    \item \label{item:opt:interval:pt-min} {\grn For any $\alpha_T = o(a_T)$ and $\delta > 0$, the interval-family $\lf\{\mathcal{I}^{\delta,\star}_{T,r}\big|_{\Theta}=\lf[\ubar {J}^\delta_{T,r}\big|_{\Theta},\bar J^\delta_{T,r}\big|_{\Theta}\ri]:T>0\ri\}$ is a pointwise minimal interval-family of \ $\mathbb{I}_r\lf(\SC{L}_{\Theta},\ \SC{U}_{\Theta}\ri)\big|_{\Theta}$ for a given {\em inflation factor} $\{\alpha_T\}$.}
    
     \item \label{item:opt:interval:sol} {\grn For any $\alpha_T = o(a_T)$ and $\delta > 0$, the interval-family $\lf\{\mathcal{I}^{\delta,\star}_{T,r}\big|_{\Theta}=\lf[\ubar J^\delta_{T,r}\big|_{\Theta},\bar J^\delta_{T,r}\big|_{\Theta}\ri]:T>0\ri\}$ is a locally uniform minimal interval-family of   $\mathbb{I}_r\lf(\equicls_{\Theta}\ri)\big|_{\Theta}$ for a given {\em inflation factor} $\{\alpha_T\}$; that is, $\lf\{\mathcal{I}^{\delta,\star}_{T,r}=[\ubar J^\delta_{T,r},\bar J^\delta_{T,r}]:T>0\ri\}$ is a solution to Problem \ref{prob:interval}-\ref{prob:item:minimality}.}
   \end{enumerate}
  If $I^M_{\theta_0}: \esttvs \rt [0,\infty]$ is continuous, then \ref{item:opt:interval:exp-accu}, \ref{item:opt:interval:pt-min} and \ref{item:opt:interval:sol} also hold for $\delta=0$. 
\end{theorem}

\begin{remark}
Taking $\alpha_T \equiv 1$ in Theorem \ref{thm:minimal:interval}-\ref{item:opt:interval:equi}, it follows that  the families $\{\ubar J^\delta_{T,r}\}$ and $\{\bar J^\delta_{T,r}\}$
   are respectively eventually  equicontinuous on $\opintm$. %
\end{remark}

Notice that Theorem \ref{thm:minimal:interval}-\ref{item:opt:interval:pt-min} indicates that the functions $\{\bar J^\delta_{T,r}\}$ and $\{\ubar J^\delta_{T,r}\}$ are still the ideal function-families for defining the upper and lower bounds for CI even in the bigger class, $\mathbb{I}_r\lf(\SC{L}_{\Theta},\ \SC{U}_{\Theta}\ri)\big|_{\Theta},$ provided the weaker notion of pointwise minimality is used as a criterion.\\ 

Our final theorem deals with asymptotic properties of the interval-family $\{\mathcal{I}^{\delta,\star}_{T,r}=[\ubar J^\delta_{T,r},\bar J^\delta_{T,r}]:T>0\}$ and shows, among other things, that it satisfies the remaining criteria in Problem \ref{prob:interval}. Theorem \ref{th:prop-J}-\ref{thm:minimal:interval:min:cons}, in particular, proves that not only are the functions, $\ubar J_{T,r}^\delta$ and $\bar J_{T,r}^\delta$, ideal for defining the lower and upper bounds for the optimal CI based on the estimator $\widehat \theta_T$, the actual (random) CI, $[\ubar J_{T,r}^\delta(\widehat \theta_T), \bar J_{T,r}^\delta(\widehat \theta_T)]$, is eventually contained in any other competing intervals if $\widehat \theta _T$ is consistent.

\begin{theorem}[Further optimality properties of $\ubar J_{T,r}^\delta$ and $\bar J_{T,r}^\delta$ ]\label{th:prop-J}
Suppose that Assumption~\ref{assum:para} and Assumption~\ref{assum:basic}  hold. 
Let $\{\mathcal{I}^{\delta,\star}_{T,r}=[\ubar J^\delta_{T,r},\bar J^\delta_{T,r}]:T>0\}$ with $\ubar J_{T,r}^\delta$ and $\bar J_{T,r}^\delta$  defined by \eqref{eq:optJ-mod}. Then for any $\delta> 0$ the following hold.
\begin{enumerate}[label=(\roman*)] 

\item \label{item:thm:consistent} {\grn Assume that $I^M_{\theta_0}(\vart)>0,$ \  for $\vart \neq 0.$ Then $\bar J^\delta_{T,r}(\widehat \theta_T), \ubar J^\delta_{T,r}(\widehat \theta_T)$ converge to $J(\theta_0)$ with speed $1/a_T$ in the sense the families $\{a_T(\bar J^\delta_{T,r}(\widehat \theta_T) - J(\theta_0))\}$ and $\{a_T(\ubar J^\delta_{T,r}(\widehat \theta_T) - J(\theta_0))\}$ are stochastically bounded under $\PP_{\theta_0}$ (c.f. Definition \ref{def:stoch-bdd}).
In other words, 
$\{\mathcal{I}^{\delta,\star}_{T,r}:T>0\}$ solves  Problem \ref{prob:interval}-\ref{prob:item:cons}. }

 \item \label{thm:minimal:interval:mischa} Let $\theta_0 \in \Theta$. Then for every $\tilde \theta_0 \in \Theta$ such that $J(\tilde \theta_0) > J(\theta_0)$
 \begin{align*}
\lim_{T \rt \infty} \frac{1}{b_T} \log \mathbb{P}_{\theta_0}\left( J(\tilde \theta_0) >  \bar{J}^{\delta}_{T,r}(\widehat\theta_T) \right) =0,
 \end{align*}
 and for every $\tilde \theta_0 \in \Theta$ such that $J(\tilde \theta_0) < J(\theta_0)$
  \begin{align*}
\lim_{T \rt \infty} \frac{1}{b_T} \log \mathbb{P}_{\theta_0}\left( J(\tilde \theta_0) <  \ubar{J}^{\delta}_{T,r}(\widehat\theta_T) \right) =0.
 \end{align*}
 In particular, $\{\mathcal{I}^{\delta,\star}_{T,r}: T>0\}$ is a solution to Problem~\ref{prob:interval}-\ref{prob:item:unbiased}.
\end{enumerate}
Suppose that  $\lf\{[\ubar H_{T,r},\bar H_{T,r}]:T>0\ri\}\in \mathbb{I}_r\lf(\equicls_{\Theta}\ri)$ (see Definition \ref{def:exponentially:accurate} and the paragraph below), and that the following hold: either
\begin{enumerate}[label = (\Alph*),ref=(\Alph*)]
\item \label{ass:item:parsp-a}  $\Theta \subset  \esttvs$ is  open, or
\item \label{ass:item:parsp-b} the estimator $\widehat \theta_T$ takes values in $\Theta$  for all $T>0$.
\end{enumerate}

  Then the following statements are true. 
\begin{enumerate}[label=(\roman*), start=3] 
\item \label{thm:minimal:interval:min:cons}  {\grn If for every $\theta_0\in\Theta$, the estimator $\widehat\theta_T$ is strongly consistent, then for any $\alpha_T = o(a_T)$
     \begin{align*}
     \limsup_{T\rt \infty} \alpha_T\lf(\bar J^\delta_{T,r}(\widehat \theta_T) - \bar H_{T,r}(\widehat \theta_T)\ri) \leq &\  0, \quad \PP_{\theta_0}-a.s.\\
          \liminf_{T\rt \infty} \alpha_T\lf(\ubar J^\delta_{T,r}(\widehat \theta_T) - \ubar H_{T,r}(\widehat \theta_T)\ri) \geq&\  0, \quad \PP_{\theta_0}-a.s.
     \end{align*}
     If for every $\theta_0 \in \Theta$, $\widehat \theta_T$ is weakly consistent then for any $\kappa>0$, 
     \begin{align*}
     \lim_{T\rt \infty}\PP_{\theta_0}\lf(\alpha_T(\bar J^\delta_{T,r}(\widehat \theta_T)-\bar H_{T,r}(\widehat \theta_T)) > \kappa\ri) =&\ 0\\
      \lim_{T\rt \infty}\PP_{\theta_0}\lf(\alpha_T(\ubar J^\delta_{T,r}(\widehat \theta_T) -\ubar H_{T,r}(\widehat \theta_T))< -\kappa\ri) =&\ 0.
     \end{align*} }

\item \label{thm:minimal:interval:uma-ci} Let $\theta_0, \tilde\theta_0 \in \Theta$ be such that $J(\theta_0) \neq J(\tilde\theta_0)$, and suppose for each $T$,  $\widehat \theta_T$ is tight (equivalently, its distribution is a Radon probability measure on $\esttvs$). Then
\begin{align*}
    &\limsup_{T\to\infty} \left( \mathbb{P}_{\theta_0}\left(J(\tilde\theta_0)\leq \bar{J}^{\delta}_{T,r}(\widehat\theta_T)\right) - \mathbb{P}_{\theta_0}\left(J(\tilde\theta_0) \leq \bar{H}_{T,r}(\widehat\theta_T)\right) \right) \leq 0,\\
     &\limsup_{T\to\infty} \left( \mathbb{P}_{\theta_0}\left(J(\tilde\theta_0)\geq \ubar{J}^{\delta}_{T,r}(\widehat\theta_T)\right) - \mathbb{P}_{\theta_0}\left(J(\tilde\theta_0) \geq \ubar{H}_{T,r}(\widehat\theta_T)\right) \right) \leq 0.
\end{align*}
In particular, $\{\mathcal{I}^{\delta,\star}_{T,r}: T>0\}$ is a solution to Problem~\ref{prob:interval}-\ref{prob:item:UMA}.     
\end{enumerate}
  If $I^M_{\theta_0}: \esttvs \rt [0,\infty]$ is continuous, then \ref{item:thm:uniformly:compact}-\ref{thm:minimal:interval:uma-ci} also hold for $\delta=0$. 
\end{theorem}

\begin{remark} 
A few remarks are in order.
\begin{enumerate}[label=(\roman*)]
\item For $\delta =0$, the continuity of $I^M_{\theta_0}$ is only needed to establish exponential accuracy of the interval-family, $\{\mathcal{I}^{0,\star}_{T,r}: T>0\}$ (Theorem \ref{thm:minimal:interval}-\ref{item:opt:interval:exp-accu}), that is, its membership in $\mathbb{I}_r\lf(\equicls_{\Theta}\ri)$, but is not needed otherwise for proofs of Theorem \ref{thm:minimal:interval}-\ref{item:opt:interval:pt-min} \& \ref{item:opt:interval:sol}  and Theorem \ref{th:prop-J}- \ref{item:thm:uniformly:compact}-\ref{thm:minimal:interval:uma-ci}.
\end{enumerate}
\begin{enumerate}
\item[(ii)] 
{\grn Notice that the tightness condition imposed on $\widehat\theta_T$ in
Theorems~\ref{th:prop-J}-\ref{thm:minimal:interval:uma-ci} is only
required for each fixed $T$; we do not require the family
$\{\widehat\theta_T:T>0\}$ to be (uniformly) tight, which is a much stronger condition.

In particular, if $\esttvs$ is Polish, then the distribution of any random variable, in particular distribution of each
$\widehat\theta_T$, is automatically tight, and hence this assumption is
superfluous. Even if $\esttvs$ is not Polish the same conclusion holds if
$\widehat\theta_T$ takes values in a subset
$\esttvs_0\subset\esttvs$ which, equipped with the induced topology, is
Polish. }
\end{enumerate}

\end{remark}

\subsection{Revisiting OU Process}\label{sec:optimal:interval}

Before we prove Theorems~\ref{thm:minimal:interval} and \ref{th:prop-J}, we derive an explicit expression for the optimal CI for asymptotic variance of the  Ornstein-Uhlenbeck process introduced in Example~\ref{ex:OU:part1}.

\begin{example}[Asymptotic variance estimation of an OU process] \label{ex:OU:part2}
It is well known~\cite[Theorem~1.2]{ref:Gao-10} that the MLE $\{\widehat \theta_T\}$ defined in \eqref{eq:MLE:OU} is such that $\{a_T(\widehat \theta_T - \theta)\}$ satisfies an LDP with speed $\{b_T\}$, where $b_T = T/a_T^2$ and rate function 
\begin{equation} 
\label{eq:OU:rate:fct}
    I_\theta^M(\vartheta) = \frac{\vartheta^2}{4\theta}.
\end{equation}
Here, the parameter space is $\Theta = \mathbb{R}_{>0}$ and the MLE is such that $\widehat \theta_T\in\esttvs=\mathbb{R}$ for all $T>0$. Therefore, Assumption~\ref{assum:para} clearly holds. Moreover, Assumption~\ref{assum:basic} also is satisfied in this setting. As the rate function~\eqref{eq:OU:rate:fct} is continuous, we consider $\delta=0$ and solution to Problem~\ref{prob:interval}, according to Theorems~\ref{thm:minimal:interval} and \ref{th:prop-J}, is then provided by
\begin{equation} \label{eq:OU:optimal:interval}
\begin{aligned}
    \bar J^0_{T,r}(\widehat\theta_T) &= \sup_{\theta>0} \{J(\theta) \ : \  I_\theta^M(a_T(\widehat\theta_T-\theta))\leq r\}
    = J(\widehat\theta_T) + \kappa^+_T   \\
     \ubar J^0_{T,r}(\widehat\theta_T) &= \inf_{\theta>0} \{J(\theta) \ : \  I_\theta^M(a_T(\widehat\theta_T-\theta))\leq r\} 
    = J(\widehat\theta_T) +  \kappa^-_T,
\end{aligned}
\end{equation}
where $r_T = r/a_T^2$ and
\begin{equation*}
    \kappa^{\pm}_T = \frac{1}{\widehat\theta_T^2}\left( r_T \pm \sqrt{r_T^2+\widehat\theta_T r_T} \right).
\end{equation*}
This leads to the confidence interval \eqref{eq:optimal:CI:OU} analyzed and compared to the CLT heuristic in Example~\ref{ex:OU:part1}.
The strong consistency of the MLE $\widehat \theta_T$  ensures that $\lim_{T\to\infty}  \kappa^+_T = \lim_{T\to\infty}   \kappa^-_T = 0$ $\mathbb{P}_\theta$-a.s. This together with the continuity of $J$ implies that the estimators \eqref{eq:OU:optimal:interval} are (strongly) consistent, i.e., $\lim_{T\to\infty} \bar J^0_{T,r}(\widehat\theta_T)=\lim_{T\to\infty} \ubar J^0_{T,r}(\widehat\theta_T) = J(\theta)$ holds $\mathbb{P}_\theta$-a.s.

To derive the formulas \eqref{eq:OU:optimal:interval}, we note that by a change of variables $\tilde\theta = 1/\theta$, the two optimization problems become convex. Then we exploit the property \cite[Lemma~9.2]{bertsimas-LPbook}, stating that for any $\theta'\in\mathbb{R}^d$ and $A\in\mathbb{R}^{d\times d}$, the optimization problem 
\begin{equation*}
    \min_{\theta\in\mathbb{R}^d} \{c^\top \theta \ : \ (\theta-\theta')^\top A^{-1} (\theta-\theta') \leq \rho\},
\end{equation*}
admits a closed form solution $\theta^\star = \theta' - \sqrt{\frac{\rho}{c^\top A c}}Ac$.
 Figure~\ref{sfig:OU:new} displays the optimal confidence interval ${\mathcal{I}}^{0,\star}_{T,r}(\widehat\theta_T) = [\ubar J^0_{T,r}(\widehat \theta_T),\bar J^0_{T,r}(\widehat \theta_T)]$ as a function of the sample size $T$.

\end{example}

\section{Preparation and proofs of main results}\label{sec:proof:main:result}

Before stating the preliminary results and the proofs of our main results, we quickly discuss the high-level proof roadmap, with the hope to guide the reader throughout this section.

\vspace{1mm}

{\grn
\noindent \textbf{A concise proof roadmap:} 
\vspace{-1mm}
\begin{itemize}
 \item The first step in our proof is to show that, for a given estimator-value $\theta'\in\esttvs$, the set of compatible parameters $\scr{I}^\delta_{T,r}(\theta')$ is bounded for all sufficiently large sample sizes $T$. Since $J$ is Lipschitz continuous on bounded sets, this guarantees that the lower and upper endpoints, defined by the optimization problems in \eqref{eq:optJ-mod}, are finite. In fact, we prove a stronger, locally uniform result: when the estimator-value $\theta'$ ranges over any bounded subset of $\esttvs$, both the compatible parameter values $\theta\in\scr{I}^\delta_{T,r}(\theta')$ and the corresponding scaled deviations from the observed estimator-value, $a_T(\theta'-\theta)$, are uniformly bounded for all sufficiently large $T$ (see Lemma \ref{lem:unif-cpt}).

\item  These boundedness properties play a crucial role in establishing the
convergence of the endpoint functions
$\ubar{J}^\delta_{T,r}(\cdot)$ and $\bar{J}^\delta_{T,r}(\cdot)$ to
$J(\cdot)$ uniformly on bounded sets with speed $1/a_T$ (see Theorem \ref{thm:minimal:interval}-\ref{item:thm:uniformly:compact}). This, together
with the fact that the MDP implies
$a_T(\widehat\theta_T-\theta)\stackrel{\PP_\theta}{\Rt}0$, establishes
the consistency of the resulting CI
$\mathcal{I}^{\delta,\star}_{T,r}(\widehat\theta_T)$ (i.e., the consistency of the endpoints $\ubar{J}^\delta_{T,r}(\widehat\theta_T),\bar{J}^\delta_{T,r}(\widehat\theta_T)$), in fact with speed $1/a_T$ (see Theorem \ref{th:prop-J}-\ref{item:thm:consistent}).

 \item The exponential accuracy of the CI $\mathcal{I}^{\delta,\star}_{T,r}(\widehat\theta_T) = [\ubar{J}^\delta_{T,r}(\widehat\theta_T),\bar{J}^\delta_{T,r}(\widehat\theta_T)]$ (see Definition \ref{def:exponentially:accurate}) relies on the MDP upper bound for $\widehat \theta_T$
 The main idea is that, when the true value of the parameter is $\theta$, $J(\theta)$ will lie outside the CI $\mathcal{I}^{\delta,\star}_{T,r}(\widehat\theta_T)$ only if $\theta$ is not in the compatibility set  $\scr{I}^\delta_{T,r}(\widehat\theta_T)$, which in turn is equivalent to the scaled deviation $a_T(\widehat\theta_T-\theta)$ lying outside $\sblvl_{\theta,r}^\delta$.
This also explains the role of the $\delta$-fattening of
$\sblvl_{\theta,r}$. Since the MDP upper bound applies to closed sets,
and the relevant non-coverage event involves the complement of the set of admissible
deviations, we enlarge $\sblvl_{\theta,r}$ to the open set
$\sblvl^\delta_{\theta,r}$, whose complement is closed. The
$\delta$-fattening thus provides a small open enlargement that allows
the MDP upper bound to be applied directly. Naturally, we would like this
enlargement to be as small as possible. In particular, when
$I^M_\theta$ is continuous, we show that one can work with the limiting
choice $\delta=0$, with $\sblvl^0_{\theta,r}\equiv\sblvl_{\theta,r}$ (see Theorem \ref{thm:minimal:interval}-\ref{item:opt:interval:exp-accu}).

\item The minimality of the interval family
$\{\mathcal{I}^{\delta,\star}_{T,r}:T>0\}$, on the other hand, relies on the
MDP lower bound. In fact, we prove a stronger result that is of independent
interest. Proposition \ref{prop:opt} shows that if one attempts to construct a
smaller CI by moving any of the endpoint functions of the optimal interval family
inward by an amount of order $1/\alpha_T$, where $\alpha_T=o(a_T)$, then the
resulting CI not only fails to be exponentially accurate with error rate
$r$, but its noncoverage probability fails to decay exponentially at all;
that is, its normalized log-error probability converges to zero.

\item Finally, Proposition \ref{prop:opt} is also used to establish a complementary interpretation of the minimality of the optimal CI in terms of its ability to exclude incorrect values. Specifically, for any
$\tilde\theta\in\Theta$ with $J(\tilde\theta)\neq J(\theta)$,
Proposition \ref{prop:opt} leads to the fact that the probability of an
incorrect value $J(\tilde\theta)$ not belonging to the optimal CI does not
have a nonzero exponential decay rate (see Theorem \ref{th:prop-J}-\ref{thm:minimal:interval:mischa}). Thus, although the probability of the true value $J(\theta)$ not belonging to the CI is exponentially small, the probability of correctly excluding
an incorrect value $J(\tilde\theta)$ is not exponentially small, providing
a complementary sense in which the optimal CI is not unnecessarily
conservative.

\end{itemize}

}

To prove our main results (Theorems~\ref{thm:minimal:interval} and \ref{th:prop-J}) we need a few preparatory steps.
We start with the following easy but important observation.
\begin{lemma}\label{lem:sub}
Let $J_{T,r}^\delta$ be as defined by \eqref{eq:optJ-mod}.
For any $T>0$ and $\theta\in\Theta$, 
$$a_T\lf((\bar J_{T,r}^\delta)^{-1}(-\infty, J(\theta)) - \theta\ri) \subset \lf(\sblvl_{\theta,r}^\delta\ri)^c.$$
\end{lemma}

\begin{proof} Fix $T>0$, and let $\vart \in a_T\lf((\bar J_{T,r}^\delta)^{-1}(-\infty, J(\theta)) - \theta\ri).$ Then $\bar J_{T,r}^\delta \lf(\theta+\vart/a_T\ri) < J(\theta).$
Suppose, if possible, $ \vart \in \sblvl_{\theta,r}^\delta$. This is equivalent to saying that $\vart = a_T\lf((\theta+\vart/a_T) -\theta\ri) \in \sblvl_{\theta,r}^\delta$. Then the definition of $\bar J^\delta_{T,r}$ implies $\bar J^\delta_{T,r}\lf(\theta+\vart/a_T\ri) \geq J(\theta)$, which is a contradiction.
\end{proof}

We now establish a few auxiliary results that are necessary for the proofs of the main results. 
Under Assumption~\ref{assum:basic}, we will show that for a bounded set $\Theta_0$,  $\cup_{\theta_0\in \Theta_0}\scr{I}^\delta_{T,r}(\theta_0)$ is bounded for sufficiently large $T$.

\begin{lemma}\label{lem:pseudo-cpt}
Suppose that $\esttvs_0 \subset \esttvs$ is bounded.
Suppose $I^M$ satisfies Assumption \ref{assum:basic}-\ref{item:4:ass:jointly:good:alternative}. Let $\{\theta_T\} \subset \Theta$ and $\{\theta'_{0,T}\} \subset \esttvs_0$ be sequences such that for a fixed $r \in (0,\infty)$, $a_T(\theta'_{0,T}-\theta_T) \in \sblvl_{\theta_T,r}^\delta$ for all $T>0.$ 
Then there exist constants $C_0 \equiv  C(\esttvs_0)$ and $T_0>0$  such that $\max\{\met{\theta_T}, a_T\|\theta'_{0,T}-\theta_T\|\} \leq C_0$ for all $T\geq T_0$.
\end{lemma}

\begin{proof}
 We first prove that $\limsup \limits_{T \rt \infty}\met{\theta_T} <\infty$.
If this assertion is not true, then $\limsup \limits_{T \rt \infty}\met{\theta_T} =\infty$. Since each $\theta'_{0,T}$ lies in the bounded set $\esttvs_0$ and $a_T \rt \infty$, both $\limsup \limits_{T \rt \infty}\met{\vart_T} =\infty$ and  $\limsup \limits_{T \rt \infty}\met{\vart_T}/\met{\theta_T} =\infty$,
where $\vart_T \equiv a_T(\theta'_{0,T}-\theta_T)$. 
 By the definition of $\sblvl_{\theta_T,r}^\delta$, there exists $\tilde \vart_T \in \sblvl_{\theta_T,r}$ such that $\|\vart_T-\tilde \vart_T\| < \delta$. It is obvious that $\limsup \limits_{T \rt \infty}\met{\vart_T} =\infty$ iff $\limsup \limits_{T \rt \infty}\met{\tilde\vart_T} =\infty$, and since $\limsup \limits_{T \rt \infty}\met{\theta_T} \neq 0$, $\limsup \limits_{T \rt \infty}\met{\vart_T}/\met{\theta_T} =\infty$ iff $\limsup \limits_{T \rt \infty}\met{\tilde\vart_T}/\met{\theta_T} =\infty$.
 
Consequently, there exists a sequence $\{T_n\}$ such that  $\lim \limits_{n \rt \infty}\met{\theta_{T_n}} =\infty$ and $\lim \limits_{n \rt \infty}\met{\tilde\vart_{T_n}} =\infty, \ \lim \limits_{n \rt \infty}\met{\tilde\vart_{T_n}}/\met{\theta_{T_n}} =\infty$. By Assumption \ref{assum:basic}-\ref{item:ass:growth:RF}, $\limsup \limits_{n \rt \infty} I^M_{\theta_{T_n}}(\tilde\vart_{T_n}) = \infty$. But this is a contradiction to the hypothesis that $I^M_{\theta_T}(\tilde\vart_T) \leq r$ for all $T>0$, which in particular implies $\limsup \limits_{n \rt \infty} I^M_{\theta_{T_n}}(\tilde\vart_{T_n}) \leq r$.

Now suppose $\limsup \limits_{T \rt \infty}\met{\theta_T} < \infty$, but $\limsup \limits_{T \rt \infty}\met{\vart_T} =\infty.$ Then Assumption \ref{assum:basic}-\ref{item:ass:cpt:RF} leads to a contradiction. 
\end{proof}

It is implied in the statement of Lemma~\ref{lem:pseudo-cpt} that the the constant $C(\esttvs_0)$ can be chosen independent of the family $\{\theta_T:T>0\}$. We next show that this is indeed the case.

\begin{lemma}\label{lem:unif-cpt}
Let $\esttvs_0 \subset \esttvs$ be bounded, and suppose that $I^M$ satisfies Assumption \ref{assum:basic}-\ref{item:4:ass:jointly:good:alternative}.  Then there exist constants $C_0 \equiv C(\esttvs_0)>0$, $T_0\equiv T(\esttvs_0)>0$, such that for all $T\geq T_0$, 
\begin{align}
\sup_{T\geq T_0}\sup_{\theta'_0 \in \esttvs_0}\sup_{\theta \in \scr{I}^\delta_{T,r}(\theta'_0)} \max\{ \met{\theta}, a_T\met{\theta'_0-\theta}\} \leq C_0,
\label{eq:unif-bd}
\end{align}
where $\scr{I}^\delta_{T,r}$ was introduced in \eqref{eq:feasible:set:ub}
\end{lemma}

\begin{proof} We first prove that $\dst \sup_{T\geq T_0}\sup_{\theta'_0 \in \esttvs_0}\sup_{\theta \in \scr{I}^{\delta}_{T,r}(\theta'_0)}\met{\theta} < \infty$ for some $T_0>0$. 
If it is not true, then for every $M$ and every $t$, there exist $T_M \geq t$, $\theta'_{0,T_M} \in \esttvs_0$ with $\scr{I}^\delta_{T_M,r}(\theta'_{0,T_M}) \neq \emptyset$,  and $\theta_{T_M} \in  \scr{I}^\delta_{T_M,r}(\theta'_{0,T_M})$, such that $\met{\theta_{T_M}} > M$. Here we are using the convention that $\sup \emptyset = -\infty$. It is clear that we can choose the sequence $\{T_M\}$ such that $T_{M+1} > T_M$. The sequence $\{\theta_{T_M}: M\geq 1\}$ clearly satisfies the hypothesis of Lemma \ref{lem:pseudo-cpt} but violates its conclusion. The other assertion follows similarly.
\end{proof}

We need another technical lemma stating a sufficient condition for a family of functions to be uniformly smaller than some given family of functions on compact sets.

\begin{lemma}[Sufficient condition for uniformly smaller on compact sets]\label{lem:ev-small-cpt}
Let $\estsp$ be a metric space, and $\{\psi_T: T>0\} $ and $\{\phi_T: T>0\}$ be  eventually equi-l.s.c. and eventually equi-u.s.c. function-families, respectively.   Suppose that
$\{\phi_T: T>0\}$ is eventually smaller than  $\{\psi_T: T>0\}$.  Then $\{\phi_T: T>0\}$ is 
eventually and uniformly smaller than $\{\psi_T: T>0\}$ on compact sets of $\estsp$. In particular, the conclusion holds if both $\{\psi_T: T>0\}$ and $\{\phi_T: T>0\}$ are eventually equicontinuous.
\end{lemma}

\begin{proof}
We prove the result by contradiction. Assume that if possible for some compact subset $\estsp_0\subset \estsp$ and $\vep_0>0$,
$$\limsup_{T\rt \infty} \sup_{\metvar \in \estsp_0}(\phi_T(\metvar) - \psi_T(\metvar)) > \vep_0.$$
In other words for every $T'_0\geq 0$, there exist $T'>T'_0$ and $\metvar' \in \estsp_0$ such that $\phi_{T'}(\metvar') - \psi_{T'}(\metvar') > \vep_0$. Consequently, there exist sequences $\{T_k\} \nearrow \infty$ and $\{\metvar_k\} \subset \estsp_0$ such that $\phi_{T_k}(\metvar_k) - \psi_{T_k}(\metvar_k) > \vep_0.$ Since $\estsp_0$ is compact, $\metvar_k \stackrel{k \rt \infty}\rt \metvar_0$ along a subsequence (which we continue to denote as $\{\metvar_k\}$) for some limit point $\metvar_0 \in \estsp_0$.
By eventual equi-u.s.c. of $\{\phi_T\}$ and eventual l.s.c. of $\{\psi_T\}$, there exist $ \delta = \delta(\metvar_0,\vep_0)$ and $T_0 = T_0(\metvar_0,\vep_0)$ such that
\begin{align*}
\lf(\phi_T(\metvar) - \phi_T(\metvar_0) \ri) \vee \lf(\psi_T(\metvar_0) - \psi_T(\metvar) \ri) < \vep_0/4, \quad \text{ for all } \metvar \in B(\metvar_0,\delta), \ T\geq T_0.
\end{align*}
Now let $k_0$ be such that for all $k\geq k_0$, $T_k \geq T_0$ and $\metvar_k \in B(\metvar_0,\delta)$. Consequently, for $k\geq k_0$,
\begin{align*}
\phi_{T_k}(\metvar_0) \!-\! \psi_{T_k}(\metvar_0)=&\ \!\lf(\phi_{T_k}(\metvar_0) \!-\! \phi_{T_k}(\metvar_k)\ri)\!+\! \lf(\phi_{T_k}(\metvar_k) - \psi_{T_k}(\metvar_k)\ri)\! + \!\lf(\psi_{T_k}(\metvar_k) - \psi_{T_k}(\metvar_0)\ri)\\
\geq&\ -\vep_0/4+\vep_0-\vep_0/4 = \vep_0/2.
\end{align*}
This shows that $\{\phi_T\}$ is not eventually smaller than $\{\psi_T\}$ at $\metvar_0$, which contradicts the hypothesis.
\end{proof}

We next need the following lemma on pointwise convergence of $\bar J_{T,r}$. But before we prove it we make a useful observation. Since $J$ is not assumed to be non-negative,   it is enough to only prove the assertions about  $\bar J^\delta_{T,r}$ in Theorem \ref{thm:minimal:interval}, Theorem \ref{th:prop-J} and the auxiliary results below. The corresponding results for $\ubar J^\delta_{T,r}$ then automatically follow upon observing 
\begin{align}\label{eq:ubar-J}
-\ubar {J}^\delta_{T,r}(\theta') = \sup_{\theta\in\Theta} \lf\{ -J(\theta)\ : \ a_T(\theta'-\theta) \in \sblvl^\delta_{\theta,r}\ri\}.
\end{align}

{\grn
\begin{lemma}[Uniform convergence]\label{lem:cont-conv-w}
Suppose that Assumption~\ref{assum:para}-\ref{ass:item:parsp-0} \& \ref{item:conti:J} and Assumption~\ref{assum:basic}  hold. Then Theorem \ref{thm:minimal:interval}-\ref{item:thm:uniformly:compact} holds.

\end{lemma}

\begin{proof}
As just noted, it is enough to prove the assertion for $\bar J^\delta_{T,r}$.

Fix a bounded set $\opintm_0 \subset \opintm$. Take $\esttvs_0 = \opintm_0 $ in Lemma \ref{lem:unif-cpt}, and let $T_0$ and $C_0 \equiv  C(\esttvs_0 \equiv \opintm_0)$ be as in that lemma. 
Fix any $\theta' \in \esttvs_0 \equiv \opintm_0$, and $T \geq T_0$. First assume that $\scr{I}^{\delta}_{T,r}(\theta') \neq \emptyset$. Then  there exists $\theta_{T}^\star \in \scr{I}^{\delta}_{T,r}(\theta') \subset \Theta$  such that 
\begin{align}\label{eq:auxJineq-2}
\bar J^\delta_{T,r}(\theta') -  1/a_T \leq J(\theta_{T}^\star) \leq \bar J^\delta_{T,r}(\theta').
\end{align} 
 By Lemma \ref{lem:unif-cpt},  $\max\{\|\theta_{T}^\star\|, \|\vart_{T}^\star\|\} \leq C_0$ for all $T\geq T_0$, where $\vart_{T}^\star \equiv a_{T}(\theta'-\theta_{T}^\star)$. By the local Lipschitz continuity of $J: \opintm \rt \R$, for any $\tilde C_0$, there exists a constant $L_J \equiv L_J(\tilde C_0)$ such that
\begin{align*}
| J(u_1) - J(u_2)| \leq L_J(\tilde C_0)\|u_1 - u_2\|, \quad  u_i\in \Theta,\  \|u_i\| \leq \tilde C_0
\end{align*}
Take $\tilde C_0 = C_0\vee \max_{\theta' \in \esttvs_0}\|\theta'\|$, and notice that \eqref{eq:auxJineq-2} imply
\begin{align*}
 J(\theta') -  \bar J^\delta_{T,r}(\theta')  \leq J(\theta') -  J(\theta_{T}^\star) \leq |J(\theta') -  J(\theta_{T}^\star)| \leq L_J \|\vart_{T}^\star\|/a_T \leq L_j C_0/a_T,
\end{align*}
and
\begin{align*}
    \bar J^\delta_{T,r}(\theta') - J(\theta') & \leq (J(\theta_{T}^\star) - J(\theta')) + 1/a_T \leq L_J \|\theta'-\theta_{T}^\star\| + 1/a_T\\
    & \leq (L_J \|\vart_{T}^\star\|+1)/a_T \leq (L_j C_0+1)/a_T.
\end{align*}
Thus
\begin{align}
    \label{eq:endpt-J-diff}
  |\bar J^\delta_{T,r}(\theta') - J(\theta')| \leq  (L_j C_0+1)/a_T 
\end{align}
 If $\scr{I}^{\delta}_{T,r}(\theta') = \emptyset$, then by definition $ \bar J^\delta_{T,r}(\theta') -J(\theta')=0$, and \eqref{eq:endpt-J-diff} is automatically true. This proves the assertion.

\end{proof}
}
{\grn 
\begin{proposition} \label{prop:opt} Suppose that Assumption~\ref{assum:para}-\ref{ass:item:parsp-0} \& \ref{item:conti:J} and Assumption~\ref{assum:basic} hold, and $J:\Theta \rt \R$ is continuous.
Suppose for some inflation factor $\alpha_T \neq 0$ and for some $\theta_0 \in \Theta$, $\Delta \equiv \liminf\limits_{T\rt \infty}\alpha_T\lf(\bar J_{T,r}^\delta(\theta_0) - \bar \Phi_{T}(\theta_0)\ri) >0$. Further assume that the family $\{\alpha_T\lf(\bar{\Phi}_{T}(\cdot) - J(\cdot)\ri) \}$ is  eventually equi-u.s.c. at $\theta_0$. Then $$\lim\limits_{T\to\infty}\frac{1}{b_{T}} \log \PP_{\theta_0}\lf(J(\theta_0) > \bar \Phi_{T}(\widehat \theta_{T})\ri)=0.$$
\end{proposition}

\begin{proof} Clearly, $\limsup\limits_{T\to\infty}\frac{1}{b_{T}} \log \PP_{\theta_0}\lf(J(\theta_0) > \bar \Phi_{T}(\widehat \theta_{T})\ri) \leq 0.$ To show the other inequality, we first establish the following claim.

\np 
{\em Claim:} For any fixed $\eta>0$, $B(0, \eta) \subset a_{T}\lf(\bar \Phi_{T}^{-1}(-\infty, J(\theta_0)) -\theta_0\ri) $ for sufficiently large $T$. 

Toward this end, let $T_0$ be such that $\alpha_T\lf(\bar J_{T,r}^\delta(\theta_0) - \bar \Phi_{T}(\theta_0)\ri) > 3\Delta/4$  for $T\geq T_0$. Since $\alpha_T/a_T \rt 0$ as $T \rt \infty$,  by Lemma \ref{lem:cont-conv-w},  there exists $T_1\geq T_0$ such that for all $T \geq T_1$,
\begin{align}\label{eq:Jineq2}
 \alpha_T\lf(J(\theta_0) - \bar \Phi_{T}(\theta_0)\ri) \geq \Delta/2.
\end{align}
Next equi-u.s.c. of $\lf\{\alpha_T\lf(\bar{\Phi}_{T} - J\ri)\ri\}$ at $\theta_0 \in \Theta \subset \opintm$ and openness of $\opintm$ imply that there exist $\rho>0$ and $T_2\geq T_1$, such that for $u \in B(\theta_0,\rho) \subset \opintm \subset \esttvs$ and $T \geq T_2$,
\begin{align}\label{eq:eqcontH}
\alpha_T\lf(\bar \Phi_{T}(u) - J(u) - \bar \Phi_{T}(\theta_0) + J(\theta_0)\ri)  < \Delta/4.
\end{align}

Let $T_3 \geq T_2$ be such that $\eta/a_{T} < \rho$ for all $T \geq T_3.$ It follows that for any $T \geq T_3$ and $\vart \in B(0, \eta)$,
$
\met{\lf(\theta_0+\vart/a_{T}\ri) - \theta_0} \leq \eta/a_{T} < \rho
$.  Consequently, by \eqref{eq:eqcontH}, we have for any $\vart \in B(0, \eta)$, 
$$\sup_{T\geq T_3} \alpha_T \lf(\bar \Phi_{T}(\theta_0+\vart/a_{T}) -J(\theta_0+\vart/a_{T})- \bar \Phi_{T}(\theta_0) + J(\theta_0)\ri) < \Delta/4.$$
It now follows from \eqref{eq:Jineq2} that for all $T \geq T_3$ and any  $\vart \in B(0, \eta)$,
\begin{align}\label{eq:Phi_ineq-0}
\begin{aligned}
\bar \Phi_{T}(\theta_0+\vart/a_{T}) <&\ \bar \Phi_{T}(\theta_0) +(J(\theta_0+\vart/a_{T}) - J(\theta_0)) +\f{\Delta}{4\alpha_T}  \\
\leq&\  J(\theta_0)+(J(\theta_0+\vart/a_{T}) - J(\theta_0)) -\f{\Delta}{4\alpha_T}.
\end{aligned}
\end{align}
Now by the local-Lipschitz property of $J\big|_{\opintm}:\opintm \to \R$ and the fact that $\alpha_T = o(a_T)$, it follows we can find $T_4 \geq T_3$ such that for all $T \geq T_4$, $\met{J(\theta_0+\vart/a_{T}) - J(\theta_0)} \leq \f{\Delta}{8\alpha_T}$. Hence from \eqref{eq:Phi_ineq-0} we have
\begin{align*}
\bar \Phi_{T}(\theta_0+\vart/a_{T}) < J(\theta_0) - \f{\Delta}{8\alpha_T} < J(\theta_0),
\end{align*}
which proves the claim. 

Hence, we have for all $T \geq T_3$
\begin{align}\label{ineq:prep:LDP:optimality:proof}
\begin{aligned}
\PP_{\theta_0}\lf( J(\theta_0) > \bar \Phi_{T}(\widehat \theta_{T})\ri) 
&= \PP_{\theta_0}\lf( a_{T}(\widehat \theta_{T} - \theta_0) \in a_{T}\lf(\bar \Phi_{T}^{-1}(-\infty, J(\theta_0)) -\theta_0\ri) \ri) \\
&\geq \PP_{\theta_0}\lf( a_{T}(\widehat \theta_{T} - \theta_0) \in  B(0, \eta)\ri).
\end{aligned}
\end{align}
By Assumption~\ref{assum:basic}-\ref{ass:item:MDP} the family $\{a_T(\widehat \theta_T - \theta_0)\}$ satisfies an LDP with speed $\{b_T\}$ and rate function $I^M_{\theta_0}$. Consequently,  \eqref{ineq:prep:LDP:optimality:proof} implies that
\begin{align}
\non
\liminf\limits_{T\to\infty}\frac{1}{b_{T}} \log \mathbb{P}_{\theta_0}\lf( J(\theta_0) > \bar \Phi_{T}(\widehat \theta_{T})\ri)
\geq &\ \liminf\limits_{n\to\infty}\frac{1}{b_{T}} \log \PP_{\theta_0}\lf( a_{T}(\widehat \theta_{T} - \theta_0) \in  B(0, \eta)\ri) \\ \non
\geq &\  - \inf_{\vartheta\in B(0, \eta)}I^M_{\theta_0} (\vartheta)
\geq \ -I^M_{\theta_0} \lf(0\ri)=0.
\end{align} 
\end{proof}
}

We are now ready to prove Theorem \ref{thm:minimal:interval}.

\begin{proof}[Proof of Theorem \ref{thm:minimal:interval}]
As mentioned in the paragraph before Lemma \ref{lem:cont-conv-w}, it is enough to prove the statements for $\bar J^\delta_{T,r}$. The corresponding assertions for $\ubar J^\delta_{T,r}$ follow from \eqref{eq:ubar-J}.

\np
\ref{item:opt:interval:exp-accu} Note that exponential accuracy of the interval family $\{\mathcal{I}^{\delta,\star}_{T,r}:T>0\}$, that is, \eqref{eq:def:UMA:cond}  easily follows from \eqref{eq:feas-1}. To see this simply observe that
\begin{align*}
\mathbb{P}_{\theta_0}\lf( J(\theta_0) \notin[\ubar J^\delta_{T,r}(\widehat \theta_T),\bar J^\delta_{T,r}(\widehat \theta_T)] \ri) 
\leq&\  \mathbb{P}_{\theta_0}\lf( J(\theta_0)>\bar J^\delta_{T,r}(\widehat \theta_T)\ri) + \mathbb{P}_\theta\lf( J(\theta_0)< \ubar J^\delta_{T,r}(\widehat \theta_T)\ri)\\
\leq&\ 2 \PP_{\theta_0}\lf( J(\theta_0)>\bar J^\delta_{T,r}(\widehat \theta_T)\ri) \vee \mathbb{P}_{\theta_0}\lf( J(\theta_0)< \ubar J^\delta_{T,r}(\widehat \theta_T)\ri),
\end{align*}
which implies that $\mfk{p}_e \equiv     \limsup\limits_{T\to\infty}\frac{1}{b_T} \log \mathbb{P}_{\theta_0}\lf( J(\theta_0) \notin[\ubar J^\delta_{T,r}(\widehat \theta_T),\bar J^\delta_{T,r}(\widehat \theta_T)] \ri)$ satisfies
\begin{align*}
\mfk{p}_e \leq &  \max\lf\{ \limsup\limits_{T\to\infty}\frac{1}{b_T} \log \PP_{\theta_0}\lf( J(\theta_0)>\bar J^\delta_{T,r}(\widehat \theta_T)\ri),\limsup\limits_{T\to\infty}\frac{1}{b_T} \log \mathbb{P}_{\theta_0}\lf( J(\theta_0)< \ubar J^\delta_{T,r}(\widehat \theta_T)\ri)\ri\}\\
& \hs{.1cm} + \limsup\limits_{T\to\infty}\frac{1}{b_T} \log 2\ \leq\ -r.
\end{align*}
We now establish the first inequality in \eqref{eq:feas-1}. 

Note that by Lemma \ref{lem:sub} and the LDP upper bound
\begin{align}
\non
    &\limsup_{T \rt \infty} \frac{1}{b_T} \log  \PP_{\theta_0}\lf(J(\theta_0)>\bar J^\delta_{T,r}(\widehat \theta_T)\ri)\\ 
 \non   
    &\qquad=  \limsup_{T \rt \infty} \frac{1}{b_T} \log  \PP_{\theta_0}\lf( a_T\lf(\widehat\theta_T -\theta_0\ri) \in a_T\lf(\lf(\bar J^\delta_{T,r}\ri)^{-1}(-\infty, J(\theta_0)) -\theta_0\ri)\ri)\\
  \non  
    &\qquad\leq  \limsup_{T \rt \infty} \frac{1}{b_T} \log  \PP_{\theta_0}\lf( a_T\lf(\widehat\theta_T -\theta_0\ri) \in \lf(\sblvl_{\theta_0,r}^\delta\ri)^c\ri)\\
  \label{eq:exp-accu-est}  
    &\qquad\leq\  - \inf_{\vartheta\in (\sblvl_{\theta_0,r}^\delta)^c}I^M_{\theta_0}(\vartheta) \leq -r.
\end{align}
The first inequality in \eqref{eq:exp-accu-est} follows as $(\sblvl_{\theta_0,r}^\delta)^c$ is closed, as mentioned in \eqref{eq:sblvl-fat}, and
the last inequality holds since $(\sblvl_{\theta_0,r}^\delta)^c  \subseteq (\sblvl_{\theta_0,r})^c = \{\vartheta\in\esttvs: I^M_{\theta_0}(\vartheta) > r\}$. 

Now suppose that $I^M_{\theta_0}:\esttvs \rt [0,\infty]$ is continuous, and take $\delta=0$. Then the assertion follows from the observation that the bound in \eqref{eq:exp-accu-est} holds with $(\sblvl_{\theta_0,r}^\delta)^c$ replaced by $\overline{(\sblvl_{\theta_0,r})^c }$, and by continuity of $I^M_{\theta_0}$,  $\overline{(\sblvl_{\theta_0,r})^c }\subseteq \{\vart\in\esttvs : I^M_{\theta_0}(\vart) \geq r\}.$\\

\np
\ref{item:thm:uniformly:compact} This is proved in Lemma \ref{lem:cont-conv-w}.\\

\np
\ref{item:opt:interval:equi} This is a direct consequence of  \ref{item:thm:uniformly:compact}.\\

\np
\ref{item:opt:interval:pt-min} {\grn First note that \ref{item:opt:interval:exp-accu} and \ref{item:opt:interval:equi} show that $\{\mathcal{I}^{\delta,\star}_{T,r}=[\ubar J_{T,r}^\delta, \bar J_{T,r}^\delta] :T>0\}$ is an element of $\mathbb{I}_r\lf(\equicls_{\Theta}\ri) \subset \mathbb{I}_r\lf(\SC{L}_{\Theta},\ \SC{U}_{\Theta}\ri)$. Now  suppose it does not satisfy the assertion of \ref{item:opt:interval:pt-min}. Then  there exists a family $\{[\ubar H_{T,r},\bar H_{T,r}]:T>0\} \in \mathbb{I}_r\lf(\SC{L}_{\Theta},\ \SC{U}_{\Theta}\ri)$ and a $\theta_0 \in \Theta$  such that for some $\alpha_T = o(a_T)$,
\begin{align*}
2\Delta \equiv \limsup_{T\rt \infty} \alpha_T\lf(\bar J_{T,r}^\delta(\theta_0) - \bar H_{T,r}(\theta_0)\ri) \vee  \limsup_{T\rt \infty}  \alpha_T \lf(\ubar H_{T,r}(\theta_0)-\ubar J_{T,r}^\delta(\theta_0)\ri) > 0,
\end{align*}
and for specificity, we assume that
\begin{align*}
2\Delta \equiv \limsup_{T\rt \infty} \alpha_T\lf(\bar J_{T,r}^\delta(\theta_0) - \bar H_{T,r}(\theta_0)\ri) > 0.
\end{align*}
Then, there exists a sequence $\{T_n\}$ such that $\alpha_{T_n}(\bar J_{T_n,r}^\delta(\theta_0) - \bar H_{T_n,r}(\theta_0)) \rt 2\Delta$ as $n \rt \infty.$
Therefore, there exists $n_0$ such that for all $n\geq n_0$, 
\begin{align*}%
\alpha_{T_n}(\bar J_{T_n,r}^\delta(\theta_0) - \bar H_{T_n,r}(\theta_0)) \geq \Delta.
\end{align*}
Consequently, by Proposition \ref{prop:opt}
\begin{align*}
\limsup\limits_{T\to\infty}\frac{1}{b_{T}} \log \mathbb{P}_{\theta_0}\lf( J(\theta_0) > \bar H_{T,r}(\widehat \theta_{T})\ri) \geq &\
\lim_{n\to\infty}\frac{1}{b_{T_n}} \log \mathbb{P}_{\theta_0}\lf( J(\theta_0) > \bar H_{T_n,r}(\widehat \theta_{T_n})\ri)
=0 > -r.
\end{align*}

Since $\{J(\theta_0) > \bar H_{T,r}(\widehat \theta_{T})\} \subset \{J(\theta_0)  \notin [\ubar H_{T,r}(\widehat \theta_{T}),\bar H_{T,r}(\widehat \theta_{T})] \}$, the above inequality implies that 
\begin{align*}
\lim_{T\to\infty}\frac{1}{b_{T}} \log \mathbb{P}_{\theta_0}\lf( J(\theta_0) \notin  [\ubar H_{T,r}(\widehat \theta_{T}),\bar H_{T,r}(\widehat \theta_{T})] \ri) =0>-r.
\end{align*}
But that is a contradiction to the assumption that $\{[\ubar H_{T,r},\bar H_{T,r}]:T>0\} \in \mathbb{I}_r\lf(\SC{L}_{\Theta},\ \SC{U}_{\Theta}\ri)\big|_{\Theta}$.\\
}

\np
\ref{item:opt:interval:sol} {\grn By \ref{item:opt:interval:exp-accu} and \ref{item:opt:interval:equi}, $\{\mathcal{I}^{\delta,\star}_{T,r}=[\ubar J_{T,r}^\delta, \bar J_{T,r}^\delta] :T>0\}$ is an element of $\mathbb{I}_r\lf(\equicls_{\Theta}\ri)$. Now, let $\{[\ubar H_{T,r},\bar H_{T,r}]:T>0\} \in \mathbb{I}_r\lf(\equicls_{\Theta}\ri)$. By \ref{item:opt:interval:pt-min}, $\alpha_T \bar J_{T,r}^\delta\big|_\Theta$ is eventually smaller than $\alpha_T\bar H_{T,r}\big|_\Theta$ pointwise on $\Theta$. Equivalently, $\alpha_T(\bar J_{T,r}^\delta-J)\big|_\Theta$ is eventually smaller than $\alpha_T(\bar H_{T,r}-J)\big|_\Theta$ pointwise on $\Theta$.

Now, since $\alpha_T(\bar J_{T,r}^\delta-J),\ \alpha_T(\bar H_{T,r}-J):\esttvs\to\R$ are eventually equi-continuous at every $\theta\in\Theta$, its restriction $\alpha_T(\bar J_{T,r}^\delta-J)\big|_\Theta, \alpha_T(\bar H_{T,r}-J)\big|_\Theta:\Theta\to\R$ are also eventually equi-continuous. Note that the difference between the equicontinuity properties in the first and the second part is that, in the latter, equicontinuity at a point $\theta\in\Theta$ is with respect to neighboring points belonging only to $\Theta$, whereas in the former it is with respect to neighboring points in the ambient space $\esttvs$.

Thus taking $\estsp = \Theta$ in Lemma \ref{lem:ev-small-cpt}, it follows that 
$\alpha_T(\bar J_{T,r}^\delta-J)\big|_\Theta$ is eventually and uniformly smaller than $\alpha_T(\bar H_{T,r}-J)\big|_\Theta$ on compact sets of $\Theta$. A similar analysis holds for comparing the lower end point functions $\ubar J_{T,r}^\delta$ and $\ubar H_{T,r}.$}

\end{proof}

\begin{proof}[Proof of Theorem~\ref{th:prop-J}]
As before, we will only prove the assertions for $\bar J^\delta_{T,r}$.

\np
\ref{item:thm:consistent} {\grn Writing 
$$a_T(\bar J^\delta_{T,r}(\widehat \theta_T) - J(\theta_0)) = a_T (\bar J^\delta_{T,r}(\widehat \theta_T) - J(\widehat \theta_T)) + a_T(J(\widehat \theta_T) - J(\theta_0))$$
the assertion follows once we  show that the first term on the right side is stochastically bounded and the second converges to zero in probability with respect to $\PP_{\theta_0}$. 

Since $\theta_0 \in \Theta_0 \subset \opintm$, and $\opintm$ is open, there exists a $\rho>0$ such that $B(\theta_0, \rho) \subset \opintm.$ By Theorem \ref{thm:minimal:interval}-\ref{item:thm:uniformly:compact}, there exists $T_0>0$ such that 
$$M_0 \equiv \sup_{T\geq T_0} a_T\sup_{\theta' \in B(\theta_0, \rho)}\lf|\bar J^\delta_{T,r}(\theta') - J(\theta')\ri| < \infty.$$
Now fix $\epsilon>0$. Since $\widehat \theta_T \stackrel{\PP_{\theta_0}} \Rt \theta_0,$ make $T_0$ larger, if necessary, such that for any $T \geq T_0$,
\begin{align}\label{eq:est-op-ball}
 \PP_{\theta_0} \lf(\widehat\theta_T \notin B(\theta_0,\rho)\ri) \leq \ep.
\end{align}
 Then notice that for $T \geq T_0$,
\begin{align*}
    \PP_{\theta_0} \lf(a_T \lf |\bar J^\delta_{T,r}(\widehat \theta_T) - J(\widehat \theta_T)\ri| > M_0\ri) \leq&\ \PP_{\theta_0} \lf(a_T \lf |\bar J^\delta_{T,r}(\widehat \theta_T) - J(\widehat \theta_T)\ri| > M_0,\ \widehat \theta_T \in B(\theta_0, \rho)\ri) \\
    & + \PP_{\theta_0} \lf(\widehat\theta_T \notin B(\theta_0, \rho)\ri)
     \leq\ \epsilon,
\end{align*}
where we used \eqref{eq:est-op-ball} and the fact the first probability on the right side is 0. This is because when $\widehat \theta_T \in B(\theta_0, \rho)$
$$a_T \lf |\bar J^\delta_{T,r}(\widehat \theta_T) - J(\widehat \theta_T)\ri| \leq a_T\sup_{\theta' \in  B(\theta_0, \rho)}\lf|\bar J^\delta_{T,r}(\theta') - J(\theta')\ri| \leq M_0.$$
This shows that $\{a_T (\bar J^\delta_{T,r}(\widehat \theta_T) - J(\widehat \theta_T))\}$ is stochastically bounded. 

Finally, by the local Lipschitz continuity of $J\big|_\opintm$, when $\widehat \theta_T \in  B(\theta_0, \rho)$
$$a_T|J(\widehat \theta_T) - J(\theta_0)| \leq L_J(B(\theta_0, \rho)) a_T\|\widehat \theta_T - \theta_0\|.$$
It is now easy to show from Lemma~\ref{lem:est-conv-rate} and \eqref{eq:est-op-ball} that as $T \rt \infty$
$$a_T|J(\widehat \theta_T) - J(\theta_0)| \stackrel{\PP_{\theta_0}} \Rt 0.$$
}

\np
\ref{thm:minimal:interval:mischa} Fix a $\tilde \theta_0 \in \Theta$ such that $\eta \equiv J(\tilde \theta_0) - J(\theta_0) >0$. For $\theta' \in \esttvs$, define $\bar \Phi^\delta_{T,r}(\theta') = \bar J^\delta_{T,r}(\theta') + J(\theta_0) - J(\tilde \theta_0)$. Then notice that 
$\bar J^\delta_{T,r}(\theta') - \bar \Phi^\delta_{T,r}(\theta') = J(\tilde \theta_0) - J(\theta_0)=\eta>0$ for any $\theta'$ (in particular, for $\theta'=\theta_0$) and 
$\{J(\theta_0) > \bar \Phi^\delta_{T,r}(\widehat \theta_T) \} = \{J(\tilde \theta_0)>\bar J^\delta_{T,r}(\widehat \theta_T)\}$. Since $\{\bar J^\delta_{T,r}: T>0\}$ (and hence also $\{\bar \Phi^\delta_{T,r}: T>0\}$)  is eventually equicontinuous on $\Theta$ (cf. Theorem~\ref{thm:minimal:interval}-\ref{item:opt:interval:equi}),
the assertion immediately follows from Proposition \ref{prop:opt}.\\

\np
\ref{thm:minimal:interval:min:cons} {\grn We first assume strong consistency of $\widehat \theta_T$. Let $\Omega_0 \subset \Omega$ be such that $\PP_{\theta_0}(\Omega_0)=1$ and $\widehat \theta_T(\om) \stackrel{T\rt \infty}\Rt \theta_0$ for all $\om \in \Omega_0$. Fix an $\omega \in \Omega_0$.  Then the set $\{\theta_0, \widehat \theta_T(\om): T>0\} \subset \esttvs$ is  compact, and under assumption \ref{ass:item:parsp-b} it is actually a subset of $\Theta$. Next consider assumption \ref{ass:item:parsp-a}. Since $\Theta \subset \esttvs$ is open under this assumption, there exists a $\delta>0$ such that $B(\theta_0,\delta) \subset \Theta$. Since $\widehat \theta_T(\om) \stackrel{T\rt \infty}\Rt \theta_0$,  there exists $T_0(\om)$ such that $\{\theta_0, \widehat \theta_T(\om): T \geq T_0(\om)\} \subset B(\theta_0,\delta) \subset \Theta$. Clearly $\{\theta_0, \widehat \theta_T(\om): T \geq T_0(\om)\}$  is  compact. Thus in either case, there is a compact set $\Theta_0(\om) \subset \Theta$ and $T_0(\om)$ such that $\widehat \theta_T(\om) \in \Theta_0(\om)$ for all $T \geq T_0(\om)$.
Hence
\begin{align*}
\limsup_{T\rt \infty}\ &\alpha_T\Big(\bar J^\delta_{T,r}(\widehat \theta_T(\om))  - \bar H_{T,r}(\widehat \theta_T(\om))\Big)\\
\leq&\  \limsup_{T\rt \infty} \sup_{\theta' \in \Theta_0(\om)}\lf(\alpha_T \lf(\bar J^\delta_{T,r}(\theta') -J(\theta')\ri) - \alpha_T\lf(\bar H_{T,r}(\theta')-J(\theta')\ri)\ri) \leq 0,
\end{align*}
where the second inequality follows by Theorem \ref{thm:minimal:interval}-\ref{item:opt:interval:sol} (also see Definition \ref{def:minimal:interval}).

We next assume that $\widehat \theta_T$ is weakly consistent for $\theta_0$, that is, $\widehat \theta_T \stackrel{\PP_{\theta_0}} \rt \theta_0$ as $T \rt \infty$. Let $\{T_n\}$ be a sequence with $T_n \uparrow \infty$. Again by invoking the fact that convergence in probability implies almost sure convergence along a subsequence of any sequence, we can assume without loss of generality that $\widehat \theta_{T_n} \rt \theta_0 $ a.s. (w.r.t.~$\PP_{\theta_0}$). By the previous part it now follows that for any $\kappa>0$,
$$1_{\lf\{\alpha_T\lf(\bar J_{T_n,r}^\delta(\widehat \theta_{T_n}) - H_{T_n,r}(\widehat \theta_{T_n})\ri) > \kappa\ri\}} \stackrel{n \rt \infty}\Rt 0, \ \ \PP_{\theta_0}\text{-}a.s.,$$
hence the dominated convergence theorem readily implies that $$\PP_{\theta_0}\lf(\alpha_T\lf(\bar J_{T_n,r}^\delta(\widehat \theta_{T_n}) - H_{T_n,r}(\widehat \theta_{T_n})\ri) > \kappa\ri) \stackrel{n \rt \infty}\Rt 0.$$
Since the limit is independent of the sequence $\{T_n\}$, the assertion follows. \\
}

\np
\ref{thm:minimal:interval:uma-ci}  Fix $\tilde\theta_0 \in \Theta$  such that $J(\tilde\theta_0) \neq J(\theta_0)$, and let $\eta \equiv |J(\theta_0) - J(\tilde\theta_0)| >0$. By continuity of $J$ at $\theta_0$, there exists a $\kappa>0$  such that for any $\theta \in \Theta$, $\met{\theta - \theta_0} \leq \kappa$ implies $|J(\theta) - J(\theta_0)| \leq \eta/4.$ Fix an $\vep>0$. We now establish the following claim.

\np
{\bf Claim:} There exists a compact set, $\bar{\Theta}_\vep \subset \overline{B(\theta_0,\kappa)} \cap\Theta$ and $T_{0,\vep}$ such that for all $T\geq T_{0,\vep}$, $\PP_{\theta_0}(\widehat\theta_T \notin \bar{\Theta}_\vep) \leq \vep$.

{\grn To see the claim, first observe that, by a result of Le Cam
\cite[Theorem 2.3.6]{Boga18}, tightness of each $\widehat\theta_T$,
together with $\widehat\theta_T \stackrel{\PP_{\theta_0}}\Rt \theta_0$,
implies that the family $\{\widehat\theta_T\}$ is (uniformly) tight.}  Thus 
there exists a compact set $\bar{\SC{A}}_\vep \subset \esttvs$ and $T^1_{0,\vep}$ such that for $T\geq T^1_{0,\vep}$
\begin{align*}
\PP_{\theta_0}(\widehat\theta_T \in \bar{\SC{A}}_\vep^c) \leq \vep/2.
\end{align*}
Notice that if assumption \ref{ass:item:parsp-b} holds, then $\bar{\SC{A}}_\vep$ can be taken to be a compact subset of $\Theta$. By weak consistency of $\widehat\theta_T$, there exists $T^2_{0,\vep}$ such that for all $T\geq T^2_{0,\vep}$, 
$$\PP_{\theta_0}\lf(\met{\widehat\theta_T - \theta_0} > \kappa\ri) \leq \vep/2.$$
Now define  $\bar{\Theta}_\vep \equiv \bar{\SC{A}}_\vep \cap \overline{B(\theta_0,\kappa)}$ and take $T_{0,\vep} = T^1_{0,\vep} \vee T^2_{0,\vep}.$ 

If assumption \ref{ass:item:parsp-a} is valid, then there is a $\kappa' \leq \kappa$ such that $\overline{B(\theta_0,\kappa')} \subset \Theta.$ Again using weak consistency of $\widehat\theta_T$, we get a $T^3_{0,\vep}$ such that for all $T\geq T^2_{0,\vep}$, 
$$\PP_{\theta_0}\lf(\met{\widehat\theta_T - \theta_0} > \kappa'\ri) \leq \vep/2.$$
Now take  $\bar{\Theta}_\vep \equiv \bar{\SC{A}}_\vep \cap \overline{B(\theta_0,\kappa')}$ and  $T_{0,\vep} = T^1_{0,\vep} \vee T^3_{0,\vep}.$ \\

\np
Next by Theorem \ref{th:prop-J}-\ref{item:thm:uniformly:compact}, there exists $\bar{T}_{0,\vep} \geq T_{0,\vep}$ such that for $T\geq \bar{T}_{0,\vep}$, $\sup_{\theta' \in \bar{\Theta}_\vep}|\ubar J_{T,r}^{\delta}(\theta') - J(\theta')| \leq \eta/4$. We now consider two cases.\\

 \np
{\em Case I: $J(\theta_0) < J(\tilde\theta_0)$.} Notice that $J(\tilde\theta_0) > J(\theta_0)+\eta/2$. Hence by Theorem \ref{th:prop-J}-\ref{item:thm:consistent},
\begin{align}\label{eq:case-i}
\limsup_{T\to\infty}\mathbb{P}_{\theta_0}\left(J(\tilde\theta_0)\leq \bar{J}^{\delta}_{T,r}(\widehat\theta_T)\right) \leq  \limsup_{T\to\infty}\mathbb{P}_{\theta_0}\left(J(\theta_0)+\eta/2\leq \bar{J}^{\delta}_{T,r}(\widehat\theta_T)\right)= 0,
\end{align}
which immediately implies the assertion.\\

\np
{\em Case II: $J(\theta_0) > J(\tilde\theta_0)$.} Notice that $J(\tilde \theta_0)< J(\theta_0) - \eta/2$. We first show that there exists $\tilde T_{0,\vep}>0$ such that for all $T \geq \tilde T_{0,\vep}$
\begin{align}\label{eq:subset-case-ii}
\lf\{\theta':J(\tilde\theta_0)\leq \bar{J}^{\delta}_{T,r}(\theta')\ri\}\cap \bar{\Theta}_\vep  \subset \lf\{\theta':J(\tilde\theta_0)\leq \bar{H}_{T,r}(\theta')\ri\}.
\end{align}
To see this, suppose that  \eqref{eq:subset-case-ii} is not true. Then there exists a sequence $\{T_n \uparrow \infty\}$ and $\{\theta'_{T_n}\} \subset \bar{\Theta}_\vep$ such that 
\begin{align*}
\bar{H}_{T_n,r}(\theta'_{T_n}) <J(\tilde\theta_0)  \leq \bar{J}^{\delta}_{T_n,r}(\theta'_{T_n}).
\end{align*}
Let $n_\vep$ be such that $T_{n_\vep} \geq {\bar T}_{0,\vep}$. Then for any $n\geq n_\vep$
\begin{align*}
\bar{J}^{\delta}_{T_n,r}(\theta'_{T_n}) - J(\theta_0) = \bar{J}^{\delta}_{T_n,r}(\theta'_{T_n}) - J(\theta'_{T_n}) + J(\theta'_{T_n}) - J(\theta_0) \geq -(\eta/4+\eta/4) = -\eta/2.
\end{align*}
It follows that for any $n\geq n_\vep$
\begin{align*}
\bar{H}_{T_n,r}(\theta'_{T_n}) 
<J(\tilde\theta_0)
 \leq J(\theta_0) - \eta/2  
\leq \bar{J}^{\delta}_{T_n,r}(\theta'_{T_n}).
\end{align*}
Consequently, 
\begin{align*}
\limsup_{T\rt \infty} \sup_{\theta' \in  \bar{\Theta}_\vep}\lf(\bar{J}^{\delta}_{T_n,r}(\theta') -\bar{H}_{T_n,r}(\theta')\ri) \geq &\ \inf_{n\geq n_\vep} \lf(\bar{J}^{\delta}_{T_n,r}(\theta'_{T_n}) -\bar{H}_{T_n,r}(\theta'_{T_n})\ri)\\
  \geq &\ J(\theta_0) - \eta/2 - J(\tilde\theta_0) \geq \eta/2,
\end{align*}
which is a contradiction to Theorem \ref{thm:minimal:interval}-\ref{item:opt:interval:sol}. Hence, \eqref{eq:subset-case-ii} holds.

Now letting $\mfk{p}^\delta_T \equiv \mathbb{P}_{\theta_0}\left(J(\tilde\theta_0)\leq \bar{J}^{\delta}_{T,r}(\widehat\theta_T)\right) - \mathbb{P}_{\theta_0}\left(J(\tilde\theta_0) \leq \bar{H}_{T,r}(\widehat\theta_T)\right)$, observe that by \eqref{eq:subset-case-ii} and the choices of $\kappa$ and $ \bar{\Theta}_\vep$, we have for $T \geq T_{0,\vep}\vee \tilde{T}_{0,\vep}$
\begin{align*}
\mfk{p}^\delta_T \leq &\ \mathbb{P}_{\theta_0}\left(\widehat \theta_T \in \lf\{\theta':J(\tilde\theta_0)\leq \bar{J}^{\delta}_{T,r}(\theta')\ri\}\cap \bar{\Theta}_\vep\right) -\mathbb{P}_{\theta_0}\left(J(\tilde\theta_0) \leq \bar{H}_{T,r}(\widehat\theta_T)\right) \\
&\ + \PP_{\theta_0}(\widehat\theta_T \in \bar{\Theta}_\vep^c) \ \leq\ 0+\vep =\vep.
\end{align*}
In other words, $\limsup_{T\to\infty} \mfk{p}^\delta_T \leq \vep,$ and since $\vep$ is arbitrary, the assertion follows.\\

\end{proof}

\section{Model classes}\label{sec:models}

This section presents several specific parametric and non-parametric model classes in addition to the Ornstein-Uhlenbeck process presented in Example~\ref{ex:OU:part1}. We show that our main results (Theorems~\ref{thm:minimal:interval} and \ref{th:prop-J}) are applicable and in particular how the confidence interval \eqref{eq:optJ-mod} can be efficiently computed.

\subsection{Continuous state i.i.d.~model with unknown mean}\label{ssec:cts:state:unknown:mean}

Consider a stochastic process $X=\{X_t\}_{t=1}^T$ consisting of $\statespace =\R^n$-valued random variables that are serially independent. 
Each random variable $X_t$ is distributed according to a distribution $F_\theta$ under $\mathbb{P}_\theta$, where $F_\theta$ is known apart from its mean, which is modelled as unknown parameter $\theta\in\Theta\subset\mathbb{R}^d$. A natural (and strongly consistent) estimator for the mean $\theta$ is its empirical counterpart
\begin{align}\label{eq:est:unknown:mean}
    \widehat \theta_T = \frac{1}{T} \sum_{t=1}^T X_t.
\end{align}
In this setting, Assumption~\ref{assum:para} clearly holds with $\esttvs=\mathbb{R}^n$.
A Cram\'er-like theorem states that the estimator \eqref{eq:est:unknown:mean} satisfies a moderate deviation principle with a quadratic rate function.
\begin{theorem}[{Moderate deviations of empirical average in i.i.d.~setting \cite[Theorem~3.7.1]{dembo2009large}}] \label{thm:MDP:cramer}
Let $X_1,\hdots,X_T$ be i.i.d.~$\statespace$-valued random variables, such that $\log \mb E[e^{{\lambda\tpose} X_1}] <\infty$ in some ball around $\lambda=0$, $\mb E[X_1]=\theta$, and let $C(\theta)$, the covariance matrix of $X_1$ be invertible. 
Consider a sequence $\{a_T\}$ such that $a_T\to\infty$ and $b_T \equiv T/a^2_T\to \infty$ and let $\widehat \theta_T$ be the empirical mean~\eqref{eq:est:unknown:mean}. Then, for every $\theta\in\Theta$ the stochastic process $\{a_T( \widehat\theta_T-\theta) \}$ satisfies an LDP with speed $\{b_T\}$ and rate function $I_\theta^M:\mathbb{R}^n\to[0,\infty]$ defined by $I^M_\theta(\vart)=\frac{1}{2}{\vart}\tpose C(\theta)^{-1}\vart$.
\end{theorem}
The rate function is characterized through the covariance matrix of the random variable $X_1$ and a list of common distributions and corresponding rate functions is provided in Table~\ref{tab:rate_functions}. From Table~\ref{tab:rate_functions} and by Theorem~\ref{thm:MDP:cramer}, it can be seen that Assumption~\ref{assum:basic} holds for all candidate distributions in the table. Consequently the optimal CI~\eqref{eq:optJ} proposed by Theorem~\ref{thm:minimal:interval} are applicable (with $\delta=0$).
As each distribution leads to a different covariance and hence a different rate function, the computation of the confidence intervals via the optimization problems \eqref{eq:optJ} needs to be discussed on a case by case basis. 
For simple loss functions, the resulting confidence intervals admit explicit closed-form expressions. Moreover, in the multivariate Gaussian setting, if $J(\theta)$ is of the form
\begin{equation*}
    J(\theta) = \mathbb{E}_{\theta}[\ell(X)],
\end{equation*}
where $\ell:\mathbb{R}^n \to \mathbb{R}$ is an affine function, then the optimal interval obtained from \eqref{eq:optJ} also admits a closed-form representation. Explicit formulas are provided in Appendix~\ref{app:closed-form-ci}.
\begin{table} {{
\centering
\begin{tabular}{c@{\;}l|cccl}
 \hline\hline
 & Distribution $F_\theta$ & Model $\theta$ & Variance $C(\theta)$& Rate function $I^M_\theta(\vart)$ \\
 \hline  &&&&&\\[-2ex]
 (a) & Normal, $\mc N(\mu,\Sigma)$& $\theta=\mu\in\R^n$ &$\Sigma$ & $\frac{1}{2}\vart\tpose \Sigma^{-1}\vart$\\[1ex]
 (b) & Exponential, Exp$(\lambda)$ & $\theta=\frac{1}{\lambda}\in\R_+$ & $\theta^2$ & $\frac{\vart^2}{2\theta^2}$\\[1ex]
 (c) & Gamma, Gam$(p,\nu)$ & $\theta=p\nu\in\R_+$ & $\theta \nu$ & $\frac{\vart^2}{2\nu\theta}$ \\[1ex]
 (d) & Poisson, Poi$(\lambda)$ & $\theta=\lambda\in\R_+$ & $\theta$ & $\frac{\vart^2}{2\theta}$ \\[1ex]
 (e) & Bernoulli, Ber$(p)$ & $\theta=p \in(0,1)$ & $\theta(1-\theta)$ & $\frac{\vart^2}{2\theta(1-\theta)}$ \\[1ex]
 (f) & Binomial, Bin$(m,p)$ & $\theta=mp\in(0,m)$ & $\theta(1-\frac{\theta}{m})$ & $\frac{\vart^2}{2\theta(1-\frac{\theta}{m})}$ \\[1ex]
 (g) & Geometric, Geom$(p)$ & $\theta=\frac{1}{p}\in(1,\infty)$ & $\theta(\theta-1)$ & $\frac{\vart^2}{2\theta(\theta-1)}$\\
 \hline
 \end{tabular}
\caption{The rate functions described by Theorem~\ref{thm:MDP:cramer} for different distributions: (a)~a normal distribution with mean~$\mu$ and covariance matrix~$\Sigma\in\mb S_+^n$; (b)~an exponential distribution with rate parameter~$\lambda$, $\lambda>0$; (c)~a Gamma distribution with shape parameter $p>0$ and scale parameter $\nu>0$;
(d)~a Poisson distribution with rate parameter~$\lambda$, $\lambda>0$; (e)~the Bernoulli distribution with success probability $p$; (f)~a binomial distribution with parameters $m$ and $p$; (g)~the geometric distribution on $\mathbb{N}$ with success probability $p$.}
\label{tab:rate_functions} }}
\end{table}

\subsection{Cox-Ingersoll-Ross process}
Consider a Cox–Ingersoll–Ross (CIR) process defined as the SDE
\begin{equation}\label{eq:CIR:process}
    \d X_t = (\delta - \theta X_t)\d t + \s \sqrt{X_t} \ \d W_t, \quad X_0 = 0,
\end{equation}
where $\delta>0$, $\s>0$ are known parameters and $\theta\in\Theta=\mathbb{R}_{>0}$ is unknown.
The goal is to build an estimator for the stationary variance of the process described as
\begin{align*}
\lim_{t\to\infty}\EE[X_t]= \delta/\theta, \qquad
\lim_{t\to\infty}\text{Var}[X_t]= \  \f{\s^2\delta}{2\theta^2} \equiv J(\theta).
\end{align*}
The confidence interval for the asymptotic variance $J(\theta)$ is built via the MLE for the unknown parameter $\theta$ given by
\begin{equation*}
    \widehat \theta_T = \frac{\delta T-X_T}{\int_0^T X_t \d t},
\end{equation*}
where $\widehat\theta_T\in\esttvs = \mathbb{R}$. In this setting, clearly Assumption~\ref{assum:para} holds.
To derive the optimal CI, according to \eqref{eq:optJ}, note in a first step that Assumption~\ref{assum:basic}, also holds. In particular, the MLE $\widehat\theta_T$ satisfies an MDP with a particularly simple rate function.
\begin{theorem}\label{thm:MDP:CIR}
Let $\{a_T\}$ be such that $a_T \rt \infty$, and $b_T \equiv T/a_T^2 \rt \infty$. For any $\theta\in\Theta$ the stochastic process $\{a_T (\widehat\theta_T - \theta)\}$ satisfies an LDP with speed $b_T$ and rate function $I_\theta^M:\esttvs\to[0,\infty]$ defined as
\begin{equation*}
    I^M_\theta(\vartheta) =  \frac{\vartheta^2\delta}{2\sigma^2 \theta}.
\end{equation*}
\end{theorem}
\begin{proof}
We introduce a change of variables by defining $Z_t = \frac{4}{\sigma^2}X_t$ and $\Delta \equiv \frac{4\delta}{\sigma^2}$,
and note that in the new coordinates, the CIR process~\eqref{eq:CIR:process} can be written as
\begin{equation*}
    \mathsf{d}Z_t = \left( \Delta - \theta Z_t \right)\mathsf{d}t + 2 \sqrt{Z_t} \mathsf{d}W_t.
\end{equation*}
Equivalently, with $\theta'=-\theta/2$, this is
\begin{equation*}
    \mathsf{d}Z_t = \left( \Delta + 2\theta' Z_t \right)\mathsf{d}t + 2 \sqrt{Z_t} \mathsf{d}W_t.
\end{equation*}
In this form, it has been shown in \cite{ref:Gao-09}\footnote{There is an algebraic error in \cite[Theorem~1.2]{ref:Gao-09}: the factor $\delta$ (equivalently $\Delta=4\delta/\sigma^2$ after the rescaling above) is printed in the denominator of the displayed rate function, whereas it should appear in the numerator. Accordingly, the correct MDP rate function for the transformed estimator is
\[
    I^M_{\theta'}(\vartheta)
    =
    -\frac{\Delta \vartheta^2}{4\theta'}
    =
    \frac{2\delta}{\sigma^2\theta}\vartheta^2.
\]
} that the estimator
\begin{equation*}
    \widehat\theta'_T = \frac{Z_T-\Delta T}{2\int_0^T Z_t \mathsf{d}t}
\end{equation*}
converges to $\theta'=-\frac{1}{2}\theta$ and satisfies an MDP with rate function
\begin{equation*}
    I^M_{\theta'}(\vartheta)
    =
    -\frac{\Delta \vartheta^2}{4\theta'}
    =
    \frac{\vartheta^2 2 \delta}{\sigma^2\theta}.
\end{equation*}
By applying the change of coordinate again, we observe that
\begin{equation*}
    -2\widehat\theta'_T
    =
    \frac{\Delta T-Z_T}{\int_0^T Z_t \mathsf{d}t}
    =
    \frac{\delta T-X_T}{\int_0^T X_t \mathsf{d}t}
    =
    \widehat \theta_T.
\end{equation*}
The contraction principle states that $\widehat \theta_T$ satisfies an MDP with rate function given as
\begin{equation*}
    I^M_\theta(\vartheta)
    =
    I^M_{\theta'}\left(-\vartheta/2\right)
    =
    \frac{\vartheta^2\delta}{2\sigma^2 \theta},
\end{equation*}
which completes the proof.
\end{proof}
Theorem~\ref{thm:MDP:CIR} ensures that Assumption~\ref{assum:basic} is satisfied and accordingly Theorem~\ref{thm:minimal:interval} can be applied (with $\delta=0$) quantifying an optimal interval estimator for $J(\theta)$. The corresponding interval~\eqref{eq:optJ} is expressed as
\begin{subequations} \label{eq:CIR:opt:problems}
\begin{equation}\label{eq:CIR:upper:bound}
\begin{aligned}
  \bar {J}^0_{T,r}(\widehat\theta_T)  &= 
  \max\limits_{\theta>0} \left\{\frac{\delta\sigma^2}{2\theta^2} \ : \ \frac{\delta (\widehat\theta_T - \theta)^2}{2 \sigma^2\theta} \leq r_T \right\} \\
  &=\max\limits_{\bar \theta>0}\left\{ \frac{\delta\sigma^2}{2}\bar\theta^2 \ : \ \bar\theta^2 \widehat\theta_T^2 + \bar \theta(- 2 \widehat\theta_T -  \frac{2 \sigma^2 r_T}{\delta}) + 1  \leq 0\right\}
\end{aligned} 
\end{equation}
and 
\begin{align}\label{eq:CIR:lower:bound}
   \ubar {J}^0_{T,r}(\widehat\theta_T)  &=\min\limits_{\bar \theta>0}\left\{ \frac{\delta\sigma^2}{2}\bar\theta^2 \ : \ \bar\theta^2 \widehat\theta_T^2 + \bar \theta(- 2 \widehat\theta_T -  \frac{2 \sigma^2 r_T}{\delta}) + 1  \leq 0\right\},
\end{align}
\end{subequations}
where $r_T = r/a_T^2$.
 The optimization problems~\eqref{eq:CIR:opt:problems} describe the maximization (resp.~minimization) of a quadratic function subject to a quadratic constraint. While these problems in general are hard to solve, due to the explicit structure and the low dimension, the problems~\eqref{eq:CIR:opt:problems} admits a closed form solution. Note that if $\widehat \theta_T <0$, the optimization problems~\eqref{eq:CIR:opt:problems}, might be infeasible. This, however, becomes more and more unlikely as $T$ grows, since $\widehat\theta_T \to \theta>0$.
\begin{proposition}[Closed-form solution of
\eqref{eq:CIR:opt:problems}]
\label{prop:CIR:SDP}
Let $\theta'>0$ and define
\[
    q_T \equiv \frac{\sigma^2 r_T}{\delta}.
\]
Then the optimization problems~\eqref{eq:CIR:opt:problems} admit the
closed-form solutions
\begin{align*}
    \ubar J^0_{T,r}(\theta')
    =
    \frac{\delta\sigma^2}{
    2\left(
        \theta'+q_T+\sqrt{q_T(2\theta'+q_T)}
    \right)^2},
    \quad
    \bar J^0_{T,r}(\theta')
    =
    \frac{\delta\sigma^2}{
    2\left(
        \theta'+q_T-\sqrt{q_T(2\theta'+q_T)}
    \right)^2}.
\end{align*}
\end{proposition}
The proof of Proposition~\ref{prop:CIR:SDP} follows directly by noting that the constraints in~\eqref{eq:CIR:opt:problems} are equivalent to
\[
    \theta^2-2(\theta'+q_T)\theta+(\theta')^2\leq0.
\] 
The result then follows by evaluating the decreasing function
$J(\theta)=\delta\sigma^2/(2\theta^2)$ at the two roots.
Figure~\ref{sfig:CIR:new} visualizes the optimal CI described by Theorem~\ref{thm:minimal:interval} via \eqref{eq:CIR:opt:problems}, where we exploit the closed form solution from Proposition~\ref{prop:CIR:SDP}.

\begin{figure}[t] 
\centering
{\begin{tikzpicture}
\begin{axis}[
  width=0.5\linewidth,
  height=0.3\linewidth,
  scale only axis,
  xmode=log,
  log basis x=10,
  xmin=100,
  xmax=20000,
  xtick={100,1000,10000,20000},
  xticklabels={$10^2$,$10^3$,$10^4$,$2\!\cdot\!10^4$},
  minor x tick num=8,
  ymin=7.04123202471,
  ymax=13.6509971043,
  xlabel={$T$},
  ylabel={confidence bounds},
  tick align=outside,
  grid=major,
  grid style={draw=black!10},
  axis line style={black!65},
  legend style={at={(0.5,-0.23)},anchor=north,legend columns=3,
    draw=black!35,fill=white,font=\small,
    /tikz/every even column/.append style={column sep=7pt}}
]

\addplot[draw=none,fill=red,fill opacity=0.18,forget plot]
table[row sep=crcr] {
x	y\\
100	8.44045441867\\
142.364595899	8.74642353751\\
202.676781654	9.04327143163\\
288.539981181	9.21062955901\\
410.778778215	9.36833145351\\
584.803547643	9.57874716209\\
832.553207402	9.66660627808\\
1185.26100936	9.70785733206\\
1687.39204632	9.78473244037\\
2402.24886796	9.84515596505\\
3419.95189335	9.88585691495\\
4868.8006929	9.93178212673\\
6931.44843155	9.95057307675\\
9867.9285495	9.98015924689\\
14048.4366031	9.99885523057\\
20000	10.009624764\\
20000	10.3442213379\\
14048.4366031	10.3870514745\\
9867.9285495	10.4444460628\\
6931.44843155	10.5334523946\\
4868.8006929	10.6014984749\\
3419.95189335	10.6939861984\\
2402.24886796	10.7989880774\\
1687.39204632	10.9536353658\\
1185.26100936	11.1132668252\\
832.553207402	11.2863712267\\
584.803547643	11.4697937878\\
410.778778215	11.7132880563\\
288.539981181	12.0361611774\\
202.676781654	12.5175235498\\
142.364595899	12.8293643172\\
100	13.3234862219\\
}--cycle;
\addplot[color=red,dashed,line width=1.5pt]
table[row sep=crcr] {
x	y\\
100	10.8070353941\\
142.364595899	10.7084559394\\
202.676781654	10.6800557982\\
288.539981181	10.5786912195\\
410.778778215	10.5295752703\\
584.803547643	10.4695922145\\
832.553207402	10.4428362658\\
1185.26100936	10.3864556815\\
1687.39204632	10.3459362248\\
2402.24886796	10.3137528408\\
3419.95189335	10.2798233717\\
4868.8006929	10.2560017259\\
6931.44843155	10.2320295033\\
9867.9285495	10.2065667086\\
14048.4366031	10.1880327318\\
20000	10.1728076919\\
};
\addlegendentry{$\bar J^0_{T,r}(\widehat\theta_T)$}

\addplot[draw=none,fill=blue,fill opacity=0.18,forget plot]
table[row sep=crcr] {
x	y\\
100	7.36874290703\\
142.364595899	7.72180938452\\
202.676781654	8.06691437896\\
288.539981181	8.2978345191\\
410.778778215	8.51683168608\\
584.803547643	8.77964786484\\
832.553207402	8.92904155032\\
1185.26100936	9.03146816626\\
1687.39204632	9.16189120946\\
2402.24886796	9.27305528739\\
3419.95189335	9.36185624231\\
4868.8006929	9.45167723798\\
6931.44843155	9.51230424055\\
9867.9285495	9.57964462751\\
14048.4366031	9.63342530338\\
20000	9.67660558395\\
20000	9.99728955193\\
14048.4366031	10.003886479\\
9867.9285495	10.0206299128\\
6931.44843155	10.0630487972\\
4868.8006929	10.0808521929\\
3419.95189335	10.1162903693\\
2402.24886796	10.1573372703\\
1687.39204632	10.2372664989\\
1185.26100936	10.3135558813\\
832.553207402	10.3929278877\\
584.803547643	10.4713024418\\
410.778778215	10.5912536604\\
288.539981181	10.7664202736\\
202.676781654	11.0602112205\\
142.364595899	11.187861576\\
100	11.4459028347\\
}--cycle;
\addplot[color=blue,dotted,line width=1.5pt]
table[row sep=crcr] {
x	y\\
100	9.34829973394\\
142.364595899	9.38997024334\\
202.676781654	9.47894969372\\
288.539981181	9.49415891761\\
410.778778215	9.54476624746\\
584.803547643	9.57693945711\\
832.553207402	9.63084264696\\
1185.26100936	9.6507098533\\
1687.39204632	9.6783064708\\
2402.24886796	9.70754961566\\
3419.95189335	9.72969289943\\
4868.8006929	9.75630103923\\
6931.44843155	9.77826667322\\
9867.9285495	9.79470119052\\
14048.4366031	9.81396813583\\
20000	9.83300894943\\
};
\addlegendentry{$\ubar J^0_{T,r}(\widehat\theta_T)$}

\addplot[color=darkgreen,solid,line width=1.1pt]
coordinates {(100,10)
              (20000,10)};
\addlegendentry{$J(\theta)$}

\end{axis}
\end{tikzpicture}%
}
\caption[]{MDP-based confidence bounds of Proposition~\ref{prop:CIR:SDP} for the stationary variance of the
Cox-Ingersoll-Ross process. Shaded regions show the pointwise $10\%$ and
$90\%$ quantiles, while the dashed and dotted curves show the corresponding
empirical means, all based on $10^3$ independent trajectories. We set
$b_T = T^{5/11}$, $\delta=5$, $\theta=1$, $\sigma=2$ and $r=10^{-2}$.}
\label{sfig:CIR:new}
\end{figure}

\subsection{Non-parametric i.i.d.~model}\label{ssec:non-parametric:iid} 
Consider a setting where $\statespace\subset\mathbb{R}^n$ is compact and  where the samples $\{X_t\}_{t\in\N}\subset \statespace$ are serially independent and distributed according to $\theta^\star\in\Theta$, for $\Theta=\mathcal{P}(\statespace)$. The empirical probability distribution is given by
\begin{equation}\label{estimator:iid:finite:state}
    \widehat\theta_T = \frac{1}{T}\sum_{t=1}^T \delta_{X_t},
\end{equation}
where $T\in\mathbb{N}$ and model space is $\Theta=\mathcal{P}(\statespace)$. We take
$\esttvs=\mathcal{M}(\statespace)$ the vector space of finite signed measures on $\statespace$. We equip $\esttvs=\mathcal{M}(\statespace)$ with the Kantorovich-Rubinstein (KR) norm. Note that the topology on $\mathcal{P}(\statespace)$ induced by the KR-norm coincides with the topology of weak convergence \cite[Section 3.2]{Boga18}.
Clearly Assumption~\ref{assum:para} is satisfied.
For any $\theta\in\Theta$, introduce the function $I_\theta^M:\mathcal{M}(\statespace)\to[0,\infty]$ defined as
\begin{equation}\label{eq:rate:function:MDP:iid}
    I_\theta^M(\vartheta) = \begin{cases}
    \frac{1}{2}\int_\statespace \left( \frac{\mathsf{d}\vartheta}{\mathsf{d}\theta} \right)^2 \mathsf{d} \theta, & \quad \vartheta\ll\theta, \ \vartheta(\statespace) = 0    \\
    \infty, &\quad \text{o.w.}
    \end{cases}
\end{equation}
Then \cite[Theorem 1]{DeZa97} (see also \cite{ref:Ermakov-15}) gives the MDP counterpart of Sanov's Theorem in the $\tau$-topology. Moreover, the argument underlying \cite[Theorem 2]{DeZa97} shows that, for the fixed scaling family $\{a_T\}$, $\widehat\theta_T$ satisfies an MDP in the scaling regime determined by $a_T$, in the sense that
$
    a_T(\widehat\theta_T-\theta)
$
satisfies an LDP in the KR-topology, provided
$
    a_T\|\widehat\theta_T-\theta\|_{\rm KR}
    \xrightarrow{\PP_\theta}0.
$
When dimension $n=1$, this condition is automatically satisfied under
$1\ll a_T\ll\sqrt{T}$, since
$\EE_\theta\|\widehat\theta_T-\theta\|_{\rm KR} = O(T^{-1/2})$.

\begin{theorem}[{Moderate deviations empirical measure---i.i.d.~setting}] \label{thm:MDP:sanov:cts}
Let $X_1,\hdots,X_T$ be $\statespace$-valued i.i.d.~random variables. Consider a sequence $\{a_T\}$ such that $a_T\to\infty$, $b_T \equiv T/a^2_T\to \infty$ and $ a_T\|\widehat\theta_T-\theta\|_{\rm KR}  \xrightarrow{\PP_\theta}0
$. Let the estimator $\{\widehat\theta_T\}$ be defined as in \eqref{estimator:iid:finite:state}.
Then, the stochastic process $\{a_T( \widehat\theta_T-\theta) \}$ satisfies an LDP with speed $\{b_T\}$ and convex, good rate function \eqref{eq:rate:function:MDP:iid}.
\end{theorem}
Consider a function $J$ expressing an expected loss
\begin{equation}\label{eq:obj:J}
    J(\theta)\equiv \mathbb{E}_\theta[\ell(X)] = \int_\statespace \ell(x) \mathsf{d}\theta(x),
\end{equation}
where $\ell:\statespace\to\mathbb{R}$ is some Lipschitz-continuous loss function. Note that $J$ is uniformly continuous and
Theorem~\ref{thm:MDP:sanov:cts} ensures that Assumption~\ref{assum:basic} holds. Therefore, Theorem~\ref{thm:minimal:interval} proposes to estimate $J(\theta)$ via the interval family $\{\mathcal{I}^{\delta,\star}_{T,r}=[\ubar J^\delta_{T,r},\bar J^\delta_{T,r}]:T>0\}$ given by \eqref{eq:optJ-mod}. In this example, we need to consider $\delta>0$ as the rate function \eqref{eq:rate:function:MDP:iid} is not continuous. The functions $\ubar J^\delta_{T,r}$ and $\bar J^\delta_{T,r}$ according to \eqref{eq:optJ-mod} are described by convex optimization problems that are infinite-dimensional and therefore cannot be readily solved with a computer.
The functions $\ubar J^\delta_{T,r}$ and $\bar J^\delta_{T,r}$ for small $\delta$ can be closely approximated by $\ubar J^0_{T,r}$ and $\bar J^0_{T,r}$. Interestingly, we show in the following, that the functions $\ubar J^0_{T,r}$ and $\bar J^0_{T,r}$ can be equivalently expressed as tractable finite-dimensional optimization problems by exploiting duality of convex optimization~\cite{barvinok2002course}.

\begin{proposition}[Computation of $\ubar J^0_{T,r}$ and $\bar J^0_{T,r}$] \label{prop:comp:iid}
Suppose that $r>0$, and let $\bar \ell = \sup_{x\in \statespace}\ell(x)$, $\ubar \ell = \sup_{x\in \statespace}\{-\ell(x)\}$ 
denote the global suprema of the loss and its negative. Then the two functions $\ubar J^0_{T,r}$ and $\bar J^0_{T,r}$ given by \eqref{eq:optJ} admit the dual representations
\begin{align*}
\ubar J^0_{T,r}({\widehat\theta_T}) &= -\min_{\alpha\in\mc A}\left\{ \alpha - \frac{1}{2r_T+1} \left( \frac{1}{T}\sum_{t=1}^T \sqrt{\alpha+\ell(X_t)}\right)^2 \right\},\\
\bar J^0_{T,r}({\widehat\theta_T}) &= \min_{\alpha\in\mc B}\left\{ \alpha - \frac{1}{2r_T+1} \left(\frac{1}{T}\sum_{t=1}^T \sqrt{\alpha-\ell(X_t)}\right)^2 \right\},
\end{align*}
where
$\mc A
=\{\alpha\in\R:
\ubar\ell\leq\alpha\leq
\frac{2r_T+1}{2r_T}\ubar\ell
+\frac{1}{2r_T}J({\widehat\theta_T})\}$,
$\mc B
=\{\alpha\in\R:
\bar\ell\leq\alpha\leq
\frac{2r_T+1}{2r_T}\bar\ell
-\frac{1}{2r_T}J({\widehat\theta_T})\}$
and $r_T=r/a_T^2$.
\end{proposition}

\begin{proof}
The duality formula is shown for the upper bound $\bar J^0_{T,r}$ as the lower bound $\ubar J^0_{T,r}$ is derived analogously.
The argument closely follows that of \cite[Proposition~2]{ref:vanParys:fromdata-17}, with the key difference that we apply \cite{ref:BenTal-13} to the $\chi^2$-distance rather than to the Burg entropy.
To start, let $P=\widehat\theta_T$ and consider the lower-semicontinuous extension to $[0,\infty)$ of
\[
    \phi(t)=\frac{(t-1)^2}{t}, \qquad t>0,
\]
so that $\phi(0)=+\infty$. Its recession slope is
\[
    \phi'_{\infty}:=\lim_{t\to\infty}\frac{\phi(t)}{t}=1.
\]
If the Lebesgue decomposition of $\theta\in\Theta=\mathcal P(S)$ with respect to $P$ is $\theta=fP+\theta^\perp$, then the corresponding extended $\phi$-divergence is
\begin{equation}\label{eq:extended:phi:divergence}
    \mathsf D_\phi(\theta,P)
    :=\int_S\phi(f)\,dP+\phi'_{\infty}\theta^\perp(S)
    =\int_S\phi(f)\,dP+\theta^\perp(S).
\end{equation}
With this convention,
\[
    \mathsf D_\phi(\theta,P)
    =
    \begin{cases}
    \displaystyle\int_S\left(\frac{dP}{d\theta}-1\right)^2d\theta,
        & P\ll\theta,\\[1.2ex]
    +\infty, & \text{otherwise}.
    \end{cases}
\]
Hence, by the form of the rate function~\eqref{eq:rate:function:MDP:iid},
\[
    I^M_\theta\bigl(a_T(P-\theta)\bigr)
    =\frac{a_T^2}{2}\mathsf D_\phi(\theta,P).
\]
The optimization problem~\eqref{eq:optJ} can therefore be expressed as
\begin{equation}\label{eq:program:phi:divergence}
    \bar J^0_{T,r}({\widehat\theta_T})
    =\sup_{\theta\in\Theta}
    \left\{\mathbb E_\theta[\ell(X)]:
    \mathsf D_\phi(\theta,\widehat\theta_T)\leq 2r_T\right\}.
\end{equation}
The conjugate of $\phi$ is
\[
    \phi^*(s)=
    \begin{cases}
        2-2\sqrt{1-s}, & s\leq 1,\\
        +\infty, & s>1.
    \end{cases}
\]
Since $P$ has finite support, the duality step can be reduced to a finite-dimensional convex program: consolidate the distinct atoms of $P$, use the finitely many values of $f$ at those atoms, and let $m=\theta^\perp(S)$ denote the singular mass. For fixed $m$, the supremal contribution of the singular component is $m$ times the supremum of $\ell$ over the $P$-null part of $S$. Moreover, $f\equiv1$ and $m=0$ is strictly feasible because $r_T>0$ and $\mathsf D_\phi(P,P)=0<2r_T$. Strong duality therefore gives
\begin{align}
\bar J^0_{T,r}({\widehat\theta_T})
={}&\inf_{\substack{\lambda\geq0,\,\eta\in\R\\
                     \eta+\lambda\geq\bar\ell}}
\left\{
\eta+2r_T\lambda
+\mathbb E_{\widehat\theta_T}\!\left[
\lambda\phi^*\!\left(\frac{\ell(X)-\eta}{\lambda}\right)
\right]
\right\}. \label{eq:correct:phi:dual}
\end{align}
At $\lambda=0$, the last term is understood as the closed perspective of $\phi^*$. The constraint $\eta+\lambda\geq\bar\ell$ is essential. The domain of $\phi^*$ requires $\ell(x)-\eta\leq\lambda$ at the atoms of $P$, whereas the singular term in \eqref{eq:extended:phi:divergence}, together with $\phi'_{\infty}=1$, imposes the same condition at $P$-null locations. Together, these conditions require
\[
    \ell(x)-\eta\leq\lambda
    \qquad\text{for all }x\in S.
\]
Using $\widehat\theta_T=T^{-1}\sum_{t=1}^T\delta_{X_t}$ and the variable transformation $\alpha=\lambda+\eta$ and $\nu=\lambda$, equation~\eqref{eq:correct:phi:dual} becomes
\begin{align*}
\bar J^0_{T,r}({\widehat\theta_T})
=\inf_{\substack{\alpha\geq\bar\ell\\\nu\geq0}}
\left\{
-2\sqrt{\nu}\frac{1}{T}\sum_{t=1}^T\sqrt{\alpha-\ell(X_t)}
+\alpha+\nu(2r_T+1)
\right\}.
\end{align*}
Denote the objective above by $g(\alpha,\nu)$. For $\nu=0$ and $\alpha\geq\bar\ell$, the closed-perspective term is zero, so $g(\alpha,0)=\alpha$. For fixed $\alpha\geq\bar\ell$, minimization over $\nu\geq0$ gives
\[
    \nu^\star(\alpha)
    =\frac{1}{(2r_T+1)^2}
    \left(\frac{1}{T}\sum_{t=1}^T
    \sqrt{\alpha-\ell(X_t)}\right)^2.
\]
This formula also covers the boundary case in which the sum is zero, in which case $\nu^\star(\alpha)=0$. Consequently,
\begin{equation}\label{eq:g:optimal:nu}
g(\alpha,\nu^\star(\alpha))
=\alpha-\frac{1}{2r_T+1}
\left(\frac{1}{T}\sum_{t=1}^T
\sqrt{\alpha-\ell(X_t)}\right)^2.
\end{equation}
It remains to restrict the minimization over $\alpha$ to the compact set $\mc B$. By Jensen's inequality,
\begin{align}
g(\alpha,\nu^\star(\alpha))
&\geq
\alpha-\frac{1}{2r_T+1}
\left(\frac{1}{T}\sum_{t=1}^T
(\alpha-\ell(X_t))\right) \notag\\
&=\frac{2r_T\alpha+J(\widehat\theta_T)}{2r_T+1}.
\label{eq:lower:bound:g:jensen}
\end{align}
The scalar objective in \eqref{eq:g:optimal:nu} is continuous. Since $r_T>0$, the lower bound in \eqref{eq:lower:bound:g:jensen} tends to $+\infty$ as $\alpha\to\infty$, and hence a minimizer $\alpha^\star\geq\bar\ell$ exists. Moreover, equation~\eqref{eq:g:optimal:nu} gives
\begin{equation}\label{eq:upper:bound:g}
g(\alpha^\star,\nu^\star(\alpha^\star))
\leq g(\bar\ell,\nu^\star(\bar\ell))
\leq\bar\ell.
\end{equation}
Combining this inequality with \eqref{eq:lower:bound:g:jensen} at $\alpha=\alpha^\star$ yields
\[
    \alpha^\star
    \leq
    \frac{2r_T+1}{2r_T}\bar\ell
    -\frac{1}{2r_T}J(\widehat\theta_T).
\]
Thus the minimization may be restricted to $\mc B$, which proves the representation for $\bar J^0_{T,r}$.
Finally, applying the upper-bound representation to the loss $-\ell$, whose global supremum is $\ubar\ell$ and whose empirical mean is $-J(\widehat\theta_T)$, gives the stated representation over $\mc A$ for $\ubar J^0_{T,r}$. This completes the proof.
\end{proof}

\begin{remark}[Finite state setting---comparison to \cite{ref:Amine-21}]
In the special setting where the random variables are finitely supported, i.e., $\statespace=\{1,\dots,d\}$, the upper confidence bound \eqref{eq:optJ} reduces to
\begin{align*}
    \bar J^0_{T,r}(\widehat \theta_T) = \left\{ \begin{array}{ll}
      \sup\limits_{\theta\in\Theta}   &  J(\theta) \\
       \st  & \frac{1}{2} \sum_{i=1}^d\frac{(\widehat \theta_T-\theta)_i^2}{\theta_i}\leq \frac{r}{a_T^2}.
    \end{array}
    \right. 
\end{align*}
In the finite state i.i.d.~setting, the recent work \cite{ref:Amine-21} considers a slightly modified optimality criteria. Namely, a notion similar to that of eventually smaller (see Definition~\ref{def:smaller}), where a family of functions $\{\phi_T: T>0\} \subset \SC{F}(\Theta, \R)$ is defined to be ''eventually smaller" than another family $\{\psi_T: T>0\} \subset \SC{F}(\Theta, \R)$ if
\begin{equation*}
    \limsup_{T\to\infty} \left| \frac{\phi_T(\theta) - f(\theta)}{\psi_T(\theta) - f(\theta)} \right| \leq 1 \quad \forall \theta\in\Theta,
\end{equation*}
for some fixed function $f:\Theta\to\mathbb{R}$. Under this modified optimality criteria, \cite{ref:Amine-21} shows that the optimal upper confidence bound is a modified DRO predictor
\begin{align*}
     \left\{ \begin{array}{ll}
      \sup\limits_{\theta\in\Theta}   &  J(\theta) \\
       \st  & \frac{1}{2} \sum_{i=1}^d\frac{(\widehat \theta_T-\theta)_i^2}{(\widehat \theta_T)_i}\leq \frac{r}{a_T^2}.
    \end{array}
    \right. 
\end{align*}
\end{remark}

\begin{remark}[Stochastic Programming]\label{Rem:SP}
The setting here can actually be generalized, where instead of estimating $J(\theta)$ of the form \eqref{eq:obj:J}, we consider a general stochastic program, i.e., $J(\theta)$ given by \eqref{eq:stochastic:program}, where we denote the loss function as $L$
\begin{equation*}
    J(\theta) = \min_{z\in\mathcal{Z}}\EE_\theta[L(X,z)].
\end{equation*}
Then under appropriate regularity assumptions, the corresponding interval family is
\begin{align*}
    \bar J_{T,r}^0(\theta') &= \sup_{\theta\in\Theta} \bigg\{\min_{z\in\mathcal{Z}} \EE_\theta[L(X,z)] : I_\theta^M(a_T(\theta'-\theta)) \leq r \bigg\} \\
    &=\min_{z\in\mathcal{Z}} \sup_{\theta\in\Theta} \{ \EE_\theta[L(X,z)] : I_\theta^M(a_T(\theta'-\theta)) \leq r\} \\
    &=\min_{z\in\mathcal{Z}}\bar J^0_{T,r}(\theta',z),
\end{align*}
where for each $z$, the term $\bar J^0_{T,r}(\theta',z)$ can be expressed via the duality formula provided by Proposition~\ref{prop:comp:iid}, with $\ell(x)=L(x,z)$. The lower confidence bound $\ubar J^0_{T,r}(\theta')$ follows analogously.
\end{remark}

\begin{remark}[On tractability of the optimal confidence intervals]
\label{rem:tractability}
The optimization problems defining the optimal interval bound functions
$\bar J^\delta_{T,r}$ and $\ubar J^\delta_{T,r}$ in \eqref{eq:optJ-mod}
are, in general, infinite-dimensional and need not admit a tractable reformulation.

The examples in this section illustrate that tractability can be obtained
when additional structure is present. In particular, one can expect a
tractable formulation when either (i) the parameter $\theta$ is
finite-dimensional, or (ii) the infinite-dimensional problem admits an
exact finite-dimensional reduction (for example, via duality), and when,
in addition, the sublevel sets induced by the MDP rate function are
convex and explicitly representable (e.g., ellipsoids or divergence-based
uncertainty sets), and the functional $J(\theta)$ is compatible with this
structure (for instance, affine).

Outside of such settings, the resulting optimization problems may be
inherently nonconvex or infinite-dimensional, and a tractable reformulation is generally not available.
\end{remark}

\begin{remark}[Choice of the speed parameter $b_T$]

For many estimators, a moderate deviation principle holds for a range of sequences $1 \ll b_T \ll T$, so $b_T$ should be viewed as a design parameter controlling the strength of the exponential accuracy requirement. 

For fixed $b_T$ and $r>0$, the resulting intervals achieve error probability of order $\exp(-r b_T)$, with larger $b_T$ yielding stronger guarantees but wider intervals, and smaller $b_T$ leading to tighter intervals but weaker guarantees.

Importantly, the optimality properties (minimality and eventual UMA) are defined relative to a fixed sequence $b_T$: minimality compares our intervals only to procedures satisfying exponential accuracy with respect to the same $b_T$, and does not extend across different choices of the speed.
\end{remark}

\section{Concluding reflections on decision making}\label{sec:conclusion}

As already indicated by \eqref{eq:stochastic:program} and Remark~\ref{Rem:SP}, the framework proposed in this article can be extended to tackle data-driven decision making problems, more precisely to estimate the optimal value $J(\theta)$ of a stochastic programming problem. With slight abuse of notation, we denote
\begin{equation}\label{eq:SP:conc}
    J(\theta) = \min_{z\in\mathcal{Z}} J(\theta,z) = \min_{z\in\mathcal{Z}}\mathbb{E}_\theta[\ell(X,z)],
\end{equation}
for $\mathcal{Z}\subset \mathbb{R}^m$. According to Remark~\ref{Rem:SP}, our paper directly proposes optimal confidence intervals for $J(\theta)$. For future work, we aim to study the derivation of a pair comprising a decision $\widehat z_T\in\mathcal{Z}$ and a corresponding CI $\widehat{\mathcal{I}}_T$ with the goal that $J(\theta,\widehat z_T)$ is ``close to" the optimal value $J(\theta)$ and $\widehat{\mathcal{I}}_T$ provides an exponentially accurate estimate for $J(\theta,\widehat z_T)$, i.e.,
\begin{equation*}
    \limsup_{T\to\infty}\frac{1}{b_T} \log \mathbb{P}_\theta\left(J(\theta,\widehat z_T)\notin \widehat{\mathcal{I}}_T\right) \leq -r \quad \forall \theta.
\end{equation*}
In addition, the CI $\widehat{\mathcal{I}}_T$ must admit the other optimality criteria \ref{eq:item:1:conf:minimality}-\ref{eq:item:3:conf:UMA} studied in this paper. Stochastic programming problems~\eqref{eq:SP:conc} are the key objects in supervised learning, and statistical learning theory~\cite{vapnik1998statistical, ref:Tibshirani-01} aims to quantify estimation guarantees for the corresponding optimizers. 
Constructing optimal confidence intervals via MDP rate functions in the context of supervised learning problems has not been considered yet. This paper provides the methods and techniques for this goal.   

Another class of problems that can be addressed as instances of \eqref{eq:SP:conc} are offline (or batch) reinforcement learning problems \cite{ref:BerTsi-96,ref:Sutton-1998,ref:Levine-20}, where previously collected data from an unknown dynamical system is available and where one aims to answer the question of how this data can be utilized to estimate an optimal (or approximately optimal) policy. 
Offline reinforcement learning is difficult as it requires the learning algorithm to derive a sufficient understanding of the underlying unknown dynamical system entirely from a fixed data set, i.e., without the possibility to interact further with the underlying dynamical system. When deriving confidence intervals for the learned offline policy, the key challenge is to systematically find a balance between increased generalization and avoiding unwanted
behaviors outside of the distribution, i.e., the available data.

\begin{appendix}
\section{Some definitions}\label{appA:MDP:basics}

For the following definitions, we will consider a generic stochastic process, $\Psi = \{\Psi_T\}$, defined on a probability space $(\Omega, \SC{F},\PP)$.

\begin{definition}[Stochastically bounded]
\label{def:stoch-bdd}
A stochastic process $\Psi = \{\Psi_T\}$ taking values in a metric space $\estsp$ is \emph{stochastically bounded} if 
$$\lim_{R \rt \infty} \limsup_{T \rt \infty} \PP(\Psi_T \notin B(w_0,R)) =0,$$
where $w_0$ is an arbitrary fixed reference point in $\estsp$.
\end{definition}

It is clear that the above definition is independent of the reference point $w_0$. If $\estsp$ is finite-dimensional, the notions of stochastic boundedness and tightness coincide.

\begin{definition}[Large deviation principle]
\label{def:LDP}
 A stochastic process $\Psi = \{\Psi_T\}$ taking values in a metric space $\estsp$ satisfies an LDP with a rate function $I^L:\estsp\to[0,\infty]$ and speed $\{b_T\}$ where $\lim_{T\to\infty}b_T = \infty$, if for all  Borel sets $\mc D \subseteq \estsp$ we have
\begin{equation}
\label{eq:ldp_exponential_rates:old}
\begin{aligned}	
	-\inf_{s \in {\mc D}^\circ} \, I^L(s)~& \leq \liminf_{T\to \infty}~\frac{1}{b_T} \log \mb P\lf( \Psi_T \in \mc D \ri) \\
	& \leq \limsup_{T\to \infty}~\frac{1}{b_T} \log \mb P\lf( \Psi_T \in \mc D \ri) \leq -\inf_{s \in \overline{\mc D}} \,  I^L(s).
\end{aligned}
\end{equation}
\end{definition}

The rate function can be taken to be l.s.c., and under this assumption it is unique. A rate function $I^L$ associated with an LDP is called {\em good} if its sublevel set, $\{w \in \estsp: I^L(w) \leq a\}$, is compact for each $a\geq 0$. The goodness property of rate function holds for large deviation asymptotics in a wide variety of models. \cite{dembo2009large} is a comprehensive source for the theory of large deviations.

\begin{definition}[Moderate deviation principle]
\label{def:MDP}
Let $\Psi = \{\Psi_T\}$ be a stochastic process taking values in a metrizable topological vector space $\esttvs$. Suppose that $\Psi_T \stackrel{\PP}\Rt \psi_0$  as $T\rt \infty$ for some deterministic $\psi_0 \in \esttvs$.  
$\Psi$ satisfies a moderate deviation principle (MDP) with a rate function $I^M:\esttvs\to[0,\infty]$ and speed $b_T$, if  $\{a_T(\Psi_T - \psi_0)\}$ with $a_T  = \sqrt {T/b_T}$ satisfies an LDP in $\esttvs$ with the rate function $I^M$ and speed $\{b_T\}$.
\end{definition}

\section{Auxiliary Proofs}\label{appB}

\begin{proof}[Justification of Remark~\ref{rem:ass:basic}]
To justify the second assertion of Remark~\ref{rem:ass:basic}, we proceed in two steps. \\
    {\em Only if part:}
Assume that condition~\ref{item:ass:cpt:RF} holds and consider a sequence $\{(\theta_n, \vart_n)\} \subset \mathbb{G}_{m_0}(\Theta_0).$ 
Since $\Theta_0$ is bounded, $\{\theta_n\}$ is clearly bounded. 
Suppose the sequence $\{\vart_n\}$ is not bounded, then there exists a subsequence $\{n_k\}$ such that $\met{\vart_{n_k}} \stackrel{k\rt \infty} \rt \infty$. But then Condition \ref{item:ass:cpt:RF} guarantees   that $\lim_{k\rt \infty} I^M_{\theta_{n_k}}(\vart_{n_k}) = \infty.$ This violates the fact that $\{(\theta_{n_k}, \vart_{n_k})\} \subset \mathbb{G}_{m_0}(\Theta_0)$ (that is, $I^M_{\theta_{n_k}}(\vart_{n_k}) \leq m_0$ for all $k\in\mathbb{N}$).  

{\em If part:} Suppose that Condition \ref{item:ass:cpt:RF} does not hold. Then there exist $\{\theta^0_n\} \subset \Theta, \{\vart_n\} \subset \esttvs$ such that $\sup_n\met{\theta^0_n} < \infty$ and $\|\vart_n\|  \stackrel{n\rt \infty}\rt \infty$, and for all $n\geq 1$, $I^M_{\theta^0_n}(\vart_n) \leq m_0$ for some constant $m_0>0$. But then it is clear that for the bounded set $\Theta_0 = \{\theta^0_n\}$,   $\mathbb{G}_{m_0}(\Theta_0)$ is not bounded.
    \phantom\qedhere
\end{proof}

{\grn
\begin{proof}[Proof of Lemma~\ref{lem:est-conv-rate}]
We need to show that for any $\kappa >0$,
\begin{align}
    \label{eq:prob-conv-mdp}
    \PP_{\theta_0}\lf[\|a_T(\widehat\theta_T - \theta_0)\| \geq  \kappa\ri] \stackrel{T \rt \infty} \Rt 0.
\end{align}
Now by the LDP upper bound, for any $\ep>0$, there exists $T_0 \equiv T_0(\ep)$ such that for all $T \geq T_0$
\begin{align*}
    \f{1}{b_T} \log \PP_{\theta_0}\lf[\|a_T(\widehat\theta_T - \theta_0)\| \geq \kappa\ri] \leq - I^M_{\theta_0}(\{\|\vart\| \geq \kappa\}) + \ep,
\end{align*}
where $I^M_{\theta_0}(\{\|\vart\| \geq \kappa\}) \dfeq \inf \limits_{\vart: \|\vart\| \geq \kappa } I^M_{\theta_0}(\vart)$. Now if $I^M_{\theta_0}(\{\|\vart\| \geq \kappa\}) >0$,  then choosing $\ep = I^M_{\theta_0}(\{\|\vart\| \geq \kappa\})/2$, it follows that for all $T \geq T_0,$
$$ \PP_{\theta_0}\lf[\|a_T(\widehat\theta_T - \theta_0)\| \geq \kappa\ri] \leq e^{- b_T I^M_{\theta_0}(\{\|\vart\| \geq \kappa\})/2}$$ 
which in turn yields \eqref{eq:prob-conv-mdp}. Thus the assertion  follows upon establishing the following claim.

\np
{\em Claim:} For any $\kappa>0,$ $I^M_{\theta_0}(\{\|\vart\| \geq \kappa\}) >0.$

To show this, suppose $I^M_{\theta_0}(\{\|\vart\| \geq \kappa\}) = 0$. Then there exists a sequence $\{\vart_n\} \subset \{\vart: \|\vart\| \geq \kappa\}$ such that $\lim_{n \rt \infty}I^M_{\theta_0}(\vart_n) =0. $ In particular, $\{I^M_{\theta_0}(\vart_n)\}$ is bounded, i.e., for some $B>0$, $I^M_{\theta_0}(\vart_n) \leq B$. Since $I^M_{\theta_0}$ is good, the sublevel set $\{\vart: I^M_{\theta_0}(\vart) \leq B\}$ is compact; hence there exists a limit point $\tilde\vart$ and a subsequence $\{\vart_{n_k}\} \subset \{\vart: \|\vart\| \geq \kappa\} \cap \{\vart: I^M_{\theta_0}(\vart) \leq B\} $ such that $\vart_{n_k} \stackrel{k\rt \infty} \Rt \tilde \vart$. Clearly, $\tilde \vart \neq 0$. By lower semicontinuity of $I^M_{\theta_0}$,   
 $$0 = \lim_{k \rt \infty}I^M_{\theta_0}(\vart_{n_k}) \geq I^M_{\theta_0}(\tilde\vart).$$
 Thus $I^M_{\theta_0}(\tilde\vart) =0$ violating the hypothesis that $I^M_{\theta_0}(\vart)>0,$   for $\vart \neq 0$. This proves the claim.

\end{proof}
}

\section{Closed-form confidence intervals for Section~\ref{ssec:cts:state:unknown:mean}}
\label{app:closed-form-ci}
This appendix provides explicit closed-form expressions for the optimal
confidence intervals derived in Section~\ref{ssec:cts:state:unknown:mean} for the continuous state
i.i.d.\ model with unknown mean. These expressions follow from the
general construction in \eqref{eq:optJ} combined with the quadratic moderate
deviation rate functions of Theorem~\ref{thm:MDP:cramer} and Table~\ref{tab:rate_functions}. We discuss two natural choices of loss functions.

\subsection{Closed-form intervals for the mean functional}
\label{app:mean-functional}

We consider the one-dimensional setting and the canonical functional
\[
J(\theta) = \theta.
\]
Let $\widehat\theta_T$ be the sample mean estimator \eqref{eq:est:unknown:mean} and define the sequence
$\eta_T := \frac{2 r b_T}{T}$ which converges to $0$ as $T\to\infty$.
From \eqref{eq:optJ}, the optimal confidence interval is obtained by 
optimizing $J(\theta)$ over the set
$$
\{\theta\in\Theta : I^M_\theta(a_T(\widehat\theta_T -\theta))\leq r \} = \{\theta\in\Theta : (\widehat\theta_T - \theta)^2 \le \eta_T C(\theta) \},
$$
where $C(\theta)$ is given in Table~\ref{tab:rate_functions}. In all cases where this inequality in the set above
reduces to a quadratic in $\theta$, the interval admits the symmetric
representation 
\begin{equation}\label{app:closed:form:optimal:CI}
\mathcal{I}^\star_{T,r}(\widehat\theta_T) = [h_T \pm \kappa_T],
\end{equation}
where $h_T$ denotes its midpoint of the two roots and $\kappa_T$ is half their
distance. The models in Table~\ref{tab:rate_functions} all lead to a closed-form expression of the optimal confidence interval \eqref{app:closed:form:optimal:CI}, as listed below.

\paragraph*{(a) Normal distribution ($C(\theta)=\sigma^2$)}
\[
h_T = \widehat\theta_T,
\qquad
\kappa_T = \sigma \sqrt{\eta_T}.
\]

\paragraph*{(b) Exponential distribution ($C(\theta)=\theta^2$, $\theta>0$)}
For $\eta_T < 1$,
\[
h_T = \frac{\widehat\theta_T}{1-\eta_T},
\qquad
\kappa_T = \frac{\widehat\theta_T \sqrt{\eta_T}}{1-\eta_T}.
\]

\paragraph*{(c) Gamma distribution ($C(\theta)=\nu\theta$, $\theta>0$)}
\[
h_T = \widehat\theta_T + \frac{\eta_T \nu}{2},
\qquad
\kappa_T =
\frac{1}{2}
\sqrt{\eta_T \nu (\eta_T \nu + 4 \widehat\theta_T)}.
\]

\paragraph*{(d) Poisson distribution ($C(\theta)=\theta$, $\theta>0$)}
\[
h_T = \widehat\theta_T + \frac{\eta_T}{2},
\qquad
\kappa_T =
\frac{1}{2}
\sqrt{\eta_T(\eta_T + 4\widehat\theta_T)}.
\]

\paragraph*{(e) Bernoulli distribution ($C(\theta)=\theta(1-\theta)$, $\theta\in(0,1)$)}
\[
h_T = \frac{2\widehat\theta_T + \eta_T}{2(1+\eta_T)},
\qquad
\kappa_T =
\frac{1}{2(1+\eta_T)}
\sqrt{\eta_T^2 + 4\eta_T \widehat\theta_T (1-\widehat\theta_T)}.
\]

\paragraph*{(f) Binomial distribution ($C(\theta)=\theta(1-\theta/m)$)}
\[
h_T = \frac{2\widehat\theta_T + \eta_T}{2(1+\eta_T/m)},
\qquad
\kappa_T =
\frac{1}{2(1+\eta_T/m)}
\sqrt{\eta_T^2 + 4\eta_T \widehat\theta_T \Bigl(1-\widehat\theta_T /m\Bigr)}.
\]

\paragraph*{(g) Geometric distribution ($C(\theta)=\theta(\theta-1)$, $\theta>1$)}
For $\eta_T < 1$,
\[
h_T = \frac{2\widehat\theta_T - \eta_T}{2(1-\eta_T)},
\qquad
\kappa_T =
\frac{1}{2(1-\eta_T)}
\sqrt{\eta_T^2 + 4\eta_T \widehat\theta_T (\widehat\theta_T - 1)}.
\]

\subsection{Affine loss in the multivariate Gaussian setting}
\label{app:affine-gaussian}

Consider the multivariate Gaussian model
\[
X \sim \mathcal{N}(\theta, \Sigma),
\]
where $\theta \in \mathbb{R}^d$ is unknown and $\Sigma$ is known and
positive definite. 
Let $\widehat\theta_T$ be the sample mean estimator \eqref{eq:est:unknown:mean}.
From Theorem~\ref{thm:MDP:cramer}, the moderate deviation rate function is
\[
I_\theta^M(\vartheta)
=
\frac{1}{2}\,\vartheta^\top \Sigma^{-1} \vartheta.
\]
Consider an affine loss
\[
\ell(x) = q^\top x + \beta,
\qquad q \in \mathbb{R}^d,\;\beta \in \mathbb{R},
\]
so that
\[
J(\theta) = q^\top \theta + \beta.
\]
Using \eqref{eq:optJ}, the constraint becomes
\[
(\widehat\theta_T - \theta)^\top \Sigma^{-1} (\widehat\theta_T - \theta)
\le \frac{2 r b_T}{T}.
\]
Thus, the optimal confidence interval is obtained by optimizing a linear
functional over an ellipsoid, yielding the symmetric form \eqref{app:closed:form:optimal:CI}
with
\[
h_T = q^\top \widehat\theta_T + \beta,
\qquad
\kappa_T =
\sqrt{\frac{2 r b_T}{T}}\, \sqrt{q^\top \Sigma q}.
\]
This shows that, in the Gaussian case with affine loss, the optimal
confidence interval reduces to an explicit ellipsoidal projection
determined by the moderate deviation rate function.

\end{appendix}

\begin{funding}
Research of A. Ganguly is supported in part by NSF DMS - 1855788 and NSF DMS - 2246815.
Research of T. Sutter is supported in part by DFG in the Cluster of Excellence EXC 2117 “Centre for the Advanced Study of Collective Behaviour” (Project-ID 390829875).
\end{funding}

\bibliographystyle{imsart-number}
\bibliography{references}

\end{document}